\documentclass[default,iicol]{sn-jnl}

\usepackage[utf8]{inputenc}         
\usepackage[T1]{fontenc}            
\usepackage{hyperref}               
\usepackage{url}                    
\usepackage{nicefrac}               
\usepackage{microtype}              
\usepackage{amsfonts}               
\usepackage{longtable}
\usepackage{lscape}
\usepackage{array,rotating,amsmath,booktabs,graphicx}
\usepackage{adjustbox}
\usepackage{geometry}
\usepackage{comment}
\usepackage{graphicx}
\usepackage{subfig}
\usepackage{gensymb}
\usepackage[super]{nth}
\usepackage{multicol, blindtext}
\usepackage{xcolor,colortbl}
\usepackage{float}
\usepackage{scrextend}
\usepackage{acronym}

\jyear{2021}%

\theoremstyle{thmstyleone}%
\theoremstyle{thmstyletwo}%

\theoremstyle{thmstylethree}%

\acrodef{DNN} 		[\textsc{DNN\xspace}]				{Deep Neural Networks}
\acrodef{CNN} 		[\textsc{CNN\xspace}]				{Convolutional Neural Network}
\acrodef{FCN} 		[\textsc{FCN\xspace}]				{Fully Convolutional Networks}
\acrodef{SVM} 		[\textsc{SVM\xspace}]				{Support Vector Machine}
\acrodef{mAP} 		[mAP]				{mean Average Precision}
\acrodef{mIoU} 		[mIoU]				{mean Intersection over Union}
\acrodef{ICDAR} 		[\textsc{ICDAR\xspace}]				{International Conference on Document Analysis and Recognition}

\acrodef{ICFHR} 		[\textsc{ICFHR\xspace}]				{International Conference on Frontiers of Handwriting Recognition}

\acrodef{DIAR}      [\textsc{DIAR\xspace}]              {Historical Document Image Analysis and Recognition}

\acrodef{FM}      [\textsc{FM\xspace}]              {F-Measure}
\begin{document}

\title[Historical Document Image Datasets]{A Survey of Historical Document Image Datasets}

\acrodef{OCR} 		[\textsc{OCR\xspace}]				{Optical Character Recognition}
\acrodef{WA} 		[\textsc{WA\xspace}]				{Weighted Average}

\acrodef{WER} 		[\textsc{WER\xspace}]				{Word Error Rate}

\acrodef{CER} 		[\textsc{CER\xspace}]				{Character Error Rate}
\acrodef{DR} 		[\textsc{DR\xspace}]				{Detection Rate}
\acrodef{VLAD} 		[\textsc{VLAD\xspace}]				{Vector of Locally Aggregated Descriptors}

\acrodef{CTC} 		[\textsc{CTC\xspace}]				{Connectionist Temporal Classification}

\acrodef{LSTM} 		[\textsc{LSTM\xspace}]				{Long Short Term Memory}

\acrodef{acc} 		[Acc]				{Accuracy}
\acrodef{RA}[\textsc{RA\xspace}]				{Recognition Accuracy}
\acrodef{ca} 		[\textsc{CA\xspace}]				{Character Accuracy}


\author*[1]{\fnm{Konstantina} \sur{Nikolaidou}}\email{konstantina.nikolaidou@ltu.se}

\author[2]{\fnm{Mathias} \sur{Seuret}}\email{mathias.seuret@fau.de}
\author[1]{\fnm{Hamam} \sur{Mokayed}}\email{hamam.mokayed@ltu.se}
\author[1]{\fnm{Marcus} \sur{Liwicki}}\email{marcus.liwicki@ltu.se}

\affil[1]{\orgdiv{EISLAB Machine Learning Group}, \orgname{Luleå University of Technology}, \orgaddress{\street{Aurorum 1}, \city{Luleå}, \postcode{97187}, \state{Norrbotten}, \country{Sweden}}}

\affil[2]{\orgdiv{Pattern Recognition Lab Computer Vision Group}, \orgname{Friedrich-Alexander-Universität}, \orgaddress{\street{Martensstr. 3}, \city{Erlangen}, \postcode{91058}, \state{Bavaria}, \country{Germany}}}


\abstract{This paper presents a systematic literature review of image datasets for document image analysis, focusing on historical documents, such as handwritten manuscripts and early prints. Finding appropriate datasets for historical document analysis is a crucial prerequisite to facilitate research using different machine learning algorithms. However, because of the very large variety of the actual data (e.g., scripts, tasks, dates, support systems, and amount of deterioration), the different formats for data and label representation, and the different evaluation processes and benchmarks, finding appropriate datasets is a difficult task. This work fills this gap, presenting a meta-study on existing datasets. After a systematic selection process (according to PRISMA guidelines), we select 65 studies that are chosen based on different factors, such as the year of publication, number of methods implemented in the article, reliability of the chosen algorithms, dataset size, and journal outlet. We summarize each study by assigning it to one of three pre-defined tasks: document classification, layout structure, or content analysis. We present the statistics, document type, language, tasks, input visual aspects, and ground truth information for every dataset. In addition, we provide the benchmark tasks and results from these papers or recent competitions. We further discuss gaps and challenges in this domain. We advocate for providing conversion tools to common formats (e.g., COCO format for computer vision tasks) and always providing a set of evaluation metrics, instead of just one, to make results comparable across studies.
}




\maketitle

\section{Introduction}
\label{sec:introduction}

In the last decade, deep neural networks have been the state-of-the-art in challenging domains \cite{jimaging6100110}. 
The key to the success of deep neural networks is the availability of data with ground truth used for training and evaluation of such methods.
Although this is the case for Document Image Analysis \cite{1359749}, especially for modern documents \cite{harley2015evaluation} and scene text detection and recognition \cite{6628859, 7333942}, the performance remains low for historical documents compared to other Computer Vision problems that use very large databases \cite{russakovsky2015imagenet}.

As deep learning networks require a large amount of data, large datasets are required.
In recent years, a variety of datasets have appeared in journals, conference proceedings, and the competitions they host, for tasks such as document classification, word spotting, layout analysis, graphical object detection and handwriting recognition for both modern and historical document images. 
Despite their complexity due to inevitable degradation by centuries of usage and their variability, the analysis of historical documents has attracted much interest. 
The growth of digital libraries has contributed to the research on historical documents by providing high-quality digitized images to researchers to process and analyze. 
As a result, researchers have introduced several datasets by managing and annotating the collections provided by these libraries.

\subsection{Purpose and Contributions}

The main goal of this work is to provide researchers with an overview of datasets and machine learning tasks. 
In particular, we give an overview of the publicly available historical document image datasets and report the benchmark tasks and results.
We also report the results based on some datasets that have been used in recent competitions.
Thus, we refer to the appropriate method for every task and dataset while finding gaps and challenges in the field.

The major contributions of this work can be summarized as follows:
\begin{itemize}
    \item A systematic literature review of historical document image datasets is presented.
    \item A summary of 65 historical document image studies grouped into document classification, document structure, and content analysis related tasks.
    \item A tabular overview of the statistics, classes, tasks, languages, type of document, input visual aspects, ground truth information, and benchmarks for every dataset.
    \item A discussion on the challenges, gaps and future research directions is provided.
\end{itemize}


\begin{figure*}[htp]
\centering
   \subfloat[]{\label{original}
      \includegraphics[width=.22\textwidth]{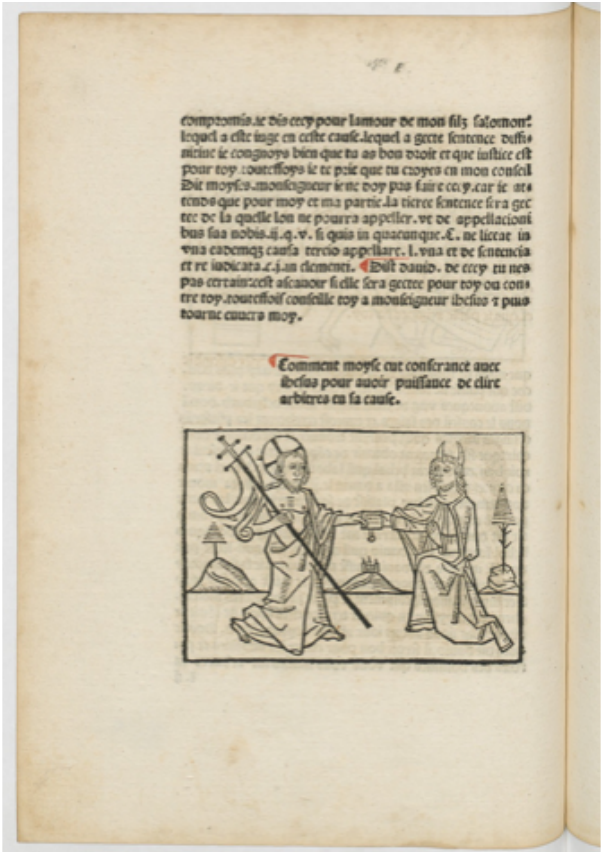}}
~
   \subfloat[]{\label{metadata}
      \includegraphics[width=.22\textwidth]{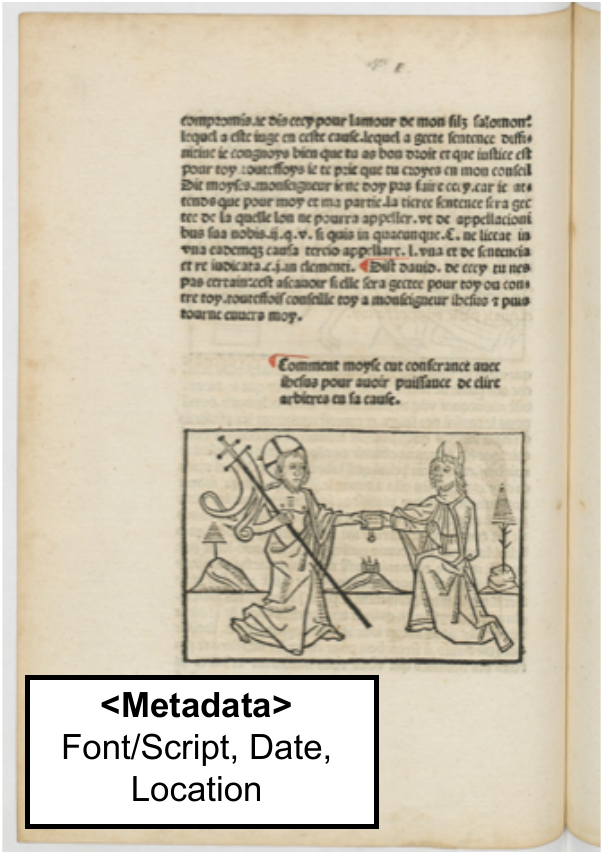}}
      ~
  \subfloat[]{\label{layout}
  \includegraphics[width=.22\textwidth]{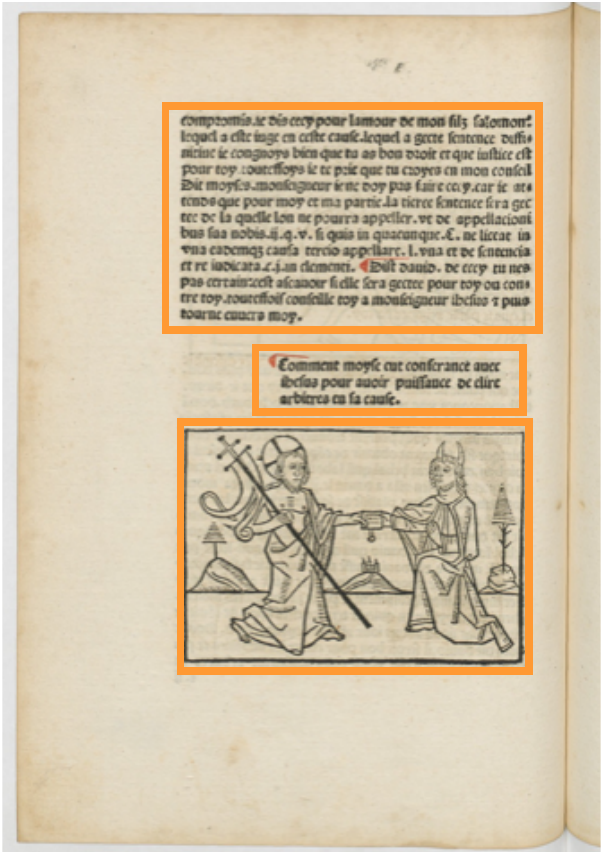}}
  ~
    \subfloat[]{\label{semantic}
      \includegraphics[width=.22\textwidth]{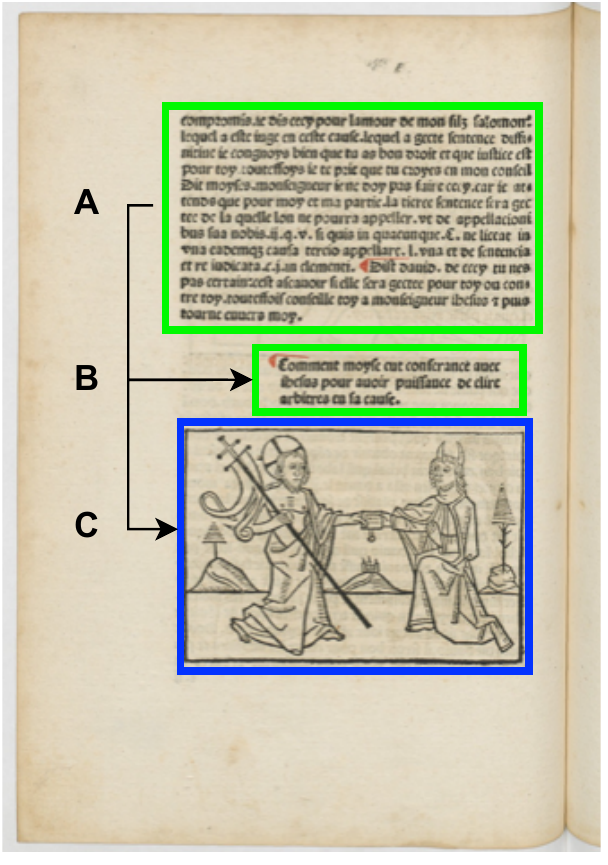}}
   \caption{(a) A document page and its example (b) metadata information related to document classification, (c) document structure, and (d) content analysis.}\label{doc_geometrical_logical}
   \label{fig:meta_structure_content}
\end{figure*}

\subsection{Selection Methodology}
For the systematic literature review, we searched for the datasets presented at the \ac{ICDAR} and the \ac{ICFHR} over a time span of six years (2016-2021). 
For dataset identification, we further used the keyword queries "historical", "document", "image", "analysis" and "dataset" in the Google Scholar academic research database. %
We limited our search to the first 98 pages, or 980 results out of the 188K total results, as the database does not seem to operate after that number of data. 
We filtered the suitable papers by reading the abstracts and excluding those unrelated to historical document image datasets.
Furthermore, through backward snow-balling related works, we included a few datasets that were not retrieved by the database.
The final number of studies that we present in this paper is 65.

\subsection{Scope}

This paper mainly focuses on technical aspects that could facilitate machine learning researchers in making full use of available datasets.
Other aspects that may be more relevant for experts from the humanities field, such as historical and paleographic analyses to understand the context of a historical document and how it was used, are beyond the scope of this paper. 
For every dataset, we summarize the size, language, targeting tasks, type of annotations, and benchmark results.
We further provide detailed tables that include dataset statistics, visual aspects of the input images, ground truth format, and the provided benchmarks.

\begin{figure*}[ht]
\centering
\includegraphics[width=0.99\textwidth, scale=0.5]{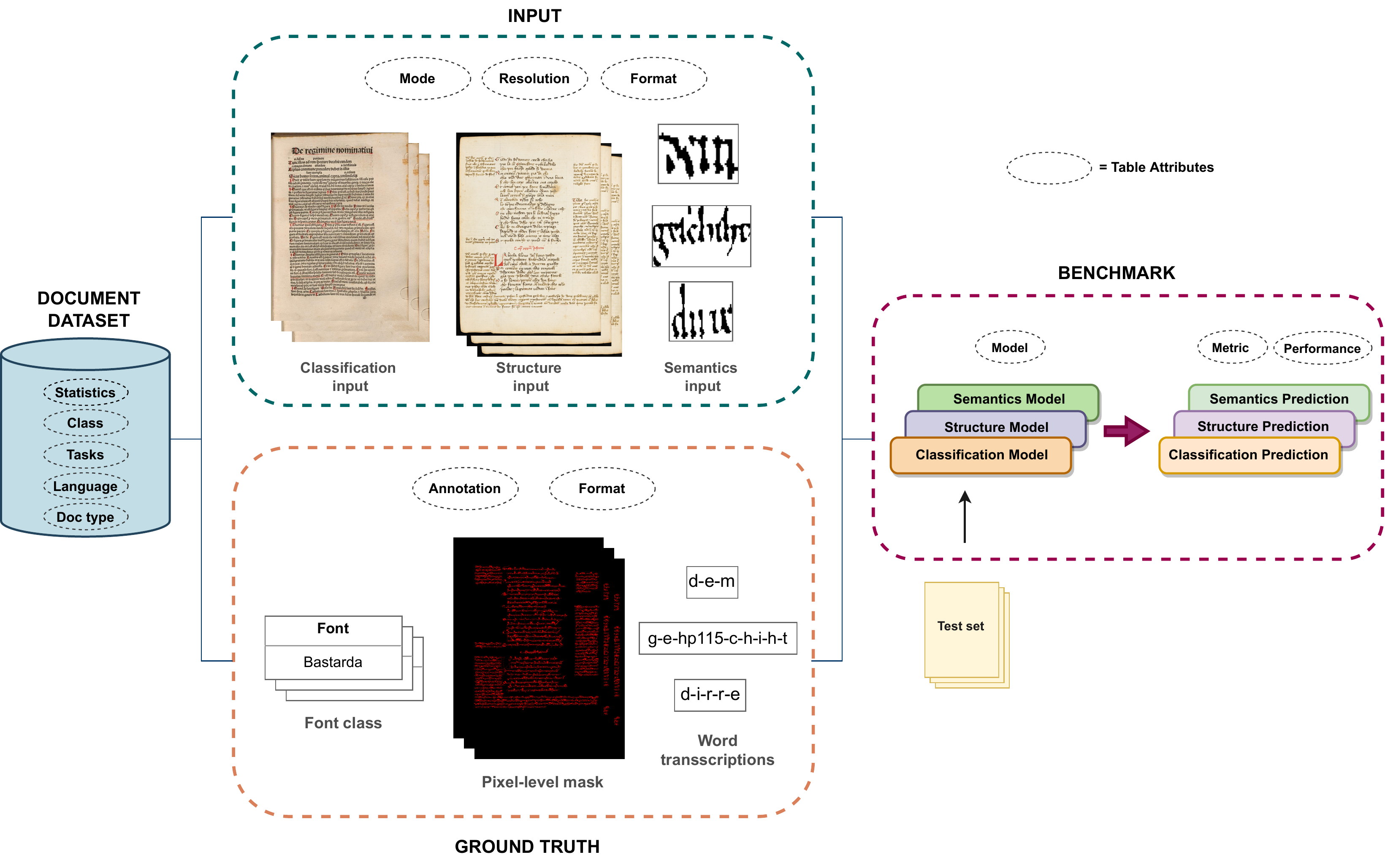}
\caption{Illustration of Table \ref{tab:large_table} and its attributes. A document image dataset that can be characterized by different statistics, languages, and types of documents. The dataset contains input images for different tasks, of specific visual aspects (mode, resolution and format) and their corresponding annotation type and format. Then benchmarks with different models, metrics, and performances are created for the different tasks.}
\label{fig:mesh1}
\end{figure*}

The study is organized as follows. 
In Section \ref{sec:datasets}, we report the existing datasets for historical document image analysis, organizing them according to the following categories: 
\begin{enumerate}
    \item Document classification
    \item Document structure 
   \item Content analysis
\end{enumerate}
Subsections \ref{ssec:classification}, \ref{ssec:structure}, and \ref{ssec:content} include studies related to the aforementioned categories.
Furthermore, Table \ref{tab:structure_content_class} shows the listed datasets and their corresponding section, release year, writing type (handwritten or printed), and checkmarks according to the information and tasks the are or they could be used for.
Finally, we present a large table (see Table \ref{tab:large_table}) with information on the reviewed datasets considering the statistics, classes, tasks, languages, document type, input image aspects, ground truth, and benchmarks. 
We should note that to fully understand the table, the reader should address the corresponding section of every dataset. 
An illustration of this table can be seen in Figure \ref{fig:mesh1}.
In Section \ref{sec:discussion}, we discuss observations, challenges and future directions for the domains of interest.
Finally, we present our conclusions in Section \ref{sec:conclusion}.

\section{Related Work}
\label{sec:related_work}

Several works addressed the topic of document image analysis as surveys. 
The work presented in \cite{Tang1996AutomaticDP} focused on the task of automatic document processing. 
This task was split into document analysis and document understanding, related to the layout structure and logical structure of a document, respectively. 
Lombardi and Marinai \cite{jimaging6100110} surveyed papers that use deep learning methods for historical document image analysis by showing the connections between input and output for all methods and according to task.
This paper further presented some historical datasets.
Other surveys are task-specific.
\cite{Hussain2015ACS} provided a survey focusing on databases and benchmarks for handwriting recognition.
Furthermore, \cite{824821} focused on both on-line and off-line handwriting recognition.
Likforman-Sulem et al. \cite{LikformanSulem2006TextLS} focused on automatic text-line segmentation methods for historical documents.
\cite{10.1145/3476887.3476888} concentrated on evaluation metrics and tools for \ac{OCR}.
Finally, other works focus on layout analysis \cite{Namboodiri2007} or word spotting techniques \cite{10.1016/j.patcog.2017.02.023}.
The work presented in \cite{10.1145/3355610}, surveyed layout analysis techniques and listed several historical and modern datasets related to the task.
To the best of our knowledge, we present the first literature review that systematically focuses on historical datasets and benchmarks.

\section{Datasets}
\label{sec:datasets}

Various datasets exist for historical document image analysis for different tasks, such as layout analysis, baseline detection, handwriting recognition, binarization, and writer identification. 
In this review, we group the tasks discovered through our methodology as subtasks of document classification, document structure, and content analysis. 
Moreover, the main components of a document are its metadata, structure, and content as shown in Figure \ref{fig:meta_structure_content}.
Notably, the actual document analysis pipelines can become much more complex in real scenarios\footnote{\url{https://ocr-d.de/en/about\#the-ocr-d-project}\label{ocr-d}}, but for a general grouping of the tasks, these three categories are sufficient.
We present the existing historical document image datasets considering these components and order them by release date (earliest to latest) within every subsection.
Furthermore, a tabular overview of these datasets and checkmarks on the subtasks they include according to their ground truth is presented in Table \ref{tab:structure_content_class}.
Since several datasets can belong to more than one task category, we place them in the first section where a benchmark task exists, however, in Table \ref{tab:structure_content_class} we checkmark all possible tasks.


\begin{table*}[hp]

\centering
\caption{Listed datasets and their release year, summary section, document type, whether the dataset is a part of a competition, and information about the different tasks that are divided according to document classification-level, structure, or content analysis content. The datasets are sorted from earliest to latest release year.}
\begin{adjustbox}{scale=0.56,center}
\renewcommand\arraystretch{1.2}
\begin{tabular}{lclcc|cccc|ccccc|cccccc}
  \multicolumn{5}{c}{} & 
  \multicolumn{4}{c}{\textbf{\begin{tabular}{c}
  Document\\Classification
  \end{tabular}}} & 
  \multicolumn{5}{c}{\textbf{\begin{tabular}{c}
  Document\\Structure
  \end{tabular}}} &
  \multicolumn{6}{c}{\textbf{\begin{tabular}{c}
  Content \\Analysis
  \end{tabular}}} 
  \\ 
  \cmidrule(lr){6-9}
  \cmidrule(lr){10-14}
  \cmidrule(lr){15-20}

 \textbf{Dataset}&\textbf{\begin{tabular}{c}
Year
 \end{tabular} }&\textbf{Section} &
 \textbf{Handwritten/Printed} &\textbf{Competition} &
 \rotatebox[origin=c]{90}{\textbf{Date}} & \rotatebox[origin=c]{90}{\textbf{Font}} & \rotatebox[origin=c]{90}{\textbf{Script}} & \rotatebox[origin=c]{90}{\textbf{Location}} &
 \rotatebox[origin=c]{90}{\textbf{Binarization}} & \rotatebox[origin=c]{90}{\textbf{Layout}} & \rotatebox[origin=c]{90}{\textbf{Text-Line}} &  \rotatebox[origin=c]{90}{\textbf{Table}} & \rotatebox[origin=c]{90}{\textbf{Graphics}} & \rotatebox[origin=c]{90}{\textbf{OCR}} & 
 \rotatebox[origin=c]{90}{\textbf{Retrieval}} 
 & \rotatebox[origin=c]{90}{\textbf{Digits}} & \rotatebox[origin=c]{90}{\textbf{Table}} & \rotatebox[origin=c]{90}{\textbf{Writer}} & \rotatebox[origin=c]{90}{\textbf{Order}}\\[3.9ex]
 \botrule

 GERMANA \cite{5277691}& 2009 &\ref{sssec:germana} &handwritten&&&&&&& \checkmark & \checkmark   & & & \checkmark &&&&&\\
 \hline
 
 RODRIGO \cite{serrano-etal-2010-rodrigo}& 2010 &\ref{sssec:rodrigo}& handwritten&& &&&& &\checkmark & \checkmark   & & & \checkmark &&&&&\\
 \hline
 
 IAM-HistDB \cite{10.1145/1815330.1815331} & 2010 &\ref{sssec:iam-histdb}&handwritten&& &&&&   & &  & & &\checkmark&&&&&\\
 \hline
 
 PHTD \cite{6121553}& 2011 &\ref{sssec:phtd} &handwritten&&&&& &&  & \checkmark   & & &\checkmark &&&&&\\
 \hline
 
 PBOK \cite{Alaei2012DatasetAG}& 2012 &\ref{sssec:pbok} &handwritten&&&&&&&  & \checkmark  & & &\checkmark&&&& &\\
 \hline
 
 IMPACT \cite{10.1145/2501115.2501130} & 2013 &\ref{sssec:impact}& both &&&&&& & \checkmark & \checkmark  &\checkmark&\checkmark&\checkmark&&&&&\checkmark\\
 \hline

 ESPOSALLES \cite{ROMERO20131658}& 2013&\ref{sssec:esposalles}& handwritten &\checkmark&&& && & \checkmark & \checkmark   & & &\checkmark &&&&&\\
 \hline
 
 BH2M \cite{6976764}& 2014 &\ref{sssec:bh2m} &handwritten&&\checkmark&&&&& \checkmark & \checkmark &  & &\checkmark &\checkmark&&& &\checkmark\\
 \hline

 HADARA80P \cite{6980990}& 2014 &\ref{sssec:hadara80p}&handwritten&& &&&& & \checkmark &   & & &\checkmark&\checkmark&&&&\\
 \hline

 ENP \cite{7333898}& 2015 &\ref{sssec:enp} &printed&&&&&&& \checkmark &    &\checkmark &\checkmark&\checkmark&\checkmark&&&&\\
 \hline

 GRPOLY-DB \cite{7333841}& 2015 &\ref{sssec:GRPOLY-DB} &both&&&&&&&  & \checkmark   & &&\checkmark&\checkmark&&&&\\
 \hline
    
 DocExplore \cite{10.1117/1.JEI.26.1.011010}& 2016 &\ref{sssec:docexplore}&both&&&&&& &  &  &  & & \checkmark & \checkmark &&&&\\
 \hline

 DIVA-HisDB \cite{7814109}& 2016&\ref{sssec:diva-hisdb}&handwritten&\checkmark &&&&&& \checkmark & \checkmark   & & &&&&& &\\
 \hline

 AMADI\_LontarSet \cite{7814058}&2016 &\ref{sssec:amadi_lontarset}&handwritten&\checkmark &&&&& \checkmark &  &  & & &\checkmark & \checkmark &&&& \\
 \hline

 ICFHR16 CLaMM \cite{7814129}& 2016 &\ref{sssec:ICFHR16CLAMM}&handwritten&\checkmark&&&\checkmark&&    &  & &  &&&&&&&\\
 \hline
 
 ICDAR17 CLaMM \cite{8270155} & 2017&\ref{sssec:ICDAR17CLAMM}&handwritten&\checkmark&\checkmark&&\checkmark&&    &  & &  &&&&&&&\\
 \hline
 
 HBA 1.0\cite{mehri:hal-01637826}& 2017&\ref{sssec:hba}&both&\checkmark&&&&&& \checkmark   &  & & \checkmark &&&&&&\\
 \hline

 SleukRith \cite{10.1145/3151509.3151510}& 2017 &\ref{sssec:sleukrith}&handwritten &\checkmark&&&&& \checkmark& & \checkmark   & & &\checkmark &&&& &\\
 \hline

 VML-HD \cite{8067751}& 2017&\ref{sssec:vml-hd}&handwritten &&&&&& &  &    & & &\checkmark&\checkmark&&&& \\
 \hline

 CFRAMUZ \cite{Arvanitopoulos2017AHF}& 2017 
 &\ref{sssec:cframuz}&handwritten&&&&&& &  & &  &  &\checkmark&\checkmark&&& &\\
 \hline
 
 Lontar Sunda \cite{8270066}& 2017 
 &\ref{sssec:SUNDANESE_PALM_LEAF}&handwritten& \checkmark&&&&&\checkmark&  &\checkmark   & & &\checkmark &&&& &\\
 \hline

 ICDAR17 REID2017 \cite{8270161} & 2017 &\ref{sssec:reid2017}&printed&\checkmark&&&&& & \checkmark  &  & & \checkmark&\checkmark&&&& &\\
 \hline

 ICDAR17 Historical-WI \cite{8270156} & 2017 &\ref{sssec:Historical-WI}&handwritten&\checkmark &&&&&  & &  & & &&\checkmark& & &\checkmark &\\
 \hline
 
 Kuzushiji \cite{Clanuwat2018DeepLF} & 2018 &\ref{sssec:K-MNIST}&printed& &&&&&&  &   & & &\checkmark&& & & &\\
 \hline

 READ-BAD \cite{8395221}& 2018 &\ref{sssec:read-bad}&printed&\checkmark &&&&& & & \checkmark  & & &&&&& &\\
 \hline
 
 Warped Arabic \cite{iet:/content/conferences/10.1049/cp.2018.1286}& 2018 &\ref{sssec:warped_arabic}&both& &\checkmark&\checkmark&\checkmark&\checkmark & &  & \checkmark  & &&&&& &&\\
 \hline
 
 MHDID \cite{8480372}& 2018 &\ref{sssec:mhdid} &handwritten&&&&&&&  & &  & & &  (\checkmark)&&& &\\
 \hline
 
 Tripitaka \cite{Yang2018DenseAT}& 2018 &\ref{sssec:tripitaka}& handwritten&&&&&& & \checkmark &   & & &\checkmark&&&&& \\
 \hline
 
 KERTAS \cite{adam2018kertas}&2018 &\ref{sssec:kertas} &handwritten&&\checkmark&&&&  & &  & & &&&&&\checkmark& \\
 \hline
 
 ICFHR18 RASM2018 \cite{8583806}&2018 &\ref{sssec:rasm2018} &handwritten&\checkmark&&&&&& \checkmark & \checkmark  & & \checkmark&\checkmark&&&& &\checkmark\\
 \hline
 
 ICFHR18 Asian Palm Leaf \cite{8583808}&2018 &\ref{sssec:ICFHR18_ASIAN_PALM_LEAF} &handwritten&\checkmark&&&&&\checkmark&  & \checkmark  & & &\checkmark&&&& &\\
 \hline
 
 Oficio de Hipotecas de Girona (OHG) & 2018 &\ref{sssec:OHG}&handwritten&&&&&&& \checkmark &    & & &\checkmark&&&&& \\
 \hline

 ARDIS \cite{kusetogullari2020ardis}& 2019 &\ref{sssec:ardis} &handwritten&&&&&&&  &  &  & & &&\checkmark&& &\\
 \hline

 Pinkas \cite{8978129}& 2019 &\ref{sssec:pinkas}&handwritten&&&&&&& \checkmark & \checkmark   & & &&&&&& \\
 \hline
 
 BADAM \cite{Kiessling2019BADAMAP}& 2019 &\ref{sssec:badam}&handwritten&&&&&& &  & \checkmark  & & &&&&& &\\
 \hline

 HORAE \cite{10.1145/3352631.3352633}& 2019&\ref{sssec:horae}&handwritten&& &&&&& \checkmark&\checkmark   & & \checkmark& &&&& &\\
 \hline
 
 
 ICDAR19 cTDaR19 \cite{8978120}& 2019 & \ref{sssec:ctdar} &handwritten&\checkmark &&&&&  &  &  &\checkmark & &&&&\checkmark&& \\
 \hline

 ICDAR19 DMAS2019\footref{dmas2019}  & 2019 &\ref{sssec:DMAS2019}& printed &\checkmark&&&& && \checkmark  &  & & \checkmark & \checkmark&&&&& \\
 \hline

 ICDAR19 DIBCO 2019 \cite{8978205}  & 2019 &\ref{sssec:DIBCO_2019}& both &\checkmark&&&& &\checkmark &  &  & &  & &&&&& \\
 \hline
 
 OBC306 \cite{8978032}& 2019 &\ref{sssec:obc306}&handwritten&&&&&& &  &    & & &\checkmark&&&& &\\
 \hline
 
 GRK-Papyri \cite{8978142} \& PapyRow \cite{10.1007/978-3-030-68787-8_16}& 2019 \& 2021 &\ref{sssec:grk-papyri}&handwritten &&&&&&\checkmark&    &  & & &&\checkmark&&&\checkmark &\\
 \hline
 
 CASIA-AHCDB \cite{8978010}& 2019 &\ref{sssec:casia-ahcdb}& handwritten&&&&&&&  &  &  &  &\checkmark&&&& &\\
 \hline

 Amharic Database \cite{8977980}& 2019 &\ref{sssec:amharic}& printed&&&&&& &  &  &  & &  \checkmark&&&& &\\
 \hline
 
 Multiple Font Groups \cite{10.1145/3352631.3352640}&2019 &\ref{sssec:fonts} &printed&\checkmark&&\checkmark&&&  &  &  & & &&&&&& \\
 \hline

 ICDAR19 HDRC-Chinese \cite{8977999}&2019 & \ref{sssec:chinese_reading_challenge}&handwritten& \checkmark &&&&&& \checkmark & \checkmark   & & &\checkmark&&&& &\\
 \hline

 ICDAR19 REID2019 \cite{8978191} & 2019 &\ref{sssec:reid2019}&printed&\checkmark&&&&& & \checkmark  &  & & \checkmark&\checkmark&&&& &\\
 \hline

 ABP \& NAF \cite{8978117}& 2020 &\ref{sssec:ABP_NAF}&handwritten&&&&&&&\checkmark &   & \checkmark  & &  & &&& & \\
 \hline
 
 FCR \cite{Quirs2020FinnishCR}& 2020 &\ref{sssec:FCR}&handwritten&&&&&& &  \checkmark & \checkmark  & \checkmark &  &\checkmark &&&\checkmark & &\\
 \hline

 IlluHisDoc \cite{monnier2020docExtractor}& 2020 &\ref{sssec:illuhisdoc}&printed&&&&&& &  & \checkmark  & & \checkmark &&&&& &\\
 \hline
 
 Newspaper Navigator \cite{10.1145/3340531.3412767}& 2020 &\ref{sssec:newspaper_navigator}&printed&&\checkmark&&&& & \checkmark&   & &\checkmark &\checkmark&&&& &\\
 \hline

 DIDA \cite{KUSETOGULLARI2021100182}& 2020 &\ref{sssec:dida}& handwritten&&&&&&&    &  & & &&&\checkmark&& &\\
 \hline
 
 ScribbleLens \cite{9257750}& 2020 &\ref{sssec:scribblelens}&handwritten&&\checkmark &&&&  &  &  & & &\checkmark&&&&\checkmark &\\
 \hline

 ICDAR19 RASM2019 &2019 &\ref{sssec:rasm2019} &handwritten&\checkmark&&&&&& \checkmark & \checkmark  & & \checkmark&\checkmark&&&& &\\
 \hline
 
 ICDAR19-HDRC-IR \cite{Christlein2019ICDAR2C}& 2019 &\ref{sssec:ICDAR2019-HDRC-IR}&handwritten&\checkmark&&&&&  &  & & &  &&\checkmark&&&\checkmark&\\
 \hline
 
 HTR Benchmarks \cite{SANCHEZ2019122}& 2019 &\ref{sssec:htr_benchmarks}&handwritten&\checkmark&&&&&  &  &\checkmark & &  &\checkmark&&&&&\\
 \hline
 
 HJDataset \cite{Shen_2020_CVPR_Workshops}& 2020 &\ref{sssec:hjdataset}& both &&&&&&& \checkmark &   & &  &&&&&&\checkmark\\
 \hline
 
 ICFHR20 HisFragIR20 \cite{Seuret2020ICFHR2C}& 2020 &\ref{sssec:HisFragIR20}&handwritten&\checkmark&&&&&  &  & & &  &&\checkmark&&&\checkmark&\\
 \hline

 ICDAR21 HDC \cite{10.1007/978-3-030-86337-1_41} & 2021 &\ref{sssec:his_doc_classification_icdar_2021}& both &\checkmark& \checkmark & \checkmark & \checkmark & \checkmark &  & &  & & &&&&&&\\
 \hline
 
 BIR-database \cite{10.1145/3476887.3476913} & 2021 & \ref{sssec:bir_database} &printed&& & \checkmark &  & && \checkmark  &  & & &&&&&&\\
 \hline

 GloSAT \cite{10.1145/3476887.3476890} & 2021 & \ref{sssec:GloSAT}&both&& &  &  && &  & &   \checkmark& &&&&&&\\
 
 \hline
 
 Digital Peter \cite{10.1145/3476887.3476892} & 2021 &
 \ref{sssec:digital_peter}& handwritten&\checkmark& &  &  && & &\checkmark &  & &\checkmark&&&& &\\
 
 \hline 
 
  BiblIA \cite{10.1145/3476887.3476896} & 2021 &
 \ref{sssec:BiblIA}& handwritten& & \checkmark&  & \checkmark && & &\checkmark &  & &\checkmark&&&& &\\
 
 \hline 
 
  HisClima \cite{9412210} & 2021 &
 \ref{sssec:hisclima}& handwritten& & &  &  && &\checkmark &\checkmark & \checkmark & &\checkmark&&&\checkmark& &\\

 \hline 
 
  Hugin-Munin \cite{10.1007/978-3-031-06555-2_27} & 2022 &
 \ref{sssec:hugin_munin}& handwritten& & &  &  && & & &  & &\checkmark&&&&\checkmark &\\
 
 \hline 
 
  POPP \cite{10.1007/978-3-031-06555-2_10} & 2022 &
 \ref{sssec:popp}& handwritten& & &  &  && & &\checkmark &  & &\checkmark&&&\checkmark&\checkmark &\\
 
 \botrule

\end{tabular}
\label{tab:structure_content_class}
\end{adjustbox}
\end{table*}


\subsection{Document Classification Datasets}
\label{ssec:classification}

To process and understand a document, it is vital to categorize it.
When and where was this document written? 
What font or script was used in this document (or parts of it, if revisited later)?
This information is helpful for the steps of document understanding, as it provides context to the researcher about the era, the writer, and more.
Document classification refers to the categorization according to a document’s geographical and chronological occurrence and its written script or font.  
In the following subsections, we summarize six datasets based on date/age, font/script, and location classifications.
Three studies refer to competitions highly related to each other, and the other three studies introduce datasets for font classification as the main task.

\subsubsection{ICFHR 2016 Competition on the Classification of Medieval Handwritings in Latin Script (ICFHR16 CLaMM) }
\label{sssec:ICFHR16CLAMM}

The ICFHR16 Competition \cite{7814129} provided a collection of grayscale Latin manuscript images for script type classification.
Two tasks were proposed in this competition: Task 1, which uses a single label for each image, and Task 2, which uses multi weighted labeling for each image. 
The dataset is comprised of 12 script classes and 3 sets, the training set contains 2K images and the test sets for Tasks 1 and 2 contain 1K and 2K images, respectively. 
The average \ac{acc} and the average intraclass distance were used for the evaluation of the proposed systems. The system that achieved the highest accuracy for Task 1 used I-vector extraction \cite{Dehak2011LanguageRV} on image patches and a classification of the extracted vectors using Latent Dirichlet Analysis (LDA). 
For Task 2, the best performing system in terms of Final Score utilized a neural network architecture named DeepScript\footnote{\url{https://github.com/mikekestemont/DeepScript}\label{deepscript}} and pre-processing to yield information on random image crops and their various perturbations to the network classifier as input.
In terms of the average intraclass distance, the higher ranked system in both tasks was the FRDC-OCR and consisted of a \ac{CNN} classifier that was applied on patches, where for every image, the result is the average of the recognition confidence and feature vector across its patches.

\subsubsection{ICDAR 2017 Competition on the Classification of Medieval Handwritings in Latin Script (ICDAR17 CLaMM) }
\label{sssec:ICDAR17CLAMM}

Similar to the ICFHR16 CLaMM presented in Section \ref{sssec:ICFHR16CLAMM}, tasks were proposed for the ICDAR17 CLaMM competition \cite{8270155}. 
Task 1 was a script type classification, and Task 2 was a script type classification on heterogeneously encoded data. The available training dataset for these tasks contains 3,500 manuscript images from the previous year’s competition. 
Three thousand of these training images were used and their labels were further extended for Task 3, manuscript date classification, and Task 4, manuscript date classification on heterogeneously encoded data. 
The date classification data were distributed across 15 classes, ranging from 500 C.E. to 1600 C.E. 
Tasks 1 and 3 were evaluated using a 2K image test set, while tasks 2 and 4 were evaluated using a 1K image test set. 
For Tasks 1 and 2, the evaluation criterion was the accuracy per script type, while for Tasks 3 and 4, the evaluation criterion was accuracy per date. 
The winners of Tasks 1 and 3 applied T-DeepCNN, a CNN with residual connections \cite{He2016DeepRL} and batch normalization \cite{10.5555/3045118.3045167} on patch images of 227 x 227 pixels, averaging over the patches of every image for a final prediction. 
They further enhanced the performance of their model by using an ensemble of 5 CNN classifiers. 
For Tasks 2 and 4, the winning approach (CK2) was a linear \ac{SVM} classifier with a squared hinge loss on vectors derived from the combination of the PCA-whitened RootSIFT local descriptors and the global vector of locally aggregated descriptors (VLAD).
This system was based on \cite{7333893, Christlein2017WriterIU}.

\subsubsection{KERTAS }
\label{sssec:kertas}

The KERTAS dataset \cite{adam2018kertas} consists of handwritten Arabic manuscripts from the Qatar National Library intended for age and writer detection.
This dataset contains 2,502 high-resolution document images and their corresponding date annotation according to the Islamic century in which they were written.
The dataset provides the additional source, manuscript name, description, writer name, and ID information in XML format for every manuscript.
Furthermore, an age detection algorithm based on sparse representations was introduced in this work and results on the dataset using different image size inputs were shown. 
This method was also compared with three different writing style feature algorithms, Run Length \cite{4107573}, Edge Direction \cite{BRINK2012162}, and Edge Hinge \cite{10.1016/j.patrec.2013.03.020}, with 3-NN as the classifier.
Both sets of experiments were made using predefined and random splits, and the predefined splits performed better in all cases.
The proposed method on the $50 \times 50$ image size achieved the best performance.

\subsubsection{Dataset of Pages from Early Printed Books with Multiple Font Groups }
\label{sssec:fonts}

This dataset \cite{10.1145/3352631.3352640} contains a set of 35,623 images for font classification and consists of 12 different classes: Antiqua, Italic, Textura, Rotunda, Gotico-Antiqua, Bastarda, Schwabacher, Fraktur, Greek, Hebrew, "Other Fonts" and "Not a Font".
Several of the data samples have multiple labels, thus this dataset is appropriate for testing multilabel classification methods.
This dataset is considered highly imbalanced.
Baseline results on ResNet with 50 and 18 layers \cite{He2016DeepRL}, VGG with 16 layers \cite{Simonyan2015VeryDC}, and DenseNet with 121 layers \cite{huang2018densely} were presented, and the \ac{mIoU} metric was reported.
This dataset was further used on the Historical Document Classification Competition \cite{10.1007/978-3-030-86337-1_41} that was part of the \nth{16} \ac{ICDAR} 2021.
Further information about this competition is presented in \ref{sssec:his_doc_classification_icdar_2021}.

\subsubsection{ICDAR 2021 Competition on Historical Document Classification (HDC 2021) }
\label{sssec:his_doc_classification_icdar_2021}

The competition on Historical Document Classification \cite{10.1007/978-3-030-86337-1_41}, which was hosted at ICDAR 2021 included three tasks for single or multilabel document classification: font/script, location, and date.
For the first task of font/script classification, the organizers provided the multiple font dataset presented in Section \ref{sssec:fonts} as the training dataset for the font classification task and the ICDAR17 (\ref{sssec:ICDAR17CLAMM}) and ICDAR16 (\ref{sssec:ICFHR16CLAMM}) CLaMM datasets for the script classification task.
Two new test sets were introduced for each task. 
For the date classification tasks, a new training and test set containing 11,294 and 2,516 images, respectively, was introduced. 
The ICDAR17 CLaMM dataset was suggested as an additional training set. 
New training, validation, and test sets of French manuscript images with 13 location labels were introduced for the location task. 
The competition results were evaluated using the top-1 accuracy on the test sets for the font/script and location tasks and the mean average error (MAE) for the date classification task. 
For all tasks, the winning team used \ac{CNN} operating either on non overlapping patches of four different scales or text lines acquired through segmentation \cite{10.1007/978-3-031-06555-2_11}. 
For the date classification task, instead of a cross-entropy loss, which was used in the other tasks, an interval regression loss was used that treated the task as a regression problem.

\subsubsection{The BIR Database}
\label{sssec:bir_database}

The Bold-Italic-Regular (BIR) database \cite{10.1145/3476887.3476913} consists of printed historical document pages with word bounding boxes and three font classes (bold, italic, and regular) meant for word detection and font style classification.
The BIR database includes 285 scanned pages from various catalogues from the \nth{19} and \nth{20} centuries written in French, Latin, or other languages.
Baseline results using 50\% training, 25\% validation, and 25\% test random splits (TVT), and a 5-fold cross-validation (CV5) were presented.
For word detection, YOLOv5m was utilized \cite{glenn_jocher_2021_4418161}, and for style classification, MobileNetV2 \cite{sandler2019mobilenetv2} and Xception \cite{chollet2017xception} were utilized.
All models were evaluated according to the F1 score.
The MobileNetV2 style classification results were also compared with the results of a human expert on 1K randomly chosen words.
The results showed a similar performance between the model and the human expert.


\subsection{Document Structure Datasets}
\label{ssec:structure}

The structure of a document refers to the organization of every element within it. 
After the detection of the different objects present in a document the aim is to classify these objects.
A document may consist of several blocks of text such as title, paragraphs, main body text, text lines, graphics, tables and more. 
The organization of these elements in specific places of the document constitutes the layout of the document. Detecting and extracting information is essential to get the geometry presented in a document.
Many datasets are publicly available to promote research that deals with the structure of documents.
In this subsection, we present the datasets aimed for tasks related to the geometric structure of documents such as layout analysis (detection and segmentation), baseline and text line detection, table detection, and graphic recognition.
We also include binarization, although it could be considered as a pre-processing step.
The number of summarized studies in this subsection is 24.

\subsubsection{Persian handwritten text dataset
(PHTD) }
\label{sssec:phtd}

PHTD \cite{6121553} is a 140-page dataset of handwritten documents in the Persian language.
The dataset includes 1,787 text-lines and 27,073 words for text recognition and word and line segmentation tasks.
For the former task, a unicode text file is included for every page, and for the latter, a pixel-labeled file is provided as the ground truth.
Two algorithms were utilized for the task of text-line segmentation.
The Potential Piece-wise Separation Line (PPSL) method \cite{10.1007/s10044-011-0226-x} obtained an 89.43\% segmentation accuracy, while the other method proposed by Alaei et al. \cite{Alaei2011ANS}, outperformed PPSL with an accuracy of 94\%.
In both methods, each document was split vertically into stripes.

\subsubsection{Persian, Bangla, Oriya and Kannada (PBOK) }
\label{sssec:pbok}

The PBOK dataset \cite{Alaei2012DatasetAG} includes images of 707 handwritten pages in four languages by numerous writers.
More specifically, this dataset provides 140 pages written in Persian, 228 pages written in Kannada, 199 pages written in Bangla, and 140 pages written in Oriya pages.
This dataset contains both pixel- and content-level annotations.
PBOK is considered quite complex, as it contains handwriting in both directions (left to right and right to left) and overlapping text.
The authors conducted line segmentation experiments for each language part and for the whole dataset using two algorithms, the Potential Piecewise Separation Line (PPSL) \cite{10.1007/s10044-011-0226-x} and the method proposed by Alaei et al. \cite{Alaei2011ANS}.
The Alaei et al. method achieved a \ac{DR} of 91.33\%, a \ac{RA} of 90.41\%, and an overall segmentation result (TLDM) of 90.87\%, and outperformed PPSL, which achieved values of 88.07\%, 86.69\%, and 87.38\%, respectively.

\subsubsection{IMPACT}
\label{sssec:impact}

IMPACT \cite{10.1145/2501115.2501130} is a large-scale dataset of over 600K images derived from different European libraries.
The provided PAGE XML files \cite{5597587} contain the layout, reading order, and text transcription annotations for over 45K samples.
This collection also offers metadata information, including bibliographic information, digitization information, physical characteristics, copyright information, administrative information, and comments.
The dataset includes documents written in 18 different languages (Bulgarian, Catalan, Czech, Dutch, English, French, German, Greek, Hebrew, Latin, Norwegian, Old Church Slavonic, Polish, Portuguese, Russian, Slovenian, and Spanish) and 10 scripts (Bohoričica, Cyrillic, French, Gaj, Greek, Hebrew, Latin, Latin/Gothic, Old Cyrillic, and Serif).
A web interface provides access to the image samples and annotations, giving various options for the users to browse and search.
This paper did not present any benchmark results for this dataset.

\subsubsection{Europeana Newspapers Project (ENP)}
\label{sssec:enp}

The Europeana Newspapers Project (ENP) dataset \cite{7333898} consists of European cultural heritage documents of 13 different languages from 12 European libraries published in newspapers from the \nth{17} to the \nth{20} century. 
All page images in the dataset are either 300 or 400 dpi, and there is a broad distribution of grayscale, bitonal, and color pages. 
The ground truth contains region outlines and their labels (text lines, text regions, tables, images/graphics, and blocks/zones), Unicode text transcriptions, and reading order.
In addition to conventional downloading, this dataset can be accessed through a web interface with several services such as document and attachment retrieval and a search function to determine if a document exists in the database.
A performance evaluation for \ac{OCR} and layout analysis was conducted using a commercial system (ABBYY FineReader 11) and an OCR system (Tesseract 3.03)\footnote{\url{https://github.com/tesseract-ocr/tesseract}\label{tesseract}}.

\subsubsection{GRPOLY-DB}
\label{sssec:GRPOLY-DB}

The Greek polytonic database (GRPOLY-DB) \cite{7333841} contains images of printed and handwritten documents from different sources distributed across four subsets: GRPOLY-DB-Handwritten, GRPOLY-DB-MachinePrinted-A, GRPOLY-DB-MachinePrinted-B, and GRPOLY-DB-MachinePrinted-C. 
The documents were written or printed in the old polytonic system from 1838 to 1977.
The overall dataset includes 399 pages, 15,084 text lines, 102,596 words, and 171,511 characters with ground truth.
The provided annotations include text and word-level segmentation information and transcriptions for text recognition and isolated character recognition.
Layout-related and content-related experimental results were shown on the dataset using various methods. 
For text-line segmentation, a shredding-based system \cite{5277573}, which achieved a value of 94.58\% for the average F-measure on the four subsets, outperformed a Hough transform method. 
For word segmentation, sequential clustering \cite{953781} and Gaussian mixture-based methods \cite{10.1016/j.patcog.2008.12.016} were applied. 
The former method achieved the highest average FM of 94.85\%.
The GRPOLY-DB-MachinePrinted-B was the only subset used for isolated character recognition in two scenarios, one scenario with all character instances and another with 30 random samples per class.
Thus, there were 143,051 and 3,750 characters per scenario, respectively.
Two algorithms were evaluated for both scenarios, HoG features \cite{1467360} with an SVM classifier and adaptive window features \cite{6065492} with a k-NN classifier.
For both systems, the first scenario obtained the highest \ac{RA}. 
Between the two methods, HoG features with the SVM classifier obtained the highest metric values in both scenarios.
In addition, \ac{OCR} experiments were performed at the character and word levels using Tesseract\footref{tesseract} and ABBY FineReader, with the latter performing the best. Finally, the \ac{mAP} was presented for query-by-example word spotting, where profile features with dynamic time warping (DTW) \cite{Rath2006WordSF} for feature vector comparison obtained the best results on the whole dataset.

\subsubsection{DIVA-HisDB}
\label{sssec:diva-hisdb}

DIVA-HisDB \cite{7814109} is a database that contains 150 images derived from three medieval manuscripts from the \nth{11} and \nth{14} centuries with complex layouts.
This database provides 20 training, 10 validation, 10 test, and 10 left out images for every manuscript annotated at pixel-level using the PAGE format for the following classes: main text body, decorations, and comments.
Benchmark results applying convolutional autoencoders (N-light-N) \cite{7814107} showed an average accuracy of approximately 95\% for pixel classification and the accuracy for every category. 
Additionally, the challenges of the dataset in terms of text-line segmentation using the Seam Carving \cite{6981106} and OCRopus \cite{Breuel2008TheOO} methods were demonstrated.
The HisDoc-Layout-Comp competition of ICDAR 2017 \cite{8270154} used the DIVA-HisDB dataset to evaluate systems on layout analysis (Task 1), baseline detection (Task 2), and text-line segmentation (Task 3).
For layout analysis, the best overall performance in terms of \ac{mIoU} was achieved by a \ac{FCN} that segmented every image at pixel-level.
The best performing system for Tasks 2 and 3 deployed Adaptive Run Length Smoothing (ARLS) \cite{Nikolaou2010SegmentationOH} to propose text lines and then processed them using the Seam Carving algorithm.

\subsubsection{HBA 1.0 }
\label{sssec:hba}

HBA 1.0 \cite{mehri:hal-01637826} is a collection of 11 books with 4,436 pages  of manuscripts and printed documents from the Gallica digital library from the \nth{13} to the \nth{19} century written in different scripts and languages. 
This dataset provides either ground truth images, where every foreground pixel has a different color according to the class it belongs to out of the six predefined classes, which include graphics, main text body, capitalized text, handwritten text, italic text, footnote text, or text files containing the label of every pixel. 
As a baseline, this paper presented the pixel-level classification accuracy (CA) of a texture-based layout segmentation method \cite{Mehri2015ATP} that, averaging over 4 books, was 75.9\%.
The ICDAR 2017 and ICDAR 2019 Competition on Historical Book Analysis \cite{8978192} introduced this dataset into two tasks: textual and graphical content discrimination at pixel-level and pixel-level annotation of textual content. 
The highest overall performance for both tasks was achieved by an \ac{FCN} that performed on $512\times512$ patches using a weighted cross entropy loss function.




\subsubsection{READ-BAD}
\label{sssec:read-bad}

The READ-BAD dataset \cite{8395221} contains 2,035 images of simple and complex documents with 132,123 baselines for baseline detection. 
The images were derived from 9 European archives written from 1470 to 1930, and as ground truth, the PAGE XML annotation format \cite{5597587} was used. 
The ICDAR 2017 Competition on Baseline Detection (cBAD) \cite{8270153} used the READ-BAD dataset for Track A, text-line segmentation on simple documents, and Track B, text-line location on complex documents with noise and various layout elements, providing only the page.
This dataset provides 216 training images for the simple layouts and 270 for the complex layouts.
An evaluation scheme for baseline detection was introduced using the R-value, P-value, and F-value. 
The R-value and the P-value have similarities to the recall and precision metrics, respectively, while the F-value is the harmonic mean of the two values. 
The best performing system for both tracks used a U-Net-based architecture (DMRZ submission) that extracted baselines and text regions of interest. 
Moreover, a layout classification was performed according to the provided classes as a preprocessing step. 
Postprocessing further improved the predicted baselines through detection error pruning and baseline fragment merging.

\subsubsection{Warped Arabic}
\label{sssec:warped_arabic}

In \cite{iet:/content/conferences/10.1049/cp.2018.1286}, a dataset of 200 historical Arabic document images from the 16th to the 19th century and four different libraries was introduced. 
The images were derived from books, newspapers, legal documents, and journals and contain a variety of layouts and states of degradation that help in facing challenges in baseline detection and text-line segmentation tasks. 
The PAGE XML ground truth \cite{5597587} contains text line-level information and metadata information according to bibliographic knowledge (author, title, date, location, document type, and page number), physical properties (language, script, font, and number of columns), and copyright data. 
Results were shown for text-line segmentation using four methods: Voronoi diagrams, a smearing method, a hybrid approach, and a projection profile-based method. 
For these methods, three warping percentages were utilized: 0\%, 25\%, and 50\%. 
The results suggested that the Voronoi diagrams achieved the highest success rate. Finally, the more warped the text lines are, the more challenging the task is. 
Thus, the performance decreases as the curvature increases.

\subsubsection{Oficio de Hipotecas de Girona (OHG)}
\label{sssec:OHG}

The OHG dataset\footnote{\url{https://zenodo.org/record/1322666\#.Ypi6Ty8RoUE}\label{OHG_link}} is a set of 596 pages of Spanish deeds written from a single writer on the \nth{18} century with a complex layout and six different layout regions that contain only text which are: page number, notarial typology, paragraph of text that begins next to a notarial typology, paragraph that begins on a previous page, marginal note, and marginal note added a posteriori to the document.
The dataset includes more than 23,700 lines and a 2,400-word vocabulary and the PAGE XML ground truth files for both layout analysis and handwritten recognition.

\subsubsection{Pinkas}
\label{sssec:pinkas}

The Pinkas dataset \cite{8978129} is a collection of 30 handwritten medieval Hebrew pages intended for page, line, and word segmentation tasks. 
The training and test sets contain 10,397 and 3,278 images, respectively.
These images are derived from records of European Jewish communities from 1500 to 1800.
The \ac{mAP} of different word spotting methods, including \ac{CNN} variations, was utilized in this study.
Three methods were set to provide a baseline for the dataset.
Siamese \ac{CNN} \cite{Bromley1993SignatureVU} and PHOCNet \cite{Sudholt2016PHOCNetAD} were compared as segmentation methods and an SVM with HOG descriptors \cite{Almazn2012EfficientEW} was used as a segmentation-free method.
The Siamese CNN, which achieved a \ac{mAP} of 61.5\%, outperformed the other methods. PHOCNet achieved a \ac{mAP} of 53.3\% using one-hot encoding and 56.6\% without it. 
The SVM did not perform well, as it achieved a \ac{mAP} of 1.5\%.

\subsubsection{BADAM}
\label{sssec:badam}

BADAM \cite{Kiessling2019BADAMAP} is a baseline detection dataset containing 320 training and 80 test pages of Arabic and Persian handwritten text.
These pages are derived from different sources and contain medical tracts and religious, legal, poetic, and other various content. 
This dataset provides 107,700 lines in two formats of annotations: PAGE XML \cite{5597587} and bitmasks.
The evaluation scheme proposed in \cite{8395221} was utilized for a convolutional baseline layout analysis (C-BLLA) system that classified every baseline pixel using a U-Net model \cite{10.1007/978-3-319-24574-4_28} and then extracted the baseline. 
The evaluation scheme from READ-BAD (Section \ref{sssec:read-bad}) was used and the P-value, R-value, and F-value metrics were presented for the model on the BADAM and Latin cBAD \cite{8270153} test sets. 
The results suggested that baseline detection in Arabic script is more challenging than in Latin script.

\subsubsection{HORAE}
\label{sssec:horae}

The HORAE dataset \cite{10.1145/3352631.3352633} contains 557 images derived from books of hours and their corresponding layout and text-related annotations.
These images originate from the full HORAE corpus, which consists of 500 manuscripts and 107,227 pages.
To create the final HORAE dataset with the annotated pages, a selection pipeline was used that initially classified pages into the following classes: binding, white page, calendar, miniature, miniature-and-text, text-with-miniature, and full-page text.
They excluded the ones that were binding or white pages and kept two images per class.
Then, the filtered pages were clustered from the initial step to keep one from every class and detect the ones considered outliers because of their rare layout. 
For the final 557 image set, the centroids from the most frequent layouts and the strongest outliers were annotated.
A PAGE XML \cite{5597587} file accompanies every image of the final set with annotations for page, miniature, border elements, initials, and other decorations found in the text body, such as line filler, music notations, and ornaments. 
Benchmark results were presented for line detection and layout analysis using the dhSegment segmentation neural network \cite{Oliveira2018dhSegmentAG} and evaluated according to the IoU with different thresholds and postprocessing.

\subsubsection{ICDAR 2019 Competition on Table Detection and Recognition (cTDaR) }
\label{sssec:ctdar}

The cTDaR competition of 2019 \cite{8978120} held two tracks, Track A for table region detection and Track B for table recognition, and two datasets, modern and historical.
The historical document dataset includes civil records containing various handwritten tables sourced from 23 different institutions.
For the table detection track, the dataset provides 600 training and 199 test images.
The table recognition track provides 600 training and 150 test samples for two subtracks: B1, which provides the tables regions, and B2, which does not provide any a priori knowledge. 
Hence, there is a need for both region and structure detection for B2. 
The results from 11 teams for track A and two teams for track B were compared. 
The winning team for track A achieved a \ac{WA} F1 score of 0.94 for the historical documents by using a classifier to categorize modern and archive samples and Faster-RCNN \cite{NIPS2015_14bfa6bb} for table detection. 
Then, they merged the overlapping regions that exceeded a given threshold value.
For the second track, the best submission achieved a \ac{WA} F1 of 0.48 for B1 and 0.47 for B2. 
An \ac{FCN} was used to obtain the tables' guiding lines and junction points for broken line repair.
Then, the cells were extracted through Connected Component Analysis and the row and column range were handled through a neighbor graph.

\subsubsection{ICDAR 2019 Competition on Digitised Magazine Article Segmentation (DMAS2019)}
\label{sssec:DMAS2019}

This competition\footnote{\url{https://www.primaresearch.org/DMAS2019/}\label{dmas2019}} aimed to recognize and classify parts of articles present in digitized historical magazines.
The competition provided 50-100 annotated images from magazines from 1800 -1938 taken by the National Library of the Netherlands and their layout and \ac{OCR} ground truth.
The annotations include cover, table of contents, content, and index as page classes and article, illustration with caption, advertisement, index, and colophon as article classes.
The competition does not seem to provide information about the submitted systems.

\subsubsection{ICDAR 2019 Competition on Document Image Binarization (DIBCO 2019)}
\label{sssec:DIBCO_2019}

The latest of the DIBCO competition series of 2019 \cite{8978205} aimed in evaluating various systems for the task of image binarization.
The series of this competition initiated in 2009 \cite{5277767} and had several rounds for printed and handwritten document images \cite{8270159, 6628857, 6065249}.
The 2019 competition included two categories, CATEGORY I, that provided 10 historical handwritten and printed test images of the \nth{19} century, and CATEGORY II, that provided 10 test images derived from papyri of various places in Egypt.
For CATEGORY I, the best performing method used noise reduction and then an ensemble of three clustering algorithms (Fuzzy C-Means, K-Medoids and K-Means++) for the step of grouping the foreground and background of the input images.
The best performing system for CATEGORY II, used the neural network architecture LadderNet \cite{Zhuang2018LadderNetMN} on $48\times48$ image patches.
All systems were evaluated using \ac{FM}, pseudo-\ac{FM} ($F_{ps}$), PSNR, and Distance Reciprocal Distortion Metric (DRD).

\subsubsection{ABP \& NAF} 
\label{sssec:ABP_NAF}

The work presented in \cite{8978117} introduces a method for layout and page sub-division that groups text-lines into semantic objects.
In order to evaluate the proposed method, they use the ABP dataset (ABP small) \cite{8395184} and they further introduce and extension of it, ABP large, and the National Archive Finland (NAF) dataset.
ABP small, ABP large, and NAF contain 180, 1,098, and 488 pages, respectively, and were used for table rows, columns, and cells segmentation, where the F1 measure is reported, and shows the most promising results in the cell partition.

\subsubsection{Finnish Court Records-sub500 (FCR)} 
\label{sssec:FCR}

The FCR dataset \cite{Quirs2020FinnishCR} includes 500 pages from the Renovated District Court
Records of Finland from the \nth{19} century.
The images are both single- or double-page which makes the dataset quite complex and the corresponding ground truth contains annotations on baseline- and layout-level.
The layout regions included are: page number, marginalia, paragraph, paragraph2, table, and table2.
The ground truth further includes the line-level transcriptions in the Swedish language.

\subsubsection{IlluHisDoc} 
\label{sssec:illuhisdoc}

In \cite{monnier2020docExtractor}, a test set of Gallica images named IlluHisDoc was presented for segmentation generalizability purposes.
This set was split into four types of documents: (a) printed documents with drawings, photos, ornaments, and paintings; and manuscripts that contain (b) scientific graphs, (c) illuminations, and (d) drawings.
Moreover, a segmentation method based on a ResNet-18 \cite{He2016DeepRL} backbone encoder-decoder architecture was proposed.
The performance of this model was compared with the performance of Tesseract4\footref{tesseract} and Mask-RCNN \cite{he2017mask} using pre-training either on the synthetic dataset PubLayNet \cite{zhong2019publaynet} or on SynDoc, a 10K image synthetic corpus created for this study.
The proposed method pretrained on SynDoc outperformed the other methods according to the \ac{mIoU} results.

\subsubsection{Newspaper Navigator}
\label{sssec:newspaper_navigator}

The Newspaper Navigator \cite{10.1145/3340531.3412767} is a dataset extracted from the Chronicling America historical newspaper collection.
This dataset was created by employing an object detection pipeline over the 16.3 million collected pages that extracts visual and headline content. 
The dataset provides 3,559 images with 48,409 COCO format annotations \cite{Lin2014MicrosoftCC} for easy detection across 7 classes: headline, photograph, illustration, comic, map, editorial cartoon, and advertisement.
The textual content of the predicted bounding boxes is further rendered for \ac{OCR} purposes and ResNet-18 and ResNet-50 embeddings \cite{He2016DeepRL} for the different visual category crops. 
Additional metadata in CSV format contain information such as file path, image URL, page URL, publication date, page sequence number, edition sequence number, batch name, LCCN, bounding box coordinates, prediction score, \ac{OCR}, place of publication, geographic coverage, newspaper name, and newspaper publisher.
A fine-tuned Faster-RCNN model \cite{NIPS2015_14bfa6bb} with an R50-FPN backbone achieved a \ac{mAP} of 63.4\% on the validation set.
These results also included the AP for every class.
The authors further chose 500 pages randomly from 1850-1875 and 1875-1900, treating them as test sets, and presented the \ac{mAP} on the most frequently appearing classes: the headline, the advertisement, the illustration, and the one class (all visual content into 1 class).
These results were slightly worse than the results on the validation set, especially in the case of the 1850-1875 test set.

\subsubsection{HJDataset}
\label{sssec:hjdataset}

HJDataset \cite{Shen_2020_CVPR_Workshops} was introduced in the Text and Documents in the Deep Learning Era Workshop hosted by CVPR 2020. 
The HJDataset contains 2,271 pages from Japanese biography scans for layout analysis with COCO annotations \cite{Lin2014MicrosoftCC}, derived by a semirule-based method. 
Furthermore, the ground truth includes reading order and dependency structure information. 
Benchmark results of experiments with popular object detection models such as Faster-RCNN \cite{NIPS2015_14bfa6bb}, Mask-RCNN \cite{he2017mask}, and Retinanet \cite{lin2017focal} provided by Detectron2 were shown \cite{wu2019detectron2}. Moreover, few-shot and zero-shot learning results using COCO weights were presented.

\subsubsection{GloSAT}
\label{sssec:GloSAT}

GloSAT \cite{10.1145/3476887.3476890} is a table structure recognition dataset of 500 archival images, printed, handwritten or mixed, of meteorological records.
There are two types of ground truth in the dataset: individual cell and coarse segmentation cell annotations.
In addition to the conventional XML cTDaR19 format annotations, the dataset provides the widely used Pascal Visual Object Classes (VOC) format \cite{Everingham2009ThePV} and extends these formats with cell information such as headers, page type, and table style.
A benchmark evaluation of GloSAT (individual cell and coarse segmentation cell separately), cTDaR19, and their combination (+cTDaR19) using CascadeTabNet \cite{Prasad2020CascadeTabNetAA} and CascadeTabNet with additional postprocessing proposed by the authors was presented.
This postprocessing step uses a 1-D DBSCAN clustering algorithm \cite{10.1145/3068335} to infer vertical and horizontal lines of a table, assuming that only a subset of cells is needed to place the rest for a rectangular table.
The results on the \ac{WA} F1 score showed that postprocessing helps the performance in all experimental cases.

\subsubsection{BiblIA}
\label{sssec:BiblIA}

BiblIA \cite{10.1145/3476887.3476896} is a publicly available dataset of Medieval manuscripts written in Hebrew and Aramaic that contains 6 different scripts: Ashkenazi, Byzantine, Italian, Oriental, Sephardi, Yemenite.
BiblIA contains more than 200 images with their corresponding annotations on baseline- and transcription-level, both focusing on the main text.
Furthermore, a segmentation and recognition model based on kraken OCR is used for evaluation.
The work presents \ac{acc} results on specific scripts (Ashkenazi, Italian, and Sephardi) and all scripts as well as training information.
Further experimental results show the \ac{CER} and \ac{WER} on images not included in the test set.

\subsubsection{HisClima}
\label{sssec:hisclima}

HisClima \cite{9412210} is a database of handwritten weather ship log book pages from 1880 to 1881 that contains both layout annotations of blocks, columns, rows, and lines and transcription annotations with relevant information such as number of cells in the tables.
The dataset comprises 208 pages with tables and 211 pages with descriptive text.
Baseline experiments are performed for text recognition, line segmentation, and information extraction.
A CRNN  with CTC loss performing on line images was used for the recognition task with and without a language model (LM) and evaluated according to the \ac{WER} and \ac{CER}.
The neural network architecture presented in \cite{Quirs2018MultiTaskHD}, that performs geometric and logical layout analysis, was used for the task of line segmentation.
Finally, for the information extraction for cell position and line geometry an information retrieval on tables method without segmentation based in \cite{8563224} was used.
The two latter tasks were evaluated according to precision, recall, and F1 scores.



\subsection{Content Analysis Datasets}
\label{ssec:content}

Content is a fundamental part of a document, as it contains the semantics that make a document perceivable to humans.
After categorizing a document and detecting its geometric structure, document understanding follows.
The mapping of the layout structure into a logical structure is the understanding of the document that is followed by the content analysis.
This section includes 35 studies related to \ac{OCR} tasks, whether they target isolated characters, words, lines, digits, or whole document transcriptions, writer identification, reading order, and any type of retrieval that could refer to writer, image, or word spotting.
Two datasets related to handwritten music recognition were further detected \cite{8269947, vorau}, but are not included in detail in the scope of this work.

\subsubsection{GERMANA Database}
\label{sssec:germana}

GERMANA \cite{5277691} is a database of 764 scanned pages from an 1,891 manuscript written in Spanish.
The pages contain 21K text lines and 217K words.
Catalan, French, Latin, German, and Italian may also appear in some parts of the text.
The ground truth comprises bounding box annotations for the text blocks, straight baselines for every text line, and line-by-line transcriptions.
Although the database annotations contain both layout and text information, the baseline experiments were limited to the task of handwriting recognition. 
The transcription \ac{WER} per block for handwriting recognition experiments were presented using a system that combines Hidden Markov Models for text recognition and n-grams for language modeling \cite{4377054}.
The results contained the first 180 pages of the database separated into blocks of 20 pages and were presented by adding each block consecutively.
A 37\% \ac{WER} was achieved for the last two blocks, while the error is higher in the first blocks, where more out-of-vocabulary words were presented.

\subsubsection{RODRIGO Database }
\label{sssec:rodrigo}

The RODRIGO database \cite{serrano-etal-2010-rodrigo} contains data derived from a manuscript written in 1545 in old Castilian by one writer.
The database follows a similar strategy as the GERMANA database in creation, ground truth, and experimental baseline.
It includes 853-page images of one column text blocks, where each block is annotated with a bounding rectangle, and then each line within it with the corresponding baseline.
The annotations further include transcriptions for every line, which results in a total of 20,357 text lines and 231K words as ground truth.
Baseline results were provided for the task of handwriting recognition using the same model and processes as in Section \ref{sssec:germana}, by using 20 blocks of 1K lines, and a \ac{WER} of 36.5\% was achieved on the last block.

\subsubsection{IAM-HistDB }
\label{sssec:iam-histdb}

IAM-HistDB \cite{10.1145/1815330.1815331} is a highly used database of handwritten historical manuscript images that contains three datasets: Saint Gall, Parzival, and George Washington (GW).
We present these datasets in the following paragraphs.


The Saint Gall database \cite{10.1145/2037342.2037348} is a set of 60 page images and 1,410 binarized and normalized text-line images of manuscripts written in the \nth{9} century in Latin language and Carolingian script by one writer.
The text edition for every page image was provided.
The pages are composed of 11,597 words, 4,890 word labels, 5,436 word spellings, and 49 letters.
The ground truth includes the line-level text transcriptions and the word and line locations.
An evaluation of a transcription alignment system based on HMM is proposed in the paper and compared with three more reference systems.

The Parzival database \cite{FISCHER2012934} provides handwritten documents from the \nth{13} century originating from three writers and was written in Old German and Gothic script.
It contains 47 pages, 4,477 text lines, 23,478 words, 4,934 word categories, and 93 letters.
Similar to St. Gall, the line and word images are binarized and normalized.
As ground truth, Parzival includes line- and word-level transcriptions.
The work presented in \cite{5306020} used a HMM-based system similar to \cite{10.5555/505741.505745} and the BLSTM introduced in \cite{Graves2009855} recognizer for automatic handwriting recognition on the Parzival dataset and achieved a word \ac{acc} of 88.69\% and 93.32\%.
Furthermore, in \cite{FISCHER2012934}, a lexicon-free word spotting method based on character HMMs was proposed and evaluated on the Parzival and GW datasets.

The GW database \cite{FISCHER2012934} is comprised of \nth{18} century documents from the George Washington Papers and contains 656 text and 4,894 word images, binarized and normalized, along with their transcription annotations.
The pages are written in English by two writers in longhand script.
The dataset statistics also include 1,471 word classes and 82 letters.
This dataset is widely used to evaluate word spotting algorithms. 
\cite{5871643} used this database and compared a proposed word spotting method that used a BLSTM and a modified CTC algorithm with a HMM \cite{10.1016/j.patcog.2009.02.005} and a DTW \cite{Rath2006WordSF} method.
The paper presented average precision results using GW and Parzival datasets and the proposed method achieved 0.84 on the GW and 0.94 average precision on Parzival.

\subsubsection{ESPOSALLES}
\label{sssec:esposalles}

The ESPOSALLES database \cite{ROMERO20131658} is a collection of ancient marriage license documents separated into the LICENSES and the INDEX subsets.
A single-writer book written in old Catalan is the main content of the LICENSES set, and is comprised of 173 pages and 1,747 licenses.
For every page, the subset includes the main text block bounding box, the text line within the text block coordinates, the license label, and the transcription for every line, word, and character in the main block.
The INDEX subset, which includes 29 pages of the initial indexes of two volumes by a single writer created between 1491 and 1495.
Similar to the LICENSES part, INDEX contains text and line layout as well as transcription annotations.
Both subsets provide dataset splits for cross-validation.
Finally, baseline results using Hidden Markov Models (HMM) \cite{Toselli2004IntegratedHR} and a BLSTM \cite{Graves2009855} with two feature sets, PRHLT \cite{Toselli2004IntegratedHR} and IAM \cite{10.5555/505741.505745}, showed the efficiency of neural networks with larger datasets for handwriting recognition.
The database was further used in the ICDAR17 Competition on Information Extraction in Historical Handwritten Records \cite{8270158}.
The aim there was to detect and assign name entities to semantic categories (name, surname, occupation, etc.) for two Tracks: Basic and Complete, which also contains the person (husband, wife, etc.).
The team that obtained the highest average score with word-level segmentation used a ResNet-based unigram system for character recognition and named entity recognition, while with line-level segmentation, the best method used a RNN-LSTM with \ac{CTC}.

\subsubsection{BH2M }
\label{sssec:bh2m}

The Barcelona Historical Handwritten Marriages Database or BH2M \cite{6976764} consists of 174 handwritten marriage record pages, where 100 pages are meant for training, 34 for validation, and 40 for testing.
The included pages were written in Old Catalan by a single writer between 1617 and 1619 and preserved in the Barcelona central archives.
The database provides the ground truth for layout analysis, text transcription, and semantic analysis.
XML annotation files are organized hierarchically into text blocks, segmented lines, and text words for layout analysis.
The additional word transcriptions and semantics about the license, appearance order, date, and information about the wife and husband, may enable handwritten text recognition, word spotting, information extraction and understanding and context-aware algorithms.
Moreover, baseline results of line segmentation \cite{Mota2014AGA} using the \ac{DR}, \ac{RA}, and \ac{FM} metrics, and segmentation-free \cite{Vinciarelli2002OfflineCW} and -based \cite{Almazn2012EfficientEW} word spotting algorithms using the \ac{mAP} were presented.

\subsubsection{HADARA80P}
\label{sssec:hadara80p}

The HADARA80P dataset \cite{6980990} contains 80 handwritten Arabic pages originating from a single-author book about the taaum disease and its connections to religion.
The XML ground truth files provide the pages, text block, word coordinates, and transcription for every word.
In some cases, tag values accompany the words.
The total number of labeled words is 16,720.
Experiments using a publicly available word spotting application\footnote{\url{http://www.corenum.com/products/ulysse/}\label{ulysse}} are presented and an extension of the methods used in the application \cite{Leydier2007TextSF, Leydier2009TowardsAO}, the HADARA word spotter, is proposed. 
The original methods work by locating the zones of interest through gradients, while the proposed method employs curvature according to a threshold instead.
The resulting \ac{mAP} based on the precision measures $p_{IR}$ and $\overline{\gamma_{LA}}$, presented in \cite{6628824}, showed that the proposed system outperformed the already existing application on the HADARA80P and the George Washington datasets.

\subsubsection{DocExplore}
\label{sssec:docexplore}

DocExplore \cite{10.1117/1.JEI.26.1.011010} is a pattern spotting dataset that contains 1.5K images with more than 1.4K queries.
The images originate from 6 different manuscripts written between the \nth{10} and \nth{16} centuries.
The annotation process ends with 1,464 labeled objects belonging to 35 graphical object categories, where one sample constitutes the query image and the remaining objects from every category avail as retrieval outcomes.
The dataset was proposed for two tasks: image retrieval and pattern localization.
As baseline for the latter task a system that consists of an offline, online, and post-processing step initially presented in \cite{EN2016149} is used.
In the offline step, the background is removed, a descriptor is used to find the object regions of interest, and finally a \ac{VLAD} is created.
Then, during the online step, a similarity distance calculation is performed between the extracted regions and the query image, then ranking was achieved through template matching.
The system achieved a 0.613 \ac{mAP} for retrieval and a 0.111 for localization, while further results on each category were shown.

\subsubsection{AMADI\_LontarSet}
\label{sssec:amadi_lontarset}

AMADI\_LontarSet \cite{7814058} is a collection of palm leaf manuscripts from Bali.
This dataset was a part of the ICFHR 2016 Competition on the Analysis of Handwritten Text in Images of Balinese Palm Leaf Manuscripts \cite{7814130}.
It contains binarized, word annotated, and isolated character annotated ground truth images used for the following challenges: Binarization of Palm Leaf Manuscript Images, Query-by-Example Word Spotting on Palm Leaf Manuscript Images, and Isolated Character Recognition of Balinese Script in Palm Leaf Manuscript Images, respectively.
For Challenge 1, binarization, the dataset includes 50 training images, 100 binarized images from two different sources (50 and 50) as ground truth, and 50 test images.
The team that outperformed the others used a pretrained \ac{FCN} on handwritten documents as presented in the work by Wolf et al. \cite{1048482} that was fine-tuned on the DIBCO \cite{5277767} and H-DIBCO \cite{8583809} images and then fine-tuned on the competition images.
The results were evaluated according to the F-Measure (FM), PSNR, and Negative Rate Metric (NRM) between the ground truth and the predicted binarized images.
For Challenge 2, word-spotting, a split of 130 train and 100 test images was provided along with 15,022 word annotated patches for training. 
Moreover, 36 word annotated patches were given as query test.
The goal was to use a query word image patch to retrieve similar word image patches in palm leaf manuscripts; however, there were no submissions for this challenge.
Finally, Challenge 3 aimed to recognize isolated Balinese characters distributed over 130 character classes.
The training set contains 11,710 labeled patch images, and the test set contains 7,673.
The method with the highest \ac{RA} (VMQDF) initially preprocessed the input images by resizing, binarizing using the OTSU method, and then defeating grayscale variation.
Then, synthetic samples were generated based on the preprocessed samples using the method in \cite{Shao2012FastSV}, gradient features were extracted for all images.
Finally, a classifier was trained on the new set that contained the original and 
generated images, while at the test phase, for every sample, 97 synthetic images were generated and treated according to the previously mentioned method.

\subsubsection{SleukRith}
\label{sssec:sleukrith}

SleukRith \cite{10.1145/3151509.3151510} is a dataset of 657 images from palm leaf manuscripts written in Khmer from 4 different sources.
This dataset includes annotations for isolated character recognition and word and line segmentation.
The most valuable aspect of this dataset is character recognition, which is the foundation for building the other two elements.
The individual character images were constructed by cutting patches for every character and removing the noise of near characters using inpainting.
For the rest of the annotations, the combination of the characters was used to determine the words and lines.
To evaluate the set for character recognition, the \ac{CER} of a \ac{CNN}, which was 6.04\%, was presented.
The dataset was further used in the ICFHR2018 Competition for Southeast Asian Palm Leaf Manuscripts presented in Section \ref{sssec:ICFHR18_ASIAN_PALM_LEAF}, which contained binarization, text-line segmentation, character recognition, and word transliteration tasks.
However, this dataset was not part of the binarization task.
The winning systems presented in the competition section also performed the best for this dataset alone.

\subsubsection{VML-HD}
\label{sssec:vml-hd}

VML-HD \cite{8067751} is a database of Arabic handwritten documents that includes 680 pages from 5 different books of different writers.
This database can be used for handwriting recognition and word-spotting.
The annotations of the dataset include the book and page number, the segment id, bounding box coordinates for 121,636 sub-words and 244,553 characters, length of subword, and Arabic and Latin symbol transcriptions in Hadara XML format.
Word spotting results using Radial Descriptor \cite{6981050} and Radial Descriptor Graph \cite{7814035} on a subset from every book and the 5 books combined were presented.
The Top1 - Top5 \ac{DR} of the Radial Descriptor Graph method showed better performance on the combined set than the Radial Descriptor.

\subsubsection{CFRAMUZ}
\label{sssec:cframuz}

The CFRAMUZ dataset \cite{Arvanitopoulos2017AHF} includes grayscale image pages from handwritten novels by Charles Ferdinand Ramuz in French between 1910 and 1946.
Text and XML annotation files contain the unique word ID, coordinates, width and height of word bounding boxes, word line number, word number in the current line, and word transcription for word spotting without segmentation purposes.
The following methods were evaluated according to Precision-Recall curves: Word Spotting and Recognition with Embedded Attributes (EAWS) \cite{6857995}, Efficient Exemplar Word Spotting (EEWS) \cite{Almazn2012EfficientEW}, Bag-of-Visual-Words Word Spotting (BoVWWS) \cite{Rusiol2011BrowsingHD}, and Fisher Kernels Word Spotting (FKWS) \cite{5277774}.
The \ac{mAP} of these algorithms were compared on the introduced datasets with the performance on the George Washington (GW) and the Lord Byron (LB) datasets.
Although this is a single-writer dataset, some variation in terms of writing style occured due to the year range.
Therefore, additional experiments using splits according to style were conducted.

\subsubsection{Lontar Sunda}
\label{sssec:SUNDANESE_PALM_LEAF}

The Lontar Sunda dataset \cite{8270066} is a collection of \nth{15} century Sundanese palm leaf manuscripts from Garut, West Java, and Indonesia.
This dataset includes 66 pages with corresponding binarization, word-level, and character-level annotations.
Lontar Sunda was one of the datasets used in the ICFHR 2018 Competition on Document Image Analysis Tasks for Southeast Asian Palm Leaf Manuscripts \cite{8583808}.
This competition hosted 4 challenges: A. Binarization, B. Text-line segmentation, C. Isolated character/glyph recognition, and D. Word transliteration.
As the original dataset paper did not include any benchmark results, the competition results are considered.
For Challenge A, systems were evaluated according to the \ac{FM}, Peak SNR (PSNR), and Negative Rate Metric (NRM).
The best performing system on the Sundanese data used Gaussian operators and a non-linear function to enhance the images. 
Then, the enhanced images were finally segmented with a threshold of 0.9.
In Challenge B, the system evaluation was made using the \ac{DR}, the \ac{RA}, and the \ac{FM}.
The system with the best values on the Sundanese collection, which was also the only submission for this task, used the binarized images from Challenge A and horizontal projection profile to perform line segmentation.
The character recognition challenge (C) was evaluated according to the recognition rate, and the highest value was obtained by a dense 100-layer \ac{CNN} architecture that classified similar characters.
Finally, in Challenge D, the best performing system achieved an 8.81\% \ac{CER} on the Sundanese set using a CNN-RNN encoder-decoder architecture with an attention mechanism.

\subsubsection{ICDAR 2017 Competition on Recognition of Early Indian Printed Documents (REID2017)}
\label{sssec:reid2017}

The REID 2017 Competition \cite{8270161} held at ICDAR 2017 includes 26 evaluation images written in Bengali from 1785-1909 and an example set of 5 images for training.
The competition originally held two tasks, the Bengali text recognition and the Quarterly Lists challenge (tabular recognition in English and Bengali); however, there were no submissions for the latter challenge.
The organizers of the competition provided the image annotations in PAGE XML format \cite{5597587} created using Aletheia \cite{6065274}.
These annotations included layout region polygons, metadata such as heading, paragraph, captions, footer, etc., and reading order information.
The Google Multilingual \ac{OCR} that uses the Google Cloud vision API \footnote{\url{https://cloud.google.com/vision/}\label{google}} achieved the highest flex \ac{ca} compared to the other submissions; however, the \ac{ca} of 75.4\% suggests that there is plenty of room for improvement.
The same system, with a success rate of 78.4\%, outperformed the other systems on the text region segmentation task.

\subsubsection{ICDAR2017 Competition on Historical Document Writer Identification (Historical-WI)}
\label{sssec:Historical-WI}

The Historical-WI competition \cite{8270156} focused on image retrieval based on writer identification.
The competition offered a set of 3,600 images of handwritten document pages ranging from the \nth{13} to the \nth{20} century for evaluation.
The test set originated from the Universitätsbibliothek Basel and included 720 different writers.
For training, 1,182 images in color and binary format from 394 writers were provided and were different from the writers in the test set.
The submitted systems were evaluated using the \ac{mAP} metric.
The system that achieved the highest \ac{mAP} used feature vectors derived from binarized samples and the concatenation of their oriented Basic Image Feature (BIFs) columns histograms \cite{7490135,NEWELL20142255}.

\subsubsection{Kuzushiji}
\label{sssec:K-MNIST}

The full Kuzushiji dataset \cite{Clanuwat2018DeepLF} consists of three parts: the Kuzushiji-MNIST, the Kuzushiji-49, and the Kuzushiji-Kanji.
The whole dataset is comprised of printed books from the \nth{18} century written in cursive Japanese or Kuzushiji.
The K-MNIST subset includes 70K 28$\times$28 grayscale images of 10 Kuzushiji character classes to resemble the MNIST and Fashion-MNIST datasets but is even more challenging.
Kuzushiji-49 contains 270,912 images of the same pixel resolution and mode as K-MNIST, including 49 character classes.
Finally, Kuzushiji-Kanji is a subset of 140,426 64$\times$64 grayscale images of 3,832 Kanji characters.
The two latter subsets are considered quite imbalanced. 
Benchmark results on the K-MNIST and Kuzushiji-49 were presented using a 4-nearest neighbor classifier, a 2-layer CNN, ResNet-18 \cite{10.1007/978-3-319-46493-0_38}, ResNet-18 with input mixup \cite{Zhang2018mixupBE}, and  ResNet-18 with manifold mixup regularizer \cite{verma2018manifold}. 
The performance of these models were compared using MNIST. All models had the highest test accuracy on the MNIST test set, followed by K-MNIST, and finally Kuzushiji-49. 
The best performing model for the K-MNIST and Kuzushiji-49 test sets was ResNet-18 with manifold mixup, while for MNIST, it was the simple ResNet-18 model.
Domain transfer was further explored from from Kuzushiji-Kanji to modern Kanji (stroke format).
Two Variational Autoencoders \cite{Kingma2014AutoEncodingVB, JimenezRezende2014StochasticBA} were used to create the old (Kuzushiji) and new (Modern) latent space embeddings.
Then, a Mixture Density Network \cite{Bishop94mixturedensity} predicted the probability of the new embedding given the old embedding. 
Finally, a Sketch-RNN \cite{Ha2018ANR} conditioned on the new latent space created modern Kanji stroke image versions of Kuzushiji.

\subsubsection{MHDID }
\label{sssec:mhdid}

MHDID is the Multi-distortion Historical Document Image Database \cite{8480372} for document quality assessment and distortion classification.
This dataset contains 335 images with four degradation types: wormholes, stains, reader annotations, and paper translucency.
The document images emanate from 130 books from the Qatar University Library and are written in Arabic.
Several users are supposed to compare pairs of images and select among three options.
These options are "The left image is better", "The images are similar", or "The right image is better".
With six outliers removed, the user interface results were normalized between 0, the lowest perceptual quality value, and 9, the highest.
Finally, the MOS value was computed for every image, which is the sum of the outcome pair comparisons divided by the number of pairs. 
A dataset analysis was further demonstrated in terms of color and spatial information to reveal the heterogeneity of the dataset. 
This database seems to be an outlier. 
Thus, we categorize it as retrieval in Table \ref{tab:structure_content_class} with a (\checkmark) since it is a database that compares pairs of images.

\subsubsection{Tripitaka Koreana in Han (TKH) and Multiple Tripitaka in Han (MTH)}
\label{sssec:tripitaka}

The work presented in \cite{Yang2018DenseAT} introduces two datasets for Chinese character detection and recognition, the Tripitaka Koreana in Han (TKH) and the Multiple Tripitaka in Han (MTH), created using the publicly available TKH images from the Tripitaka Koreana Institute.
For every character bounding box that the dataset includes, a character label was further provided.
The TKH consists of 1K pages, 23,471 lines, 323,491 characters, and 1,471 character classes, while the MTH contains 500 images, 17,178 lines, 197,886 characters, and 3,664 character classes.
The two datasets differ in terms of challenge, as the MTH dataset has a more complex character size uniformity, making the creation of bounding box annotations even harder.
A three-part pipeline called Recognition Guided Detector (RGD) was proposed. 
First, segmentation is performed for every line. 
A Recognition Guided Proposal Network (RGPN) generates context information, and finally, a detector uses that information to find the characters in every line.
Finally, several experiments were performed on the two datasets using the proposed system with and without a VGG-16 \cite{Simonyan2015VeryDC} backbone and its performance was compared with other well-known object detection frameworks using either the whole image or text lines as input.
This method seems to perform comparably to other methods using fewer parameters.

\subsubsection{ICFHR 2018 Competition on Recognition of Historical Arabic Scientific Manuscripts (RASM2018)}
\label{sssec:rasm2018}

The RASM 2018 competition \cite{8583806} was part of ICFHR 2018 and targeted the recognition of Arabic historical scientific manuscripts through three tasks: page segmentation, text-line detection, and \ac{OCR}. 
An example set of 15 single-column page images with PAGE XML ground truth format \cite{5597587} was provided for training and 85 for evaluation to handle these tasks.
As in similar competitions, the ground truth included polygon regions, text transcriptions, and metadata for each region, such as headings, paragraphs, captions, footers, and reading order.
For system evaluation, the competition used the success rate and errors for the page and text-line segmentation predictions and the \ac{ca} for the \ac{OCR}.
For page segmentation, the winning system used an \ac{FCN} applied on extracted patches.
The page layout predictions were further cropped for the text-line segmentation, which was performed at pixel-level using anisotropic Gaussian smoothing. 
The highest performance was achieved for the rest of the tasks by a Historical Arabic Handwritten/Typewritten \ac{OCR} system framework.
This system can handle various fonts and layouts, and in the case of the competition, an instance-based segmentation on extracted lines was performed.

\subsubsection{ICFHR 2018 Competition on Document Image Analysis Tasks for Southeast Asian Palm Leaf Manuscripts}
\label{sssec:ICFHR18_ASIAN_PALM_LEAF}

The Southeast Asian palm leaf manuscripts competition \cite{8583808} offered four tasks: binarization, text-line segmentation, isolated character/glyph recognition, and word transliteration. 
This competition included manuscripts in three languages: Balinese, Khmer, and Sundanese.
For Balinese, Khmer, and Sundanese language sets, the Amadi\_Lontarset dataset presented in Section \ref{sssec:amadi_lontarset}, the SleukRith dataset presented in Section \ref{sssec:sleukrith}, and the Lontar Sunda dataset presented in Section \ref{sssec:SUNDANESE_PALM_LEAF}, respectively, were utilized. 
 The results for each separate language set are presented in the corresponding dataset sections.
In this section, we will present the overall results of the competition.
The system that obtained the highest FM and PSNR values for binarization used Gaussian operators and a nonlinear function to enhance the images. 
Then, the enhanced images were finally segmented with a threshold of 0.9.
The same system achieved a value of 0.17 NRM that underperformed the best value by only 0.01.
In challenge B, text-line segmentation, the best and only \ac{DR}, \ac{RA}, and \ac{FM} values were achieved by using the binarized images from Challenge A and the horizontal projection profile to perform line segmentation.
Challenge C, character recognition, was evaluated according to the recognition rate, and the highest value was obtained by using a dense 100-layer \ac{CNN} architecture that classifies similar characters.
Finally, in Challenge D, the best performing system achieved a 5.62\% \ac{CER} on the mixed sets using a CNN-RNN encoder-decoder architecture with an attention mechanism.

\subsubsection{ARDIS} 
\label{sssec:ardis}

Arkiv Digital Sweden (ARDIS) \cite{kusetogullari2020ardis} is a handwritten digit dataset collection derived from historical church records. 
ARDIS contains four different datasets: \textit{Dataset I}, which contains 10K 4-digit string images that represent a year, \textit{Dataset II}, which contains single digits of classes 0-9, \textit{Dataset III}, which is the same as \textit{Dataset II} but is cleansed from noise, and \textit{Dataset IV}, which is the same as \textit{Dataset III} but is in grayscale and is similar to the highly used MNIST Database \cite{lecun1998mnist}.
\textit{Dataset II - IV} contain 7,600 digit images.
Several experiments using CNN, SVM, HOG+SVM, k-NN, random forest, and RNN classifiers were presented. 
The results of training using the MNIST and USPS \cite{291440} datasets and testing on ARDIS reveal the diversity of the ARDIS dataset as the highest \ac{RA} obtained reaches $58.80\%$, while training and testing on ARDIS gives a performance of $98.6\%$.
In all experimental cases, the best performance was achieved by the \ac{CNN} digit classifier.


\subsubsection{OBC306}
\label{sssec:obc306}

OBC306 \cite{8978032} is a dataset of 309,551 images for Oracle Bone character recognition distributed across 306 character classes.
This dataset consists of patch samples derived from different sources from full image publications of oracle bones.
For patch extraction, an oracle bone character list and dictionary were used to retrieve all characters and extract them from the source images to assign them to a class and a specific encoding.
The challenges faced in the dataset are the class imbalance and the numerous variants of each character.
The evaluation results of widely used \ac{CNN} architectures \cite{He2016DeepRL, Simonyan2015VeryDC, NIPS2012_c399862d, 7298594}, and a classical method of HOG descriptors with SVM \cite{1467360} were presented, and Inception-v4 \cite{7298594} achieved the best performance.
Although the dataset is hand carved, we characterize it as handwritten in Table \ref{tab:structure_content_class} for homogeneity reasons. 

\subsubsection{GRK-Papyri \& PapyRow}
\label{sssec:grk-papyri}

The GRK-Papyri dataset \cite{8978142} provides 50 handwritten Greek papyrus images from the \nth{6} century A.D for writer identification. It includes color and grayscale documents with 4-7 samples each from 10 different writers. 
This dataset provides a leave-one-out option that contains all images without any split and a train-test split with a balanced training set of 20 samples, keeping the rest with different numbers of samples per writer for testing.
The dataset has a high complexity, as the images are heavily degraded and low in quality, making pre- and post-processing inevitable.
Due to size limitations, the method used for evaluation was a Normalized Local Na\"{\i}ve Bayes Nearest-Neighbour (NBNN) classifier with FAST keypoints \cite{mohammed2017normalised}.
The authors suggested that this dataset can be further used for image processing tasks or line/word segmentation.
An extension of the GRK-Papyri is presented in \cite{10.1007/978-3-030-68787-8_16}, where enhancement techniques, such as background smoothing, line resizing, and image rotation,  were used to obtain images with less degradation.
In this extended version of the dataset, named PapyRow, 6,498 images were obtained using a row segmentation method and included with their corresponding XML ground truth.

\subsubsection{CASIA-AHCDB}
\label{sssec:casia-ahcdb}

CASIA-AHCDB \cite{8978010} is a database of 11,937 handwritten Chinese document pages.
For the task of character recognition, the database provides 2.2M handwritten characters belonging to 10,350 different classes.
The database distributes these elements across two different datasets: (a) the Complete Library in Four Sections (AHCDB - style1) and (b) the Ancient Buddhist Scriptures (AHCDB - style2).
Then, each dataset is split into a Basic Category Set (BC) for basic character recognition, an Enhanced Category Set (EC) for open-set character recognition, and a Reserved Category Set (RC) for other recognition purposes.
To benchmark the database, a \ac{CNN} \cite{Zhang2017OnlineAO} and a Convolutional Prototype Network (CPN) \cite{Yang2018RobustCW} were used and experiments were performed with only the \textit{Basic Category Set} and with the combination of the \textit{Basic} and \textit{Enhanced Category Sets} for every dataset. 
Moreover, the transfer of information from the style1 to the style2 dataset with direct train-test and fine-tuning was attempted, which performed best among the two methods.

\subsubsection{Amharic Text Image Recognition Database }
\label{sssec:amharic}

The Amharic database \cite{8977980} presents a collection of 40,929 printed images with Amharic script text lines originating from pages of different documents written in Amharic and 296,403 synthetic images created using OCRopus \cite{Breuel2008TheOO}.
The generated synthetic images include the Power Geez and the Visual Geez fonts.
A Bidirectional \ac{LSTM} followed by a softmax function that produced 281 probability values, which is the number of unique characters in the database, and a \ac{CTC} output layer were proposed for text-line recognition.
This method achieved an 8.54\% \ac{CER} for the printed Power Geez documents, 2.28\% \ac{CER} for the Visual Geez synthetic images and 4.24\% \ac{CER} for the Power Geez synthetic images.

\subsubsection{ICDAR 2019 Historical Document Reading Challenge on Large Structured Chinese Family Records (ICDAR19 HDRC-Chinese)}
\label{sssec:chinese_reading_challenge}

This competition \cite{8977999} presented a database of approximately 10K historical Chinese family record pages to evaluate systems for the tasks of (1) text recognition on extracted lines, (2) pixel-level layout analysis, and (3) text-line detection and recognition.
More specifically, the training set includes 11,715 pages derived from 37 different books along with their PAGE XML \cite{5597587} and pixel-wise annotations, while the test set includes 1,135 images from 12 books.
To evaluate the submitted systems, Task 1 uses the edit distance (editDistance), Task 2 uses the \ac{mIoU}, and Task 3 uses the total counted errors (totalErrors) of the output XML file.
The team that achieved the best results for all tasks used a Convolutional Recurrent Neural Network (CRNN) \cite{Shi2017AnET} to recognize Chinese text, a Cascade R-CNN \cite{Cai2018CascadeRD} to detect text lines, and a U-Net-shaped network for the pixel-wise classification.
For Task 2, the system that outperformed the others achieved a 99.96\% IoU for the background class and 99.24\% for the text class.

\subsubsection{ICDAR 2019 Competition on Recognition of Early Indian Printed Documents (REID2019) }
\label{sssec:reid2019}

The REID2019 competition \cite{8978191} is an extension of the previously mentioned REID2017 competition (Section \ref{sssec:reid2017}).
This competition provided 25 labeled images with the same annotation format and content as the previous competition and a balanced test set of 56 images written in English and Bengali.
This competition hosted two tasks, layout analysis and text recognition, but focused mostly on Bengali text recognition.
Again, the Google Multilingual \ac{OCR} achieved the highest flex \ac{ca} in the text recognition task and success rate in the text region page segmentation.
The results from this competition were slightly better than those in 2017 but still remain quite low, and the authors suggested a focus on preprocessing for better performance.

\subsubsection{DIDA} 
\label{sssec:dida}

The Digit Dataset DIDA \cite{KUSETOGULLARI2021100182} is an extension of the previously mentioned ARDIS digit dataset. 
DIDA is composed of three datasets: Dataset I, with 250K single-digit color images of 10 classes (0-9), Dataset II, with 200K multi-digit year string samples, and Dataset III, with 25K digits with bounding boxes meant for object detection.
A digit detection and recognition system named DIGITNET was proposed that initially detects handwritten digits and passes the output to a recognition network to classify them.
The results of this system were further evaluated on DIDA, comparing it to other classical methods \cite{merabti2018segmentation, Chen2000SegmentationOS,Gattal2017SegmentationAR} and network architectures such as YOLOv3 and YOLOv3-tiny \cite{Redmon2018YOLOv3AI}. 
Similar to ARDIS, several experiments with different combinations of datasets were performed and state-of-the-art results in digit detection were achieved.

\subsubsection{ScribbleLens }
\label{sssec:scribblelens}

In \cite{9257750}, a corpus for automatic manuscript transcription was presented. 
This dataset contains 1K pages from early modern Dutch manuscripts spanning over 150 years with line, character, year, and writer ground truth. 
It further provides a set of unlabeled images for unsupervised or weakly-supervised learning investigation. 
As a baseline, a network that combines Convolutional Neural Networks and bi-directional \ac{LSTM} with a \ac{CTC} (CNN/BLSTM/\ac{CTC}) \cite{7814068, Nina2018NephiA} was used, and it was shown that the \ac{CER} would be reduced in the presence of additional annotated data.

\subsubsection{ICDAR 2019 Competition on Recognition of Historical Arabic Scientific Manuscripts (RASM2019)}
\label{sssec:rasm2019}

The next RASM competition after the one presented in section \ref{sssec:rasm2018} is the ICDAR RASM19 \footnote{\url{https://www.primaresearch.org/RASM2019/}\label{rasm2019}}, which focused on the recognition of archival Arabic scientific manuscripts.
This version of the competition offers 20 training images with PAGE XML annotations \cite{5597587} and 100 test images to evaluate the systems.
The ground truth has the same format and content as the previous competition for the three tasks: text block detection, text-line detection/segmentation, and text recognition.
Although the competition did not provide any detailed information due to the absence of a published paper, a graph was provided for every task containing the results of the submitted systems.
A Google submission shows the highest success rate for the first task and an RDI system shows the highest success rate for the second task. 
For text recognition in normalized text, a 77.58\% flex \ac{ca} is achieved again by the RDI system.
We suspect that the RDI winning systems are the same as those in the previous round of the competition, however it is not clear in the competition's website.

\subsubsection{ICDAR 2019 Competition on Image Retrieval for Historical Handwritten Documents (ICDAR19-HDRC-IR) }
\label{sssec:ICDAR2019-HDRC-IR}

This competition \cite{Christlein2019ICDAR2C} followed the previous competition mentioned in \ref{sssec:Historical-WI} and handled the task of image retrieval according to writer style by providing a larger test set of 20K images from over 10K different writers.
For training, the competition proposed the dataset from the previous competition and further enlarged the training with images from Letters A and Manuscripts.
The \ac{mAP} constituted the evaluation metric, similar to the previous competition. 
The winning system obtained a 92.5\% \ac{mAP} using SIFT \cite{LoweDavid2004DistinctiveIF} and Pathlet features \cite{8978107} projected into a lower dimension space using SVD on the ICDAR17 Historical-WI data feature matrices, and then concatenated and normalized them to compute global descriptors using Euclidean distance.

\subsubsection{Handwritten Text Recognition (HTR) Benchmarks}
\label{sssec:htr_benchmarks}

The work published in \cite{SANCHEZ2019122} presents four benchmarks for historical document HTR and achieves state of the art results for four different competitions: ICFHR-2014 \cite{6981116}, ICDAR-2015 \cite{7333944}, ICFHR-2016 \cite{Snchez2016ICFHR2016CO}, and ICDAR-2017 \cite{8270157}. 
The ICFHR-2014 dataset is a subset of the Bentham Papers \cite{Causer2012BuildingAV} that contains 433 images with line detection and recognition ground truth in PAGE XML format.
Similarly, the ICDAR-15 competition  contains Bentham page images, but presenting a more difficult layout than those of ICFHR-2014.
This dataset consists of different subsets that include line images with their corresponding line transcriptions aligned, or images with page-level transcriptions, but no alignment.
The ICFHR-2016 dataset includes 450 single-block page images derived from the German Ratsprotokolle collection, that contain approximately 10K lines and 43K running words.
The provided ground truth is at line-level.
The three mentioned competitions include a Restricted and an Unrestricted Track.
Finally, the ICDAR-2017 competition provides 10,172 images distributed across two training and two test subsets.
The data provided come from the Alfred Escher Letter (AEC) and other German collections and present heterogeneous writing styles.
The competition includes a Traditional challenge for simple transcription and an Advanced challenge for transcription, but with a pre-step of line detection.
For the benchmarking, a CRNN with four convolutional and three recurrent layers is used for character optical modeling and enhanced with the use of N-gram language models on the output character probabilities.
With this enhancement, the work achieves the lowest \ac{CER} and \ac{WER} for all cases.

\subsubsection{ICFHR 2020 Competition on Image Retrieval for Historical Handwritten Fragments (HisFragIR20) }
\label{sssec:HisFragIR20}

Another competition, which is similar to the ICDAR17 Historical-WI (Section \ref{sssec:Historical-WI}) and the ICDAR2019-HDRC-IR (Section \ref{sssec:ICDAR2019-HDRC-IR}) is the HisFragIR20 \cite{Seuret2020ICFHR2C}.
This competition further increased the size of the dataset by generating 120K fragments, randomly shaped and rectangular, from 20K documents and 9.8K writers.
The test data come from European Middle Age books (\nth{9} to \nth{15} century CE).
Fragments extracted from the ICDAR2019-HDRC-IR test set comprise the training set.
The competition evaluated the test set for two tasks, retrieval per writer and per image.
For the writer task, the best system in terms of \ac{mAP} used a ResNet \cite{He2016DeepRL} with 20 layers trained on SIFT keypoints and multi-\ac{VLAD} encoding, PCA for descriptor dimensionality reduction, k-means clustering on the descriptors, and cosine similarity for the final results.
The whole process was based on the work presented in \cite{christlein2017unsupervised}.
Accuracy, Pr@10, and Pr@100 metrics were used.
The system that achieved the highest values used a ResNet50 \cite{He2016DeepRL} feature extractor with whole fragment image input and the $\chi^2$ distance.
This system also obtained the highest values in all retrieval per image task metrics.

\subsubsection{Digital Peter}
\label{sssec:digital_peter}

Digital Peter \cite{10.1145/3476887.3476892} is a dataset of 9,694 images and their corresponding text from manuscripts, written by Peter the Great from 1709 to 1713 for handwriting recognition.
This dataset provides a 6,237 training, 1,930 validation, and 1,527 test splits that can be used either for line segmentation or line recognition.
A competition\footnote{\url{https://github.com/sberbank-ai/digital\_peter\_aij2020}\label{digital_peter_note}} on text-line recognition was launched using this dataset.
As a baseline, a 7-layer \ac{CNN} was used for image feature extraction and then a bidirectional GRU network with \ac{CTC} loss \cite{10.1145/1143844.1143891} was used to predict the image text.
The model performance was further optimized using different hyperparameter values and beam search.
The task was evaluated according to the \ac{CER}, the \ac{WER}, and the string \ac{acc}.

\subsubsection{Hugin-Munin}
\label{sssec:hugin_munin}

The Hugin-Munin dataset \cite{10.1007/978-3-031-06555-2_27} is the first handwritten recognition dataset for text written in Norwegian.
This dataset contains images derived from diaries and private correspondences written from 1820 to 1950 from 12 different writers.
The ground truth includes the transcriptions of 164,922 words or 23,732 lines in PAGE XML format.
The authors provide a 80\% training - 10\% validation - 10\% test random split and another split with 3 unseen writers in the test set.
They further present a survey of open-source handwritten text recognition libraries used since 2019 and compare the performance using the \ac{CER} and \ac{WER} on the random split data.
The lowest \ac{CER} is obtained using PyLaia \cite{puigcerver2018pylaia}, while the lowest \ac{WER} is obtained using Kaldi \cite{Arora2019UsingAM}.
These best methods were further deployed on the writer split and achieve much lower performance than the random split.

\subsubsection{POPP}
\label{sssec:popp}

The POPP dataset \cite{10.1007/978-3-031-06555-2_10} contains lines extracted from Paris census tables of 1926 and consists of three sub-datasets: the "Generic dataset", the "Belleville", and the "Chaussée d'Antin".
The Generic dataset contains 80 double page images, one for every Paris district, each one from a different writer, and 4,800 lines divided in 3,840 train, 480 validation, and 480 test lines.
The Belleville dataset contains 49 pages and 1,470 lines from the Belleville district written from a single writer.
The Chaussée d'Antin is a 10-writer set of 780 lines and 26 pages from the Chaussée d'Antin census.
POPP includes grayscale images, their corresponding line bounding boxes in XML files, and the line labels in JSON format.
This work presents line recognition results of the \ac{CER} and \ac{WER} for each of the three datasets using an end-to-end hybrid attention network \cite{Coquenet2022}.
Finally, the paper presents a complete pipeline, with the steps of pre-processing, handwriting recognition, and domain knowledge integration, that extracted a vast number of information from the Paris census to be used as annotated dataset and improves the \ac{CER} with the use of self-training.


\section{Observations \& Discussion}
\label{sec:discussion}

Several datasets exist for the tasks in the three categories that we presented in this paper: document classification, document structure, and content analysis.
There is a variation in languages, tasks, and sizes; however, no large-scale dataset seems to exist that can address various tasks and be used by the community for pretraining or transfer learning.
Various evaluation methods are also presented when benchmarking. 
Nevertheless, it is difficult to directly compare datasets and techniques, as there is no universal evaluation that can directly compare the performance of systems on datasets.

The classification of objects on a page level is highly represented in document structure tasks. 
Nevertheless, a document could also be considered as the whole manuscript collection.
We found six studies related to document classification, which means that this task is rarely addressed.
Only one dataset offers more than 35K pages, but the overall amount is relatively low.
The main focus of datasets is on Latin scripts, while others such as Arabic are also represented.
Some scripts, such as Hebrew or Greek, are rarely represented.
We detected more meta-data information in several datasets that we categorized in the document structure and content analysis sections, as they are not used or included in the benchmarks.

Considering the document structure studies, we found only two datasets containing more than 10K images.
Again, there is a significant focus on Latin scripts; however, more languages are observed for this task as it is much more represented than document classification.
A noticeable issue, in this case, is the comparison across databases as a variety of evaluation measures and benchmarks are used.
We propose harmonizing the evaluation metrics using the \ac{mIoU} and \ac{mAP} metrics (at 50\%, 60\%, 70\%, and 80\%).
In terms of annotation format, we note that most datasets use PAGE XML, three datasets use the COCO format, and only one dataset uses the VOC format.
It would be beneficial to establish a conversion between annotation formats to promote the use of state-of-the-art computer vision models for historical document analysis.

The content analysis task seems to have the most prominent representation in the set of datasets.
In this case, approximately 30\% of the semantics-related studies include more than 1K images.
The majority of this percentage appears for isolated character recognition, which is reasonably the easiest case of samples one could obtain and manage in a database.
In general, there is an emphasis on \ac{OCR}, but the level of detail differs (character, word, or line).
Retrieval further focuses on text on the word, image, or writer level.
Likewise, there is a focus on Latin scripts.
Still, there is also a high representation of Asian scripts and the least representation on Arabic scripts.
Finally, there is more interest in paleography, but we lack the representation of digits and tables as content.

\section{Conclusion}
\label{sec:conclusion}

We demonstrated a survey of historical document image datasets following a systematic literature review methodology.
We summarized 65 studies and clustered them considering the related general tasks that we defined.
We list the datasets in a table, connecting them to their corresponding section, and mark the possible tasks they include.
For every study, we tabulate detailed information about the statistics, tasks, document type, languages, input image visual aspects, annotations, and benchmark and quantitative performance analysis information.
This way, we facilitate researchers in finding the most fitting datasets and enable historical document image analysis.

Our findings unveil a focus on Latin scripts and several evaluation methods, but not much compliance with deep learning trends.
A clear size limitation on dataset samples is also obvious.
As future directions, we urge the need for large-scale datasets to apply state-of-the-art deep learning methods, the inclusion of more classification tasks using metadata information, and the harmonization of evaluation schemes for direct comparison across datasets.

\newgeometry{width=3cm, height=27.7cm, left=8cm}
\thispagestyle{empty}
\begin{sidewaystable}
\small
\sidewaystablefn%
\begin{center}
\begin{adjustbox}{scale=0.9,center}
\begin{minipage}{\textheight}
\renewcommand\arraystretch{1.4}
  \centering
  \addtolength{\tabcolsep}{-2.5pt}
  
  \caption{Historical document image datasets with information about statistics, classes, tasks, language, document type, input visual aspects, ground truth, and benchmarks present in their original papers or competitions. The datasets are presented in the same order as in Table \ref{tab:structure_content_class} (earliest to latest).}
  \scalebox{0.52}{
  \begin{tabular}{l c c c c c c c c c c c c c}
  \hline
  \toprule
  
  \multicolumn{6}{c}{\textbf{General Information}} & 
  \multicolumn{3}{c}{\textbf{Input}} &
  \multicolumn{2}{c}{\textbf{GT}} &
  \multicolumn{3}{c}{\textbf{Benchmark}}\\ 
  \cmidrule(lr){1-6}
  \cmidrule(lr){7-9}
  \cmidrule(lr){10-11}
  \cmidrule(lr){12-14}
  \begin{tabular}{l}\textbf{Dataset name}\\ \textbf{[reference] - section}\end{tabular}&\textbf{Statistics}&\textbf{Classes}&\textbf{Task}&\textbf{Language}&  \textbf{Document type}
         & 
  \textbf{Mode} &
  \textbf{Resolution} &
  \textbf{Format}& \textbf{Annotation} & \textbf{Format} & \textbf{Model} & \textbf{Metric} & \textbf{Performance}\\
  
  \bottomrule
  \hline 

  \textbf{\begin{tabular}{l}GERMANA\\ \cite{5277691} - \ref{sssec:germana}\end{tabular}} &
  \begin{tabular}{c}764 pages\\20,529 text lines\\217.2K words
  \end{tabular} & \begin{tabular}{c}27.1K

  50word lexicon\\115 character set \end{tabular}& \begin{tabular}{c}Handwriting recognition \\Text block detection\\Baseline detection\end{tabular} &  
  \begin{tabular}{c}Spanish, Catalan\\Latin, French\\German, Italian
  \end{tabular} & \begin{tabular}{c}Page scans of Spanish\\ manuscript from 1891 \\on the life of\\ Germana de Foix \end{tabular}& Color & 300 dpi & N/A & \begin{tabular}{c}Text block rectangles\\Baselines\\Line transcriptions\end{tabular}& N/A& \begin{tabular}{l}HMM-based text \\image modeling\\ and n-gram\\ language modeling \cite{4377054}\end{tabular} & \begin{tabular}{c}Transcription\\WER\\(\%)\end{tabular} & 37 \\

  \hline


  \textbf{\begin{tabular}{l}RODRIGO\\ \cite{serrano-etal-2010-rodrigo} - \ref{sssec:rodrigo}\end{tabular}} &
  \begin{tabular}{c}853 pages\\20,357 lines\\232K words
  \end{tabular} & \begin{tabular}{c}17K word lexicon\\115 character set \end{tabular} & \begin{tabular}{c}Handwriting recognition \\Text block detection\\Baseline detection\end{tabular} &  \begin{tabular}{c}Old Castilian
  \end{tabular} &\begin{tabular}{c}Page images of\\ a manuscript from \\1545 written in \\old Castilian by\\one writer \end{tabular} & Color & 300 dpi & N/A & \begin{tabular}{c}Text block rectangles\\Baselines\\Line transcriptions\end{tabular}& N/A
   & \begin{tabular}{l}HMM-based text \\image modeling\\ and n-gram\\ language modeling \cite{4377054}\end{tabular} & \begin{tabular}{c}Transcription\\WER\\(\%)\end{tabular} & 36.5 \\
  
  \hline


  \textbf{\begin{tabular}{l}IAM-HistDB\\
  Saint Gall\\\cite{10.1145/2037342.2037348} - \ref{sssec:iam-histdb}\end{tabular}} &
  \begin{tabular}{c}60 pages\\1,410 text-lines\\11,597 words\\5,436 word spellings
  \end{tabular} & \begin{tabular}{c}4,890 word labels\\49 letters\\
  \end{tabular} & \begin{tabular}{c}Handwriting recognition\\Layout analysis\end{tabular} &  \begin{tabular}{c}Latin
  \end{tabular} &\begin{tabular}{c}Page images of\\ a \nth{9} century manuscript \\ written in Carolingian script \\by a single writer \end{tabular} & Color & 300 dpi & \begin{tabular}{c}JPG\\PNG\end{tabular} & \begin{tabular}{c}Text-line locations \\Word locations\\Line-level transcriptions\end{tabular}& \begin{tabular}{c}SVG\\TXT\end{tabular}
   & \begin{tabular}{c}HMM (proposed) \cite{10.1145/2037342.2037348}\end{tabular} & Alignment \ac{acc} (\%) &  92.07\\
  
  \hline


  \textbf{\begin{tabular}{l}IAM-HistDB\\
  Parzival\\\cite{5306020} - \ref{sssec:iam-histdb}\end{tabular}} &
  \begin{tabular}{c}47 pages\\4,477 text lines\\23,478 words
  \end{tabular} & \begin{tabular}{c}4,934 word labels\\93 letters\\
  \end{tabular} & \begin{tabular}{c}Handwriting recognition\\Word spotting \\Layout analysis\end{tabular} &  \begin{tabular}{c}Medieval German
  \end{tabular} &\begin{tabular}{c}Page images of\\ a \nth{13} century manuscript \\ written in Gothic script \\by 3 writers \end{tabular} & \begin{tabular}{c}Color\\Binary\end{tabular} & 300 dpi & \begin{tabular}{c}JPG\\PNG\end{tabular} & \begin{tabular}{c}Word-level transcriptions\\Line-level transcriptions\\Word labels\end{tabular}& \begin{tabular}{c}TXT\end{tabular} 
   & \begin{tabular}{l}HMM-based recognizer \cite{10.5555/505741.505745}\\BLSTM \cite{Graves2009855}\end{tabular} & \begin{tabular}{c}\\Word \ac{acc} (\%)\\\ac{WER} (\%)\end{tabular} &
   \begin{tabular}{cc}HMM \cite{10.5555/505741.505745}& BLSTM \cite{Graves2009855}\\ \hline 88.69 & 93.32\\11.31 & 6.68\end{tabular}\\
  
  \hline


  \textbf{\begin{tabular}{l}IAM-HistDB\\
  George \\Washington (GW)\\\cite{FISCHER2012934} - \ref{sssec:iam-histdb}\end{tabular}} &
  \begin{tabular}{c}20 pages\\656 text lines\\4,894 words
  \end{tabular} & \begin{tabular}{c}1,471 word labels\\82 letters\\
  \end{tabular} & \begin{tabular}{c}Handwriting recognition\\Word spotting\end{tabular} &  \begin{tabular}{c}English
  \end{tabular} &\begin{tabular}{c}Page images of\\ a \nth{18} century manuscript \\ written in longhand script \\by 2 writers \end{tabular} & Binary & 300 dpi & PNG & \begin{tabular}{c}Word-level transcriptions\\Line-level transcriptions\end{tabular}& \begin{tabular}{c}TXT\end{tabular} 
   & \begin{tabular}{l}BLSTM + CTC new (proposed)\\HMM \cite{10.1016/j.patcog.2009.02.005}\\DTW \cite{Rath2006WordSF}\end{tabular} & \begin{tabular}{c}Average precision\\of spotting task \end{tabular} & \begin{tabular}{c}0.84\\0.60\\0.48\end{tabular}  \\
  
  \hline

  
  \textbf{\begin{tabular}{l}PHTD\\ \cite{6121553} - \ref{sssec:phtd} \end{tabular}} & 
    \begin{tabular}{c} 140 handwritten documents \\
    1,787 handwritten text-lines \\
    27,073 words\end{tabular}& 3 Types of text & \begin{tabular}{c} Text-line segmentation \\ Sentence recognition/understanding \\ Word segmentation/recognition\\ Characters segmentation \\ Word spotting\\ Text-line extraction\\Writer identification 
    \end{tabular}& Persian & \begin{tabular}{c} Persian handwritten\\ text documents\end{tabular} & Grayscale & 300 dpi& TIFF & \begin{tabular}{c}
             Pixel-based \\ Text content-based
        \end{tabular}& \begin{tabular}{c}
             DAT \\ TXT
        \end{tabular} & \begin{tabular}{l}
             Alaei et al.\cite{Alaei2011ANS}\\ PPSL \cite{10.1007/s10044-011-0226-x}
        \end{tabular} &  \begin{tabular}{c}Segmentation \\\ac{acc} \% \end{tabular}& \begin{tabular}{c}94.00 \\ 89.43\end{tabular}\\
  \hline


  \textbf{\begin{tabular}{l}PBOK\\ \cite{Alaei2012DatasetAG} - \ref{sssec:pbok}\end{tabular}} &\begin{tabular}{c}707 pages\\12565 text-lines\\
  104541 words\\553536 characters\\436 writers
  \end{tabular} & \begin{tabular}{c}4 languages\\436 writers
  \end{tabular} &\begin{tabular}{c}Text-line segmentation\\Word segmentation\\
  Word recognition\end{tabular} &  \begin{tabular}{c}Persian\\Bangla\\Oriya\\Kannada
  \end{tabular} & \begin{tabular}{c}Handwritten\\ documents\end{tabular} & Grayscale& 300 dpi & TIFF &\begin{tabular}{c}Pixel-level\\Content-level\end{tabular} & \begin{tabular}{c}DAT\\TXT\end{tabular} 
   & \begin{tabular}{l}
             Alaei et al. \cite{Alaei2011ANS} \\ PPSL \cite{10.1007/s10044-011-0226-x}
        \end{tabular} & \begin{tabular}{c}(\%)\\\ac{DR}\\\ac{RA}\\TLDM \end{tabular} & 
        \begin{tabular}{cc}
             Alaei et al. \cite{Alaei2011ANS}& PPSL \cite{10.1007/s10044-011-0226-x}\\ \hline
             91.33 & 88.07\\
             90.41 & 86.69\\
             90.87 & 87.38
        \end{tabular}\\
   
  \hline

  
  \textbf{\begin{tabular}{l}IMPACT\\ \cite{10.1145/2501115.2501130} - \ref{sssec:impact}\end{tabular}} & 
    \begin{tabular}{c}
     600K images\\ 45K ground-truthed\\images\\70K word outlines
        \end{tabular} & \begin{tabular}{c}
     Text (+subcatecories), \\ Graphics(+subcategories), \\ Image, Line drawing, \\Separator, Table\\Chart, Maths\\
        \end{tabular}& Layout analysis & 18 languages & \begin{tabular}{c}European printed \\documents\end{tabular}  & 
        Color & N/A &  \begin{tabular}{c}
     TIFF \\ JPEG \\ JPEG2000
        \end{tabular} & \begin{tabular}{c}
             Region outlines\\ Region text content \\Reading order\\Text, Word outlines
        \end{tabular} & \begin{tabular}{c}PAGE \\XML \end{tabular} & N/A & N/A & N/A\\
  \hline
  
  
  \textbf{\begin{tabular}{l}ESPOSALLES\\LICENCES\\ \cite{ROMERO20131658} - \ref{sssec:esposalles}\end{tabular}} & 
     \begin{tabular}{c}173 pages\\1,747 licences\\
     5,447 lines\\60,777 words\\ 328,229 characters\end{tabular}& \begin{tabular}{c}3,465 lexicon size\\85 character classes\end{tabular}& \begin{tabular}{c}Off-line handwriting\\ recognition\end{tabular} & Spanish & \begin{tabular}{c}Handwritten documents \\ from marriage \\license books\end{tabular}  & Color & 300 dpi &  TIFF & \begin{tabular}{c}Text blocks \\ Lines \\Transcriptions\end{tabular} & N/A&\begin{tabular}{l}HMM - PRHLT \cite{Toselli2004IntegratedHR}\\HMM - IAM \cite{Toselli2004IntegratedHR, 10.5555/505741.505745}\\BLSTM - PRHLT \cite{Graves2009855, Toselli2004IntegratedHR}\\BLSTM - IAM
     \cite{Graves2009855, 10.5555/505741.505745}\end{tabular} & \begin{tabular}{c}Transcription\\WER\\(\%)\end{tabular} &\begin{tabular}{c}11.0\\14.6\\13.1\\9.0
     \end{tabular}\\
  \hline

  
  \textbf{\begin{tabular}{l}ESPOSALLES\\INDEX\\ \cite{ROMERO20131658} - \ref{sssec:esposalles}\end{tabular}} & 
     \begin{tabular}{c}29 pages \\1,563 lines\\
     6,534 words\\30,809 characters\end{tabular}& \begin{tabular}{c}1,725 lexicon size\\68 character classes\end{tabular}& \begin{tabular}{c}Off-line handwriting\\ recognition\end{tabular} & Spanish & \begin{tabular}{c}Handwritten documents \\ from marriage \\license books\end{tabular}  & Color & 300 dpi &  TIFF & \begin{tabular}{c}Text blocks \\ Lines \\Transcriptions\end{tabular} &N/A &\begin{tabular}{l}HMM - PRHLT \cite{Toselli2004IntegratedHR}\\HMM - IAM \cite{Toselli2004IntegratedHR, 10.5555/505741.505745}\\BLSTM - PRHLT \cite{Graves2009855, Toselli2004IntegratedHR}\\BLSTM - IAM
     \cite{Graves2009855, 10.5555/505741.505745}\end{tabular} & \begin{tabular}{c}Transcription\\WER\\(\%)\end{tabular} &\begin{tabular}{c}31.1\\44.7\\70.1\\59.8
     \end{tabular}\\
  \hline
  
  
  \textbf{\begin{tabular}{l}BH2M\\ \cite{6976764} - \ref{sssec:bh2m}\end{tabular}} & 
     \begin{tabular}{c}174 pages\\1,740 licenses\\5,498 lines\\
     56,645 words\end{tabular} & 3,360 word classes
      & \begin{tabular}{c}Layout analysis, \\ recognition and\\ understanding  \end{tabular} & Old Catalan & \begin{tabular}{c}Handwritten marriage \\ record archives \\ of Barcelona Cathedral\end{tabular}  & Color & 300 dpi & N/A  & \begin{tabular}{c}Layout\\Text lines \\Word bbox\\ Transcription\\ Semantic information  \end{tabular} & XML & \begin{tabular}{l}Line segmentation \cite{Mota2014AGA}\\DTW + Vinciarelli \cite{Vinciarelli2002OfflineCW}\\HOG+EWS \cite{Almazn2012EfficientEW}\end{tabular}& \begin{tabular}{c}DR - RA - FM (\%)\\mAP (\%)\\mAP (\%)\end{tabular} &\begin{tabular}{c}83.1 - 81.3 - 82.1\\31.51\\51.35\end{tabular} \\
  \hline

  
  \textbf{\begin{tabular}{l}HADARA80P\\ \cite{6980990} - \ref{sssec:hadara80p}\end{tabular}} & 
     \begin{tabular}{c}80 pages \\ 16,720 words\end{tabular}& 
     Word segment classes & Word-spotting & Arabic & \begin{tabular}{c}Handwritten  \\ Arabic documents\\of one writer\\about the taaum\\disease \end{tabular}  & Color & \begin{tabular}{c}$2882\times3650$\\pixels \end{tabular}& TIFF & \begin{tabular}{c}Page segments \\ Text blocks \\Word polygons \\Word transcriptions\end{tabular} & XML
     &\begin{tabular}{l}Ulysse\footref{ulysse} \cite{Leydier2007TextSF, Leydier2009TowardsAO}\\HADARA (proposed)\end{tabular}
     &\begin{tabular}{c}\\mAP ($p_{IR}$)\\mAP ($\overline{\gamma_{LA}}$)\end{tabular}
     &\begin{tabular}{cc}Ulysse & HADARA \\\hline 0.35&0.41\\0.27&0.31\end{tabular}\\
  \hline

  \textbf{\begin{tabular}{l}ENP\\ \cite{7333898} - \ref{sssec:enp}\end{tabular}} & 
    \begin{tabular}{c} 
    528 pages, 61K regions\\208 tables, 1K graphics \\
    47K text regions\\202K text-lines\\
    \end{tabular} & \begin{tabular}{c} 
    Regions, Tables\\ Images/Graphics \\
    Text regions, Text lines\\
    \end{tabular}& \begin{tabular}{c} \ac{OCR} \\ Text recognition \\ Layout analysis
    \end{tabular}& \begin{tabular}{c}
       Dutch, English, Estonian  \\
       Finnish, French, German \\ 
       Latvian, Polish, Russian \\
       Serbian, Swedish \\
       Ukrainian, Yidish \\
  \end{tabular} & \begin{tabular}{c}
       European Cultural  \\
       Heritage 
  \end{tabular}  & \begin{tabular}{c}
       Color  \\Grayscale \\B/W
  \end{tabular} & \begin{tabular}{c}
             300 dpi\\
             400 dpi
        \end{tabular} & TIFF & \begin{tabular}{c}
             Unicode text\\ Layout\\ Type labels\\
             Reading order
        \end{tabular}& PAGE & \begin{tabular}{l}\\
             ABBYY FineReader\\ Tesseract 3.03\footref{tesseract}
        \end{tabular}&\begin{tabular}{c}
             \\Success \\ rate (\%)
        \end{tabular}&\begin{tabular}{cc}
             Keyword search &Content retrieval\\ \hline 78.9 & 95.9\\69.8 & 95.5
        \end{tabular}\\
  \hline

  \end{tabular}}
  \label{tab:large_table}


\end{minipage}
\end{adjustbox}
\end{center}

\end{sidewaystable}

\restoregeometry


\newgeometry{width=3cm, height=29cm, left=8cm}
\thispagestyle{empty}
\begin{sidewaystable}
\small
\sidewaystablefn%
\begin{center}
\begin{adjustbox}{scale=0.8,center}
\begin{minipage}{\textheight}
\renewcommand\arraystretch{1.4}
\ContinuedFloat
  \centering
  \addtolength{\tabcolsep}{-3pt}
  
  \caption{Historical document image datasets with information about statistics, classes, tasks, language, document type, input visual aspects, ground truth, and benchmarks present in their original papers or competitions. The datasets are presented in the same order as in Table \ref{tab:structure_content_class} (earliest to latest).}
  \scalebox{0.52}{
  \begin{tabular}{l c c c c c c c c c c c c c}
  \hline
  \toprule
  
  \multicolumn{6}{c}{\textbf{General Information}} & 
  \multicolumn{3}{c}{\textbf{Input}} &
  \multicolumn{2}{c}{\textbf{GT}} &
  \multicolumn{3}{c}{\textbf{Benchmark}}\\ 
  \cmidrule(lr){1-6}
  \cmidrule(lr){7-9}
  \cmidrule(lr){10-11}
  \cmidrule(lr){12-14}
  \begin{tabular}{l}\textbf{Dataset name}\\ \textbf{[reference] - section}\end{tabular}&\textbf{Statistics}&\textbf{Classes}&\textbf{Task}&\textbf{Language}& \textbf{Document type} & 
  \textbf{Mode} &
  \textbf{Resolution} &
  \textbf{Format}& \textbf{Annotation} & \textbf{Format} & \textbf{Model} & \textbf{Metric} & \textbf{Performance}\\
  
  \bottomrule
  \hline 
  
  \textbf{\begin{tabular}{l}GRPOLY-DB\\ \cite{7333841} - \ref{sssec:GRPOLY-DB}\end{tabular}} & 
    \begin{tabular}{c} 399 pages, 15,084 text \\
    lines, 102,596 words\\
    171,511 characters
    
    \end{tabular} & \begin{tabular}{c} More than 270\\
    character classes
    \end{tabular}& \begin{tabular}{c} Word and text \\ line segmentation,\\  Text and isolated\\ character recognition, \\ Word spotting
    \end{tabular}& Greek & \begin{tabular}{c}
       Handwritten and printed  \\
       document pages\\ from 1838-1977 that\\ contain greek\\ polytonic characters
        
  \end{tabular}  & \begin{tabular}{c}Color\\Grayscale
       
  \end{tabular} & N/A & N/A & \begin{tabular}{c}
        Text line and word-\\level for segmentation\\
        Text line, word and\\ character-level transcriptions\\
        Query-by-examples
        \end{tabular}& PAGE & \begin{tabular}{l}
            Text line segm: Shredding-based method \cite{5277573}\\
            Word segm: Sequential clustering \cite{953781}\\Isolated char rec: HoG features \cite{1467360} + SVM\\
            Text recognition : ABBYY FineReader\\
            Word spotting: Profiles + DTW \cite{Rath2006WordSF}
        \end{tabular}&\begin{tabular}{c}
             FM (\%)\\FM (\%)\\ \ac{RA} (\%)\\\ac{CER} (\%)\\mAP (\%)
        \end{tabular}&\begin{tabular}{c}
             94.58\\94.85\\98.37\\19.20\\73.93
        \end{tabular}\\
  \hline

  
  \textbf{\begin{tabular}{l}DocExplore\\ \cite{10.1117/1.JEI.26.1.011010} - \ref{sssec:docexplore}\end{tabular}} & 
     \begin{tabular}{c}1,500 images\\ 1,464 queries\end{tabular}& \begin{tabular}{c}35 object categories\\ (human faces, decoration\\ objects, ornate initial\\letters, etc.)\end{tabular}& \begin{tabular}{c}Image retrieval\\ Pattern localization\end{tabular} & N/A & \begin{tabular}{c}Page images of\\ manuscripts from \\the \nth{10} and \nth{16}\\from
the Municipal\\ Library of Rouen\\with graphics\end{tabular}  & \begin{tabular}{c}Color\\Grayscale\end{tabular} & 72 dpi& JPEG  &\begin{tabular}{c}Name of belonging \\query category \\for every image\end{tabular} &N/A& \begin{tabular}{l}Patter spotting \\system with VLAD \cite{EN2016149}\end{tabular}& mAP (\%) &\begin{tabular}{l} Retrieval: 0.613 \\Localization: 0.111 \end{tabular}\\
  \hline

  \textbf{\begin{tabular}{l}DIVA-HisDB\\ \cite{7814109} - \ref{sssec:diva-hisdb}\end{tabular}} & \begin{tabular}{c} 150 images: 20 train,\\
        10 validation, 10 test, \\10 left out images\\for every manuscript\\
        493M pixels
       \end{tabular} & \begin{tabular}{c} Background\\ Main text\\
       Decorations\\ Comments\\ 
       \end{tabular} & \begin{tabular}{c} Layout analysis\\
        Element extraction
       \end{tabular} & \begin{tabular}{c} Latin\\
        Italian\\ \end{tabular} & \begin{tabular}{c} Pages from 3 \\
        medieval manuscripts\\ from \nth{11} and \nth{14}\\century \end{tabular}  & Color &600 dpi & JPG & Pixel-level & PNG & \begin{tabular}{l}Pixel class: N-light-N  \cite{7814107}\\Task 1: FCN  \cite{8270154}\\
        Task 2: ARLS+SC  \cite{Nikolaou2010SegmentationOH}\\
        Task 3: ARLS+SC  \cite{Nikolaou2010SegmentationOH}
        \end{tabular} & \begin{tabular}{c}\ac{acc} (\%)\\IoU (\%)\\F1-score (\%)\\Line IoU (\%)
        \end{tabular} & \begin{tabular}{c}95.55\\99.00\\98.22\\96.99
        \end{tabular}\\
  \hline

  
  \textbf{\begin{tabular}{l}AMADI\_LontarSet\\ \cite{7814058} - \ref{sssec:amadi_lontarset}\end{tabular}} & 
     \begin{tabular}{c}100 pages\\
     Binarization: 50 train,\\50 test images, 100\\ binarized images\\
     Word spotting: 130 train,\\100 test images,
     15K \\word patched\\
     Character recognition: 11K train-\\7K test patch images\end{tabular}& \begin{tabular}{c}133 character\\ classes\end{tabular}& \begin{tabular}{c}Binarization\\ Word spotting\\ Isolated character recognition\end{tabular} & Balinese & \begin{tabular}{c}Palm leaf\\ Balinese\\ manuscripts\end{tabular}  & Color & N/A &  JPG & \begin{tabular}{c}Binarized,  \\ word and character \\annotated images \end{tabular} & \begin{tabular}{c}PNG\\BMP \\ TXT\\XML\end{tabular} &
     \begin{tabular}{l} Challenge 1: FCN \cite{1048482, 7814130}\\Challenge 3: VMQDF \cite{7814130}\end{tabular} & \begin{tabular}{c}FM - NRM - PSNR \\Rec rate(\%)\end{tabular} & \begin{tabular}{c}68.76 - 0.13 - 33.39 \\88.39 \end{tabular} \\
  \hline


  \textbf{\begin{tabular}{l}ICFHR 2016\\ CLAMM\\ \cite{7814129} - \ref{sssec:ICFHR16CLAMM}\end{tabular}} &
  \begin{tabular}{c} 2K training images\\\textbf{Task 1:} 1K test images\\ \textbf{Task 2:} 2K test images  \end{tabular} & \begin{tabular}{c} Uncial, Half-uncial, Caroline, Humanistic\\
  Humanistic Cursive, Praegothica, Southern\\ Textualis, Semitextualis, Textualis, Hybrida\\Semihybrida, Cursiva (12 classes)\end{tabular}&  \begin{tabular}{c}Script type\\classification
  \end{tabular}&  Latin & \begin{tabular}{c}Manuscripts from the\\ French catalogues,\\ the BVMM, and Gallica\end{tabular}  & Grayscale & 300 dpi & TIFF & \begin{tabular}{c}Script class label\\ index in\\ alphabetical order\end{tabular}& CSV
   & \begin{tabular}{l}\textbf{Task 1:} I-vector \cite{Dehak2011LanguageRV} + LDA \\
   \textbf{Task 1:} FRDC-OCR \cite{7814129}\\
    \textbf{Task 2:} DeepScript\footref{deepscript}\\
   \textbf{Task 2:} FRDC-OCR \cite{7814129}\end{tabular} & \begin{tabular}{c}\ac{acc} (\%)\\AID\\Final Score \\AID\end{tabular}& 
   \begin{tabular}{c}83.900\\0.018 \\2.967\\0.120\end{tabular}\\
  \hline
  

  \textbf{\begin{tabular}{l}ICDAR 2017\\ CLAMM\\ \cite{8270155} - \ref{sssec:ICFHR16CLAMM}\end{tabular}} &
  \begin{tabular}{c} ICFHR16 CLaMM train images\\\textbf{Task 1, 3:} 2K test images\\
  \textbf{Task 2, 4:} 1K test images \end{tabular} & 
  \begin{tabular}{c} Same script classes as ICFHR16\\ CLaMM, 15 date classes starting from\\before 1000 C.E. to 1600 C.E.\end{tabular}& \begin{tabular}{c}Script type and \\date classification
  \end{tabular} & Latin & \begin{tabular}{c}Manuscripts from the\\ French catalogues,\\ the BVMM, and Gallica\end{tabular}  & \begin{tabular}{l}Task 1,3: Grayscale\\Task 2,4: Grayscale\\and Color
  \end{tabular} & \begin{tabular}{l}Task 1,3: 300 dpi\\Task 2,4: 300\\and 400 dpi
  \end{tabular} & \begin{tabular}{l}Task 1,3: TIFF\\Task 2,4: TIFF\\and JPEG
  \end{tabular} & \begin{tabular}{c}Script and date \\class label index \\ next to image name\end{tabular}& CSV
   & \begin{tabular}{l}Task 1: T-Deep CNN\\Task 2: CK2
   \\Task 3: T-Deep CNN\\Task 4: CK2\end{tabular} & \ac{acc} (\%) & 
   \begin{tabular}{c}85.20\\76.50
   \\59.00\\49.90\end{tabular}\\
  \hline

  \textbf{\begin{tabular}{l}HBA\\ \cite{mehri:hal-01637826} - \ref{sssec:hba}\end{tabular}} & \begin{tabular}{c} 4K images\\
        7.58B pixels
      \end{tabular} & \begin{tabular}{c} Graphics\\ Normal text \\
       Capitalized text\\ Handwritten text \\
        Italic text\\ Footnote text\\
       \end{tabular} & \begin{tabular}{c} Layout analysis\\
        Pixel-level annotation
       \end{tabular} & \begin{tabular}{c} Latin\\
        Italian\\ \end{tabular} & \begin{tabular}{c} Manuscripts\\
        Printed pages\\ \end{tabular}  & \begin{tabular}{c} Color\\
        Grayscale\\ \end{tabular} & \begin{tabular}{c}
             300 dpi\\
             400 dpi
        \end{tabular} & TIFF & Pixel-level & \begin{tabular}{c}PNG\\TXT
        \end{tabular} & \begin{tabular}{l}\ac{FCN} \\ on pathes \cite{8978192}\end{tabular}& \begin{tabular}{c}\\\ac{acc} (\%)\\F-measure (\%)\\Weighted F (\%) \end{tabular}& \begin{tabular}{c c}Challenge 1 & Challenge 2
        \\ \hline99.80 & 99.08\\98.47 & 91.27\\99.80 & 99.08
        \end{tabular}\\
  \hline

  
  \textbf{\begin{tabular}{l}SleukRith\\\cite{10.1145/3151509.3151510} - \ref{sssec:sleukrith}\end{tabular}} & 
  \begin{tabular}{c}301,626 characters\\73,359 words\\3,245 text lines \end{tabular}& \begin{tabular}{c}207 character classes\\6,284 unique words \end{tabular} & \begin{tabular}{c}Isolated character recognition\\ Word, line Segmentation\end{tabular}& Khmer &
  \begin{tabular}{c}Palm leaf\\ manuscript\\ pages\end{tabular} & Color & N/A & JPG & \begin{tabular}{c}Character, word annotation\\polygon coordinates\\Character ID\\Line ID \end{tabular}& XML & CNN & \begin{tabular}{c}\ac{CER}\\(\%)\end{tabular} & 6.04\\
  
  \hline

  
  \textbf{\begin{tabular}{l}VML-HD\\\cite{8067751} - \ref{sssec:vml-hd}\end{tabular}} & \begin{tabular}{c}680 pages \\ 121,636 sub-words \\ 244,553 characters\end{tabular}& \begin{tabular}{c}Book, Page number\\Sub-word id\\Location coordinates\\Arabic, Latin annotation\\Sub-word length\end{tabular}& \begin{tabular}{c}Word-spotting\\Handwritten\\ sub-word recognition\end{tabular}& Arabic & \begin{tabular}{c}Handwritten scripts\\from 5 books\\by different writers\\from 1088-1451\end{tabular} & Color & \begin{tabular}{c}$6000\times 4000$ \\pixels\end{tabular} & TIFF & \begin{tabular}{c}Bounding boxes\\ Sequences of characters\\ per page\\Location coordinates
  \end{tabular} & Hadara XML & \begin{tabular}{c}\\Radial Descriptor \cite{6981050}\\Radial Descriptor Graph \cite{7814035}
  \end{tabular} & \begin{tabular}{c}\\\ac{DR} (\%)
  \end{tabular}  & \begin{tabular}{ccccc}Top1&Top2&Top3&Top4&Top5 \\\hline
  68.15&78.44&84.71&88.50&90.15\\
  83.40&89.84&92.48&94.00&95.11
  \end{tabular}\\
  
  \hline



  
  \textbf{\begin{tabular}{l}CFRAMUZ\\\cite{Arvanitopoulos2017AHF} - \ref{sssec:cframuz}\end{tabular}} & \begin{tabular}{c}7 novels \\ 64 pages \\ 18,027 words\end{tabular}& 2,998 unique words& \begin{tabular}{c}Word spotting\\ (segmentation-free)\end{tabular}& French & \begin{tabular}{c}Charles Ferdinand\\ Ramuz's novels\\written from\\1910 - 1946\end{tabular} & Grayscale & N/A & TIFF & \begin{tabular}{c}word id \\word bbox coords \\bbox width and height\\word location\\word transcription
  \end{tabular} & \begin{tabular}{c}Text \\XML
  \end{tabular} & \begin{tabular}{l}EAWS \cite{6857995}\\EEWS \cite{Almazn2012EfficientEW}\\BoVWWS \cite{Rusiol2011BrowsingHD}\\FKWS \cite{5277774}
  \end{tabular} & mAP (\%) & \begin{tabular}{c}88.07 \\29.20 \\50.47\\46.05
  \end{tabular}\\
  
  \hline
  
  
  \textbf{\begin{tabular}{l}Lontar Sunda\\\cite{8270066} - \ref{sssec:SUNDANESE_PALM_LEAF}\end{tabular}} &\begin{tabular}{c}66 pages, 1,526 train and \\317 test word images\\4,555 train and 2,816 test\\character images \end{tabular}
  & \begin{tabular}{c}61 character\\classes \end{tabular}& \begin{tabular}{c}A. Binarization\\B. Text-line segmentation\\C. Character recognition\\D. Word transliteration\\Word spotting \end{tabular} & Sundanese & \begin{tabular}{c}Sundanese manuscripts\\ of the \nth{15} century\\ from Garut, West\\ Java, and Indonesia\end{tabular}&  Color & N/A & \begin{tabular}{c}PNG\\TIFF\end{tabular} & \begin{tabular}{c}Word-level\\Character-level\\Binarized images\end{tabular} & \begin{tabular}{c}TXT \& PNG\\TXT\\BMP\end{tabular} & \begin{tabular}{l} \textbf{A.} Gaussian+non-linear enhance\\\textbf{B.} Horizontal projection profile\\\textbf{C.} Dense CNN\\\textbf{D.} CNN-RNN encoder-decoder\end{tabular} & \begin{tabular}{c}FM - NRM - PSNR\\DR - RA - FM (\%)\\Rec rate(\%)\\\ac{CER} (\%)\end{tabular} & \begin{tabular}{c}56.72 - 0.20 - 25.82\\63.55 - 46.87 - 53.95 \\86.54\\8.81 (\%)\end{tabular} \\
  
  \hline


  \textbf{\begin{tabular}{l}ICDAR 2017\\ REID2017\\ \cite{8270161} - \ref{sssec:reid2017}\end{tabular}} &
  \begin{tabular}{c} 5 train images
  \\26 test images \end{tabular} & 
  \begin{tabular}{c}Text\\Separator\\Graphic\\Image \end{tabular}& 
  \begin{tabular}{c}Layout analysis\\Text recognition
  \end{tabular} & \begin{tabular}{c}Bengali\\English 
  \end{tabular} & \begin{tabular}{c}Scanned images from \\printed books in\\ Bengali from\\ 1785-1909
  \end{tabular} & Color & N/A
   & TIFF &  \begin{tabular}{c}Layout polygons\\Transcriptions
  \end{tabular} & PAGE XML
   & \begin{tabular}{c}Google \ac{OCR}
  \end{tabular}   & \begin{tabular}{c}\textbf{Layout:} Success rate (\%) \\\textbf{Text:} Flex \ac{ca} (\%)
  \end{tabular} & \begin{tabular}{c}78.4 \\75.4
  \end{tabular}\\
   
  \hline


  \textbf{\begin{tabular}{l}ICDAR 2017\\ Historical-WI\\ \cite{8270156} - \ref{sssec:Historical-WI}\end{tabular}} &
  \begin{tabular}{c}  1,182 train images\\3,600 test images\\\end{tabular} & 
  \begin{tabular}{c} 394 train writers\\720 test writers\\\end{tabular}& \begin{tabular}{c}Image retrieval \\based on writer\\identification
  \end{tabular} & \begin{tabular}{c}German \\French\\Arabic
  \end{tabular} & \begin{tabular}{c}Handwritten document \\pages from  \\\nth{13} to \nth{20} century\\originating from\\Universitätsbibliothek Basel
  \end{tabular} & \begin{tabular}{c}Color\\Binary 
  \end{tabular} & 300 dpi
   & \begin{tabular}{c}JPG\\PNG
  \end{tabular} & \begin{tabular}{c}Writer ID on\\image name
  \end{tabular}  & image file name
   & oBIFs \cite{7490135,NEWELL20142255} & mAP (\%) & 55.6\\
   
  \hline

  \end{tabular}}

  
\end{minipage}
\end{adjustbox}
\end{center}
\end{sidewaystable}

\restoregeometry

\newgeometry{width=3cm, height=27.5cm, left=8cm}
\thispagestyle{empty}
\begin{sidewaystable}
\small
\sidewaystablefn%
\begin{center}
\begin{adjustbox}{scale=0.86,center}
\begin{minipage}{\textheight}
\renewcommand\arraystretch{1.52}
  \ContinuedFloat
  \centering
  \addtolength{\tabcolsep}{-3.5pt}
  
  \caption{Historical document image datasets with information about statistics, classes, tasks, language, document type, input visual aspects, ground truth, and benchmarks present in their original papers or competitions. The datasets are presented in the same order as in Table \ref{tab:structure_content_class} (earliest to latest).}
  \scalebox{0.52}{
  \begin{tabular}{l c c c c c c c c c c c c c}
  \hline
  \toprule
  
  \multicolumn{6}{c}{\textbf{General Information}} & 
  \multicolumn{3}{c}{\textbf{Input}} &
  \multicolumn{2}{c}{\textbf{GT}} &
  \multicolumn{3}{c}{\textbf{Benchmark}}\\ 
  \cmidrule(lr){1-6}
  \cmidrule(lr){7-9}
  \cmidrule(lr){10-11}
  \cmidrule(lr){12-14}
  \begin{tabular}{l}\textbf{Dataset name}\\ \textbf{[reference] - section}\end{tabular}&\textbf{Statistics}&\textbf{Classes}&\textbf{Task}&\textbf{Language}& \textbf{Document type} & 
  \textbf{Mode} &
  \textbf{Resolution} &
  \textbf{Format}& \textbf{Annotation} & \textbf{Format} & \textbf{Model} & \textbf{Metric} & \textbf{Performance}\\
  
  \bottomrule
  \hline

   \textbf{\begin{tabular}{l}Kuzushiji\\ \cite{Clanuwat2018DeepLF} - \ref{sssec:K-MNIST}\end{tabular}} &
  \begin{tabular}{l}\textbf{K-MNIST:} 60K train - 10K\\ test images\\
  \textbf{K-49:} 232,365 train -\\ 38,547 test images\\
  \textbf{K-Kanji:} 140,426 images\end{tabular} & 
  \begin{tabular}{l}\textbf{K-MNIST:} 10 character classes\\
  \textbf{K-49:} 49 character classes\\
  \textbf{K-Kanji:} 3,832 character classes\end{tabular} & \begin{tabular}{c}Kuzushiji isolated \\character recognition
  \end{tabular} & \begin{tabular}{c}Kuzushiji \\(cursive Japanese)
  \end{tabular} & \begin{tabular}{c}Character images from\\ scanned documents
  \end{tabular} & Grayscale & \begin{tabular}{c}28$\times$28\\64$\times$64\\pixel  resolution 
  \end{tabular}
   & PNG & \begin{tabular}{c}Character-level
  \end{tabular}  & \begin{tabular}{c}MNIST \\NumPy\\ formats
  \end{tabular}
   &  \begin{tabular}{l}\\4-NN\\2-layer CNN\\ResNet-18 \cite{10.1007/978-3-319-46493-0_38}\\ResNet-18 + input mixup \cite{Zhang2018mixupBE}\\ResNet-18 + manifold mixup \cite{verma2018manifold}
  \end{tabular}  & \ac{acc} (\%) & \begin{tabular}{c c}K-MNIST & K-49\\ \hline 91.56 & 86.01\\95.12 & 89.25\\97.82 & 96.64\\98.41 & 97.04\\ 98.83 & 97.33
  \end{tabular} \\
   
  \hline

  \textbf{\begin{tabular}{l}READ-BAD\\ \cite{8395221} - \ref{sssec:read-bad}\end{tabular}} & \begin{tabular}{c}
    2,036 pages\\ 132,124 baselines\\
  \end{tabular} & \begin{tabular}{c}Layout classes \\for text regions, \\e.g. paragraphs\end{tabular} & Baseline detection & Latin & \begin{tabular}{c}European archival\\documents from\\ 1470-1930 \end{tabular}& Color & N/A & JPG &\begin{tabular}{c}
   Text region and\\baseline information\\
  \end{tabular} & PAGE XML & \begin{tabular}{c}U-Net based\\ network (DMRZ) \cite{8270153}\end{tabular} & \begin{tabular}{c}\\P-value \\R-value \\F-value
  \end{tabular} & \begin{tabular}{cc} A: Simple & B: Complex\\
  \hline 0.973 & 0.854 \\0.970 & 0.863 \\0.971 & 0.859\\
  \end{tabular} \\
  \hline
 
  \textbf{\begin{tabular}{l}Warped Arabic\\ \cite{iet:/content/conferences/10.1049/cp.2018.1286} - \ref{sssec:warped_arabic}\end{tabular}} & \begin{tabular}{c}
    200 pages
  \end{tabular} & \begin{tabular}{c}4 centuries\\4 document types: \\book page, newspaper,\\ legal document, journal, other\\ document, unclassified\end{tabular} & \begin{tabular}{c}Text line\\ segmentation\end{tabular} & Arabic & \begin{tabular}{c}Arabic historical \\documents from the \\ \nth{16}-\nth{19} century with \\curls and warping\\from 4 document types\end{tabular} & Color & 350 dpi & \begin{tabular}{c}TIFF\\JPG\end{tabular}  &\begin{tabular}{c}
   Text line regions\\Metadata: author, title, date\\location, document type\\page number, language, script, \\font, number of columns
  \end{tabular} & PAGE XML & \begin{tabular}{c}Warping percentage\\\hline Voronoi diagrams\\ Smearing method\\Hybrid approach\\Projection profiles\end{tabular} & \begin{tabular}{c}Success rate (\%)
  \end{tabular} & \begin{tabular}{ccc} 0\% & 25\% & 50\%\\
  \hline  97.5& 91.5& 77.5\\ 94.9&85.9 &62.9 \\93.2 &82.2 &51.2\\90.9 & 75.9 &45.9
  \end{tabular} \\
  \hline

  \textbf{\begin{tabular}{l}MHDID\\ \cite{8480372} - \ref{sssec:mhdid}\end{tabular}} & 335 images & \begin{tabular}{c}
    Paper translucency\\ Stain\\ Reader's annotations \\ Worn holes
  \end{tabular} &  \begin{tabular}{c}
    Distortion classification\\ Visual quality\\ evaluation
  \end{tabular}& Arabic & \begin{tabular}{c}
       Book pages edited  \\
       from \nth{1} - \nth{14}\\ Islamic Centuries
  \end{tabular}  & Color & $1024\times 1280$ & JPG & \begin{tabular}{c}
       Image comparison: \\
       ">" left image better \\
       "<" right image better \\
       "=" similar images
  \end{tabular} & \begin{tabular}{c}Through\\ interface\end{tabular}&N/A&N/A&N/A\\
  \hline


  \textbf{\begin{tabular}{l}Tripitaka Koreana\\ in Han (TKH)\\ \cite{Yang2018DenseAT} - \ref{sssec:tripitaka}\end{tabular}} &
  \begin{tabular}{c} 1K images \\23,471 lines\\323,491 characters\end{tabular} & \begin{tabular}{c} 1,471 character\\ classes\end{tabular} & 
  \begin{tabular}{c} Character detection\\Character recognition \end{tabular} &  Chinese&  \begin{tabular}{c}Chinese historical\\ documents\end{tabular} & N/A & N/A & N/A &\begin{tabular}{c} Character-level\\bounding boxes\\and character label \end{tabular} & N/A
   & \begin{tabular}{c} RGD-VGG16\\text line input \end{tabular} & \begin{tabular}{c}\\ P\\R\\F \end{tabular} & 
   \begin{tabular}{cc} IoU:0.6 & IoU:0.7\\\hline 97.64 & 95.49\\96.39 & 94.56\\97.01 & 94.98
   \end{tabular}\\
  \hline
  
  \textbf{\begin{tabular}{l}Multiple Tripitaka\\ in Han (MTH)\\ \cite{Yang2018DenseAT} - \ref{sssec:tripitaka}\end{tabular}} &
  \begin{tabular}{c} MTH\\500 images\\17178 lines\\
  197886 characters \end{tabular} & \begin{tabular}{c} 3664 Character\\ classes\end{tabular} & 
  \begin{tabular}{c} Character detection\\Character recognition \end{tabular} &  Chinese&  \begin{tabular}{c}Chinese historical\\ documents\end{tabular} & N/A & N/A & N/A &\begin{tabular}{c} Character-level\\bounding boxes\\and character label \end{tabular} & N/A
   & \begin{tabular}{c} RGD-VGG16\\text line input \end{tabular} & \begin{tabular}{c}\\ P\\R\\F \end{tabular} & 
   \begin{tabular}{cc} IoU:0.6 & IoU:0.7\\\hline 96.44 & 92.17\\94.61 & 90.42\\95.52 & 91.29
   \end{tabular}\\
  \hline

  \textbf{\begin{tabular}{l}KERTAS \\\cite{adam2018kertas} - \ref{sssec:kertas}\end{tabular}} & \begin{tabular}{c}2,502 images\\135 books\end{tabular} & \begin{tabular}{c}14 islamic\\ centuries\end{tabular} & 
  \begin{tabular}{c}Age detection\\Writer identification\end{tabular} 
  & Arabic & \begin{tabular}{c}Arabic manuscripts\\spanning 14\\islamic centuries\end{tabular}  & Color & High-resolution & N/A&
  \begin{tabular}{c}Date, Source\\ Writer, ID\\ Manuscript name\\Description\end{tabular}  & \begin{tabular}{c}Sub-directories\\XML\end{tabular}& 
  \begin{tabular}{c}Splits (Train/test)\\\hline Proposed ($50\times50$)\\Run Length \cite{4107573}\\ Edge Direction \cite{BRINK2012162}\\ Edge Hinge \cite{10.1016/j.patrec.2013.03.020}\end{tabular}
  & \ac{acc} (\%)&\begin{tabular}{cc} Pre-defined &Random \\\hline 94.77 & 42.51\\88.57 & 85.71\\70.48 &66.66\\73.33 & 71.40\end{tabular} \\
  \hline

  \textbf{\begin{tabular}{l}ICFHR18\\ RASM2018 \\\cite{8583806} - \ref{sssec:rasm2018}\end{tabular}} &
  \begin{tabular}{c}10-15 train images\\50-80 test images\end{tabular}& \begin{tabular}{c}Text\\Graphic\\Text line\end{tabular} & 
  \begin{tabular}{c}Page segmentation\\Text line detection\\\ac{OCR}\end{tabular} 
  & Arabic & \begin{tabular}{c}Arabic scientific \\manuscripts from \nth{8}-\nth{9}\\ centuries CE\end{tabular}  & Color & N/A & TIFF &
  \begin{tabular}{c}Page-level: polygon text \\regions, paragraphs, graphics/\\line drawing, and text lines\\Transcriptions\end{tabular}  & \begin{tabular}{c}PAGE XML\end{tabular}& \begin{tabular}{c}\textbf{Page:} FCN (patches)\\\textbf{Text-line:} RDI \ac{OCR}\\\textbf{OCR:} RDI OCR\end{tabular} & 
  \begin{tabular}{c}Success rate (\%)\\Success rate (\%)\\Flex char acc (\%)\end{tabular} & \begin{tabular}{c}87.9\\81.6\\85.4\end{tabular}\\
  \hline

  
  \textbf{\begin{tabular}{l}ICFHR18 \\Asian Palm Leaf\\\cite{8583808} - \ref{sssec:ICFHR18_ASIAN_PALM_LEAF}\end{tabular}} &\begin{tabular}{c}\textbf{Bin/tion:} 50 train-50 test Balinese, 23 train-23\\ test Khmer, 31 train-30 test Sundanese images\\\textbf{Text line:} 47 train-49 test Balinese, 50 train-200\\ test Khmer, 31 train-30 test Sundanese images\\\textbf{OCR:} 11K train-7K test Balinese, 113K train-90K\\ test Khmer, 4.5K train-2.8K test Sundanese images\\
  \textbf{Translit.:} 15K train-10K test Balinese, 16K-7.7K test\\Khmer, 1.4K train-318 test Sundanese images\end{tabular}
  & \begin{tabular}{c} 133 Balinese,\\111 Khmer,\\ and 60 Sundanese\\character classes\end{tabular}& \begin{tabular}{c}Binarization\\Text line segmentation\\ Character recognition\\Transliteration \end{tabular} & \begin{tabular}{c}Balinese\\Khmer\\Sundanese\\Latin  \end{tabular} & \begin{tabular}{c}Manuscripts\\ \end{tabular}&  Color & N/A & \begin{tabular}{c}PNG\\TIFF\end{tabular} & \begin{tabular}{c}Binarized images\\Text line polygons
  \\Character-level\\\end{tabular} & \begin{tabular}{c}TXT \& BMP\\PNG\\TXT\end{tabular} & \begin{tabular}{c}\textbf{A:} Gaussian+non-linear enhance\\\textbf{B:} Horizontal projection profile\\\textbf{C:} Dense CNN\\\textbf{D:} CNN-RNN encoder-decoder\end{tabular} & \begin{tabular}{c}FM - NRM - PSNR\\DR - RA - FM (\%)\\Rec rate(\%)\\\ac{CER} (\%)\end{tabular} & 
  \begin{tabular}{c}58.87 - 0.17 - 28.71\\75.68 - 61.00 - 67.55 \\ 
  91.97\\ 5.62
  \end{tabular} \\

  \hline
  
  
  \textbf{\begin{tabular}{l}OHG\\  \ref{sssec:OHG}\end{tabular}} & \begin{tabular}{c}596 pages, 23,700 \\text-lines, 2,400 words\end{tabular} & \begin{tabular}{c}6 layout regions \end{tabular} & \begin{tabular}{c}Layout analysis\\
            Handwritten text recognition
       \end{tabular} & \begin{tabular}{c}Spanish\end{tabular} & \begin{tabular}{c} Spanish notarial \\deeds of the \nth{19} century
       \end{tabular}  &\begin{tabular}{c} Color\
       \end{tabular} & \begin{tabular}{l}300 dpi\end{tabular} & TIF & \begin{tabular}{c}Region coordinates \end{tabular} & PAGE XML & N/A& N/A & N/A\\
  \hline

  \hline
  
  
  \textbf{\begin{tabular}{l}ARDIS\\ \cite{kusetogullari2020ardis} - \ref{sssec:ardis}\end{tabular}} & \begin{tabular}{l}\textit{Dataset I}: 10K 4-digit images
  \\\textit{Dataset II - IV}: 7.6K digit images\\
   (6.6K train/1K test)\end{tabular} & \begin{tabular}{c}\textit{Dataset I}: 75 year 
   classes\\
  \textit{Dataset II - IV}: 10 classes (0-9) \end{tabular} & \begin{tabular}{c}
            Handwritten digit\\
        recognition
       \end{tabular} & \begin{tabular}{c}Swedish\\Latin\end{tabular} & \begin{tabular}{c} Swedish handwritten \\ document crops of\\digits from 1800 - 1940
       \end{tabular}  &\begin{tabular}{c} Color\\ Grayscale 
       \end{tabular} & \begin{tabular}{l}\textit{Dataset I}: 175$\times$95 pixels
  \\\textit{Dataset IV}: 28$\times$28 pixels\end{tabular} & JPG & \begin{tabular}{c} Images organized in \\ labeled folders\\ Binary in 0-9\end{tabular} & CSV & \begin{tabular}{c}CNN\\SVM\\HOG-SVM\\kNN\\Random forest\\RNN \end{tabular}& \ac{RA} (\%) & \begin{tabular}{c}98.60\\92.40\\95.50\\89.60\\87.00\\91.12 \end{tabular}\\
  \hline

  
  \textbf{\begin{tabular}{l}Pinkas\\ \cite{8978129} - \ref{sssec:pinkas}\end{tabular}} & \begin{tabular}{c}30 pages\\1,013 lines\\13,744 words \end{tabular}& \begin{tabular}{c}3,117 train classes\\1,251 test classes\\main text, line, \\and word \\segmentation class\end{tabular} & \begin{tabular}{c} Word spotting \\Page segmentation \end{tabular}& Hebrew & \begin{tabular}{c}
      Manuscript records \\ of Jewish communities  \\
      in Europe from\\ 1500-1800
  \end{tabular} & Color & High-resolution  & JPG & \begin{tabular}{c}Page-level\\Line-level\\Word-level\end{tabular} & \begin{tabular}{c}PAGE\\XML\end{tabular} & \begin{tabular}{c}Siamese CNN \cite{Bromley1993SignatureVU}\\PHOCNet \cite{Sudholt2016PHOCNetAD}\\PHOCNet \cite{Sudholt2016PHOCNetAD} (One hot)\\Exemplar SVM \cite{Almazn2012EfficientEW}\end{tabular}& \begin{tabular}{c}mAP (\%)\end{tabular} & \begin{tabular}{c}61.5\\56.6\\53.3\\1.5\end{tabular}\\
  \hline

  

  \end{tabular}}

  
\end{minipage}
\end{adjustbox}
\end{center}
\end{sidewaystable}

\restoregeometry


\newgeometry{width=3cm, height=27.5cm, left=8cm}
\thispagestyle{empty}
\begin{sidewaystable}
\small
\sidewaystablefn%
\begin{center}
\begin{adjustbox}{scale=0.83,center}
\begin{minipage}{\textheight}
\renewcommand\arraystretch{1.6}
\ContinuedFloat
  \centering
  \addtolength{\tabcolsep}{-3pt}
  \caption{Historical document image datasets with information about statistics, classes, tasks, language, document type, input visual aspects, ground truth, and benchmarks present in their original papers or competitions. The datasets are presented in the same order as in Table \ref{tab:structure_content_class} (earliest to latest).}
  \scalebox{0.52}{
  \begin{tabular}{lccccccccccccc}
  
  \hline
  
  \toprule
  
  \multicolumn{6}{c}{\textbf{General Information}} & 
  
  \multicolumn{3}{c}{\textbf{Input}}&
  \multicolumn{2}{c}{\textbf{GT}}&
  \multicolumn{3}{c}{\textbf{Benchmark}} \\ 
  
  \cmidrule(lr){1-6}
  \cmidrule(lr){7-9}
  \cmidrule(lr){10-11}
  \cmidrule(lr){12-14}
  \begin{tabular}{l}\textbf{Dataset name}\\ \textbf{[reference] - section}\end{tabular}&\textbf{Statistics}&\textbf{Classes}&\textbf{Task}&\textbf{Language}& \textbf{Document type} & 
  \textbf{Mode} &
  \textbf{Resolution} &
  \textbf{Format}& \textbf{Annotation} & \textbf{Format} & \textbf{Model} & \textbf{Metric} & \textbf{Performance}\\
  
  \bottomrule
  \hline

  \textbf{\begin{tabular}{l}BADAM\\ \cite{Kiessling2019BADAMAP} - \ref{sssec:badam}\end{tabular}} & \begin{tabular}{c}400 images\\(320 train/80 test)\end{tabular} & N/A & Baseline detection & \begin{tabular}{c} Arabic\\Persian\end{tabular}
  & \begin{tabular}{c} Manuscripts of\\various topics\\and dates\end{tabular}  & Color & \begin{tabular}{c} 200, 300\\and 500 dpi\end{tabular} & PNG&\begin{tabular}{c} Dense\\pixel-labeled \\baselines\end{tabular} & \begin{tabular}{c}PAGE XML\\PATH \end{tabular}& C-BLLA (proposed) & \begin{tabular}{c}
       P-val\\R-val\\F-val
  \end{tabular}& \begin{tabular}{c}
       0.941\\0.901\\0.924
  \end{tabular}\\
  \hline

  
  \textbf{\begin{tabular}{l}HORAE\\\cite{10.1145/3352631.3352633} - \ref{sssec:horae}\end{tabular}} & 
  \begin{tabular}{c} 557 images, 797 pages\\843 text regions, 1112 line fillers\\ 12512 text-lines, 284 miniatures\\ 892 decorated and  118 illustrated borders\\ 2776 decorated initials, 551 simple initials\\ 22 historiated initials, 5 ornamentation\\4 music notations
       \end{tabular} & \begin{tabular}{c}Page classes: calendar,\\ miniature, miniature-and-\\text, text-with-miniature,\\ full-page text\end{tabular} & \begin{tabular}{c}Line detection\\Layout analysis\end{tabular} & Latin & \begin{tabular}{c}Pages from\\Books of Hours\end{tabular}  & Color & N/A &  JPG & \begin{tabular}{c}Layout and text\\ line-level with \\segmentation boxes\end{tabular} & \begin{tabular}{c} PAGE\\XML\end{tabular} & dhSegment \cite{Oliveira2018dhSegmentAG} & \begin{tabular}{l}IoU ($t=0.7$)\\+post-processing \\\hline IoU ($t=0.5$)\\+post-processing \end{tabular}&
       \begin{tabular}{l} Line detection: 0.84\\Line detection: 0.88\\\hline Layout analysis: 0.73\\Layout analysis: 0.75\end{tabular}\\
  \hline

  
  \textbf{\begin{tabular}{l}ICDAR19\\ cTDaR19\\\cite{8978120} - \ref{sssec:ctdar} \end{tabular}} & \begin{tabular}{c}\textbf{Track A:} 600 train,\\199 test archival images\\\textbf{Track B:} 600 train,\\150 test archival images\end{tabular} & \begin{tabular}{c}Archival\\Modern\end{tabular} & \begin{tabular}{c}Table detection\\Table recognition\end{tabular} & N/A &  \begin{tabular}{c} Images with tables \\from archival accounting \\books, record books,\\ timetables, etc.\end{tabular} & Color & N/A & JPG & \begin{tabular}{c}Bounding polygon coords\\Cell element attributes \end{tabular} & XML &
  \begin{tabular}{l} \textbf{Track A:} Faster-RCNN \cite{NIPS2015_14bfa6bb}\\\textbf{Track B:} FCN+CCA\end{tabular} & \begin{tabular}{c} \ac{WA}-F1 score\end{tabular} & \begin{tabular}{c}0.94\\ B1: 0.48, B2. 0.47\end{tabular}\\
  \hline

 
  \textbf{\begin{tabular}{l}ICDAR19 \\DMAS2019\\ \ref{sssec:DMAS2019}\end{tabular}} &
  \begin{tabular}{c} 50-100 PAGE\\ annotated images \end{tabular} &
  \begin{tabular}{c}\textbf{Page classes:} 
  cover, table\\ of contents, content, index\\
  \textbf{Article classes:} article,\\ illustration with caption, \\advertisement, index, colophon
  \end{tabular}& \begin{tabular}{c} Article segmentation\\
  Page segmentation\\
  Text recognition\end{tabular} &  N/A & \begin{tabular}{c} Digitised image pages \\from magazines from \\1800 - 1938 by the\\National Library of\\ the Netherlands\end{tabular} & Color & N/A & JP2 & \begin{tabular}{c} Layout annotations in\\ page and article-level\\ Transcriptions \end{tabular}& \begin{tabular}{c} PAGE XML\\ALTO \end{tabular}
   & N/A & N/A& N/A\\
  \hline

 
  \textbf{\begin{tabular}{l}ICDAR19 \\DIBCO 2019\\ \cite{8978205} - \ref{sssec:DIBCO_2019}\end{tabular}} &
  \begin{tabular}{c} 10 historical machine-printed\\ and handwritten test images\\10 papyri test images\end{tabular} &
  N/A& Binarization & N/A  & \begin{tabular}{c} Historical machine-printed\\ and handwritten images \\of the \nth{19} century\\and papyri images from \\various periods of Antiquity\end{tabular} & \begin{tabular}{c}Color\\Grayscale
  \end{tabular} & N/A & BMP & Image bitmasks & BMP
   &\begin{tabular}{c}Ensemble clustering\\ algorithms (Fuzzy C-Means,\\ K-Medoids, K-Means++)
  \end{tabular}  & \begin{tabular}{c}\ac{FM} (\%)\\$F_{ps}$ (\%)\\ PSNR\\DRD
  \end{tabular}& \begin{tabular}{c}72.875\\72.15\\ 14.475\\16.235 
  \end{tabular}\\
  \hline

  
  \textbf{\begin{tabular}{l}OBC306 \\\cite{8978032} -  \ref{sssec:obc306}\end{tabular}} & \begin{tabular}{c}4,024 distinct characters\\ 309,551 samples \end{tabular}& \begin{tabular}{c}306 character \\categories \end{tabular} & \begin{tabular}{c} Oracle-bone\\character recognition\end{tabular} & Oracle
  & \begin{tabular}{c}Patch samples\\derived from \\oracle-bone\\publications\end{tabular}& Grayscale& \begin{tabular}{c} 1:2 ratio\\Height: 50-150 pixels\\Width: 0-100 pixels\end{tabular} & N/A & Character-level &\begin{tabular}{c}6-digit\\ encoding  \end{tabular}& \begin{tabular}{c} Top-K\\ \hline HOG+SVM \cite{1467360}\\AlexNet \cite{NIPS2012_c399862d}\\VGG16 \cite{Simonyan2015VeryDC}\\ResNet-50 \cite{He2016DeepRL}\\ResNet-101 \cite{He2016DeepRL}\\Inception-v4 \cite{7298594}\end{tabular} & \begin{tabular}{c}\ac{acc} (\%)\end{tabular}& \begin{tabular}{cccc}Top-1 & Top-3 & Top-5 & Top-10\\\hline 4.29 & 25.93 & 32.15 & 41.26\\ 66.75 & 75.31 & 77.46 & 80.56\\ 67.20 & 76.21 & 78.57 & 81.51 \\ 69.09 & 78.06 & 80.50 & 83.00\\
  69.50 & 77.92 & 80.00 & 81.66\\70.28 & 78.74 & 80.24 & 82.28\end{tabular}\\
  \hline

  
  \textbf{\begin{tabular}{l}GRK-Papyri \& PapyRow\\\cite{8978142} \& \cite{10.1007/978-3-030-68787-8_16} - \ref{sssec:grk-papyri}\end{tabular}} 
  & \begin{tabular}{c}\textbf{GRK-Papyri:} 50 images\\\textbf{PapyRow:} 6,498 samples\end{tabular} & 10 writers & \begin{tabular}{c} Writer identification\\Image enhancement\\Binarization\\Line/word segmentation\end{tabular} & Greek
  & \begin{tabular}{c} Handwritten\\ Greek papyri\\ from the \nth{6}\\ century A.D.\end{tabular}
   & \begin{tabular}{c}Color\\Grayscale \end{tabular} & \begin{tabular}{c}\textbf{GRK-Papyri:} 96-2000 dpi\\
  Height: 796-6818\\Width: 177-2000\\\textbf{PapyRow:} 1200 pixels width\\500 pixels width
  \end{tabular} & JPG & 
  \begin{tabular}{c}Name of \\writer in\\ file name\end{tabular}& File & NBNN \cite{mohammed2017normalised} & \begin{tabular}{c}Identification\\ rate (\%)\end{tabular}& \begin{tabular}{cc}Leave-one-out & Train-test\\ \hline 30.0 &26.6\end{tabular}\\
  \hline

  
  \textbf{\begin{tabular}{l}CASIA-AHCDB \\
   \cite{8978010} - \ref{sssec:casia-ahcdb} \end{tabular}} 
   & \begin{tabular}{c} 11,937 images\\2.2M characters\end{tabular} & 
   \begin{tabular}{c}10,350 character\\categories\end{tabular} & \begin{tabular}{c}Character\\ recognition \end{tabular}& Chinese & \begin{tabular}{c}Chinese ancient\\handwritten document\\ pages from the\\ Complete
   Library in \\Four Sections (style-1)\\
   and the Ancient \\Buddhist Scriptures\\ (style-2)
   \end{tabular}& Grayscale & N/A & GNTX &\begin{tabular}{c}Background pixel \\labeling and\\ Chinese character\\ labels \end{tabular}& Unicode & \begin{tabular}{c}\\CNN\\CPN\end{tabular} & \begin{tabular}{c}Character \ac{RA} (\%) \\for BC \& EC\end{tabular} & \begin{tabular}{cc}style-1 &style-2 \\\hline96.93 & 92.08\\96.95 & 91.99\end{tabular}\\
  \hline
    
  
  \textbf{\begin{tabular}{l}Amharic\\Database\\\cite{8977980} - \ref{sssec:amharic} \end{tabular}} & \begin{tabular}{c}Character dataset: 80K \\character images, 231 characters\\
  Text-line dataset: 40,929 printed\\ Power Geez text-line images, 197,484\\ synthetic Power Geez text-line\\ images, 98,924 synthetic Visual\\ Geez text-line images, 280\\ unique Amharic characters \end{tabular} & \begin{tabular}{c}280 unique\\ Amharic characters\end{tabular} & \begin{tabular}{c}
        Text-line image\\
        recognition,\\OCR, Page binarization\\Line segmentation
       \end{tabular} & Amharic & \begin{tabular}{c} Printed, synthetic\\ Amharic socuments
       \end{tabular}  & Grayscale & \begin{tabular}{c}300 dpi\\Characters:\\ $32\times32$ pixels\\Text-lines:\\ $48\times128$ pixels\end{tabular} & PNG & \begin{tabular}{c} Character-level\\Text line-level\end{tabular} & \begin{tabular}{c} TXT\\NPY\end{tabular} & BLSTM-CTC & \begin{tabular}{c} \ac{CER} (\%)\\on text-line\\ dataset\end{tabular} & \begin{tabular}{l}Printed - Power Geez: $ 8.54$\\Synth - Power Geez:  $4.24$\\Synth - Visual Geez:  $2.28$\end{tabular}\\
  \hline

  \textbf{\begin{tabular}{l}Multiple\\Font \\Groups \\\cite{10.1145/3352631.3352640} - \ref{sssec:fonts}
  \end{tabular}} & 35,623 images &\begin{tabular}{c}
    Textura, Rotunda\\ Gotico-antiqua, Hebrew\\ Bastarda, Schwabacher \\ Fraktur, Antiqua\\
    Italic, Greek\\ Other Font, Not a Font
  \end{tabular} & Font recognition & \begin{tabular}{c}Latin, Greek\\Hebrew, and more \end{tabular} & \begin{tabular}{c}Pages from early\\ printed books \\from \nth{15}-\nth{18}\\ centuries\end{tabular} & \begin{tabular}{c}Color \\Grayscale\end{tabular} & \begin{tabular}{c}side length: 79-14K pixels\\
  median surface:\\ $5.3\times10^6$ pixels\end{tabular}& \begin{tabular}{c}JPG \\TIFF\end{tabular} & \begin{tabular}{c}Font name \\next to \\image name\end{tabular}& CSV & \begin{tabular}{c}
    ResNet-50 \cite{He2016DeepRL}\\ResNet-18 \cite{He2016DeepRL}\\VGG-16 \cite{Simonyan2015VeryDC}\\DenseNet-121 \cite{huang2018densely}
  \end{tabular} & \begin{tabular}{c}
     mIoU $\pm$ std. dev. (\%) 
  \end{tabular} & \begin{tabular}{c}
    $82.51\pm$ 0.15\\83.34$\pm$ 0.19\\83.44$\pm$ 0.53\\84.06$\pm$ 0.23
  \end{tabular}\\
  \hline

  \end{tabular}}

  
\end{minipage}
\end{adjustbox}
\end{center}
\end{sidewaystable}

\restoregeometry


\newgeometry{width=3cm, height=29.7cm, left=8.2cm}
\thispagestyle{empty}
\begin{sidewaystable}
\small
\sidewaystablefn%
\begin{center}
\begin{adjustbox}{scale=0.8,center}
\begin{minipage}{\textheight}
\renewcommand\arraystretch{1.62}
\ContinuedFloat
  \centering
  \addtolength{\tabcolsep}{-3pt}
  \caption{Historical document image datasets with information about statistics, classes, tasks, language, document type, input visual aspects, ground truth, and benchmarks present in their original papers or competitions. The datasets are presented in the same order as in Table \ref{tab:structure_content_class} (earliest to latest).}
  \scalebox{0.52}{
  \begin{tabular}{lccccccccccccc}
  
  \hline
  
  \toprule
  
  \multicolumn{6}{c}{\textbf{General Information}} & 
  
  \multicolumn{3}{c}{\textbf{Input}}&
  \multicolumn{2}{c}{\textbf{GT}}&
  \multicolumn{3}{c}{\textbf{Benchmark}} \\ 
  
  \cmidrule(lr){1-6}
  \cmidrule(lr){7-9}
  \cmidrule(lr){10-11}
  \cmidrule(lr){12-14}
  \begin{tabular}{l}\textbf{Dataset name}\\ \textbf{[reference] - section}\end{tabular}&\textbf{Statistics}&\textbf{Classes}&\textbf{Task}&\textbf{Language}& \textbf{Document type} & 
  \textbf{Mode} &
  \textbf{Resolution} &
  \textbf{Format}& \textbf{Annotation} & \textbf{Format} & \textbf{Model} & \textbf{Metric} & \textbf{Performance}\\
  
  \bottomrule
  \hline


  \textbf{\begin{tabular}{l}ICDAR19\\ HDRC\\Chinese DB\\ \cite{8977999} - \ref{sssec:chinese_reading_challenge}\end{tabular}} &
  \begin{tabular}{c} 11,715 train images\\
  1,135 test images
  \end{tabular} & 
  \begin{tabular}{c} Text\\Non-text (background) \end{tabular} & 
  \begin{tabular}{c}Text recognition on\\ extracted lines\\
  Layout analysis (pixel-level)\\
  Text line detection \& \\ recognition
  \end{tabular}&  \begin{tabular}{c}Chinese
  \end{tabular} & \begin{tabular}{c}Historical Chinese\\family records pages
  \end{tabular} & Grayscale & Various & \begin{tabular}{c}PNG\\
  JPG
  \end{tabular} & \begin{tabular}{c}Pixel-level\\Layout-level\\
  Text line-level
  \end{tabular} & PAGE XML
   & \begin{tabular}{c}CRNN \cite{Shi2017AnET} for text \\recognition, Cascade R-CNN \cite{Cai2018CascadeRD}\\ for text-line detection, and U-Net \\
  for pixel-wise classification
  \end{tabular} & \begin{tabular}{c}editDistance\\IoU (\%)\\totalErrors\end{tabular} & \begin{tabular}{c}2539\\99.54\\5557\end{tabular}\\
  
  \hline


  \textbf{\begin{tabular}{l}ICDAR 2019\\REID 2019\\ \cite{8978191} - \ref{sssec:reid2019}\end{tabular}} &
  \begin{tabular}{c} 25 annotated \\train images,
  \\56 test images \end{tabular} & 
  \begin{tabular}{c}Text\\Separator\\Graphic\\Image \end{tabular}& 
  \begin{tabular}{c}Layout analysis\\Text recognition
  \end{tabular} & \begin{tabular}{c}Bengali\\English 
  \end{tabular} & \begin{tabular}{c}Scanned images from \\printed books in\\ Bengali from\\ 1713-1914
  \end{tabular} & Color & N/A
   & TIFF &  \begin{tabular}{c}Page-level\\Transcriptions
  \end{tabular} & PAGE XML
   & \begin{tabular}{c}Google Multi-\\Lingual OCR
  \end{tabular}   & \begin{tabular}{c}Layout: Success rate (\%) \\Text: Flex character \ac{acc} (\%)
  \end{tabular} & \begin{tabular}{c}80.4\\77.68
  \end{tabular}\\
  
  \hline

  \textbf{\begin{tabular}{l}ABP \& NAF\\ \cite{8978117} - \ref{sssec:ABP_NAF}\end{tabular}} & \begin{tabular}{c}ABP small: 180 pages\\
  ABP large: 1,098 pages\\ NAF: 488 pages
       \end{tabular} & \begin{tabular}{c}Text region\\ Table region
       \end{tabular} & \begin{tabular}{c}Table detection\\ and partition
       \end{tabular} & N/A & \begin{tabular}{c}Death, birth,\\ marriage, and\\ tax records \end{tabular} & \begin{tabular}{c} Color\\Grayscale\end{tabular} & N/A & JPG & \begin{tabular}{c}Page layout\\\end{tabular} &  \begin{tabular}{c}PAGE XML\end{tabular} &
        \begin{tabular}{c} ECN - 8 layers \cite{8978117} \end{tabular} 
        & F1 (\%) & \begin{tabular}{cccc}&ABP small & ABP large & NAF\\
        rows& 99.7 &98.2 & 98.6\\
        columns & 99.8 & 98.6 & 99.0\\
        cells & 99.1 & 99.1 & 99.5\end{tabular}\\
  \hline

  \textbf{\begin{tabular}{l}FCR\\ \cite{Quirs2020FinnishCR} - \ref{sssec:FCR}\end{tabular}} & 
  500 pages & 6 layout regions & \begin{tabular}{c}Layout analysis\\HTR\end{tabular} & Swedish & \begin{tabular}{c}Pages of records of deeds,\\ mortgages, traditional life-annuity,\\ from the Renovated District\\ Court Records (\nth{19} century)\end{tabular} & Grayscale  & N/A  & & \begin{tabular}{c}Layout regions\\Baselines\\Line-level transcriptions\end{tabular} & PAGE XML &
 N/A & N/A & N/A\\
  \hline
  
  \textbf{\begin{tabular}{l}IlluHisDoc\\ \cite{monnier2020docExtractor} - \ref{sssec:illuhisdoc}\end{tabular}} & 400 images & \begin{tabular}{c}Illustration\\
        Text
       \end{tabular}& \begin{tabular}{c} Baseline detection\\
        Illustration segmentation
       \end{tabular} & N/A & \begin{tabular}{c}Document images\\from Gallica\\with various\\ illustrations\end{tabular} & \begin{tabular}{c} Color\\
        Grayscale\end{tabular} & N/A & JPG & \begin{tabular}{c}Segmentation \\masks\end{tabular} &  \begin{tabular}{c}PNG \\JSON\end{tabular} &
        \begin{tabular}{c}Tesseract4\footref{tesseract}\\Mask-RCNN \cite{he2017mask} (PubLay.)\\Proposed (PubLay.) \\Mask-RCNN \cite{he2017mask} (SynDoc)\\Proposed (SynDoc)\end{tabular} 
        & mIoU (\%) & \begin{tabular}{c}14.8\\11.5\\24.3\\55.4\\76.1\end{tabular}\\
  \hline


  \textbf{\begin{tabular}{l}Newspaper\\Navigator\\ \cite{10.1145/3340531.3412767} - \ref{sssec:newspaper_navigator}\end{tabular}} & \begin{tabular}{c}3,559 pages\\ 48,409 annotations\end{tabular}& \begin{tabular}{c}Photograph, Illustration\\
  Map, Comic/Cartoon\\Editorial cartoon\\
  Headline, Advertisement
  \end{tabular}& \begin{tabular}{c}Visual content\\ recognition\\
  \ac{OCR}
  \end{tabular}& English & \begin{tabular}{c}Historical newspaper\\ pages from \\the Chronicling \\America corpus\\\end{tabular} & Grayscale & N/A & JPG & \begin{tabular}{c}Bounding boxes\\Metadata\\Transcriptions
  \end{tabular}&
  \begin{tabular}{c}JSON (COCO)\\CSV
  \end{tabular} & \begin{tabular}{c}Faster-RCNN \cite{NIPS2015_14bfa6bb}\\with R50-FPN\\ backbone\end{tabular} & \begin{tabular}{c}mAP (\%)\end{tabular} & 63.4\\
  
  \hline

  \textbf{\begin{tabular}{l}DIDA\\
  \cite{KUSETOGULLARI2021100182} - \ref{sssec:dida}\end{tabular}} & \begin{tabular}{c} \textbf{Dataset I:} 250K single digits\\
        \textbf{Dataset II:} 200K multi-digits \\\textbf{Dataset III:} 25K bounding boxes
       \end{tabular} & \textbf{Dataset I:} 0 - 9 & \begin{tabular}{c}
            Handwritten digit\\
        recognition and detection
       \end{tabular} & Swedish & \begin{tabular}{c} Swedish handwritten \\ document crops \\ from 1800 - 1940
       \end{tabular}  & Color & \begin{tabular}{c} $240\times 210$ \\Pixels \end{tabular}& JPG & \begin{tabular}{c}Images organized in \\labeled folders \end{tabular} & Folders &\begin{tabular}{c}DIGITNET-dect \cite{KUSETOGULLARI2021100182} \end{tabular} &\begin{tabular}{c}\ac{DR} (\%)\end{tabular}&\begin{tabular}{c}75.96\end{tabular} \\
  \hline

  
  \textbf{\begin{tabular}{l}ScribbleLens\\\cite{9257750} - \ref{sssec:scribblelens}\end{tabular}} & \begin{tabular}{c}1K pages \\28,255 lines \\281,914 characters\\85 writers\end{tabular} & \begin{tabular}{c}Writter-id and\\150 years of origin\\ classes\end{tabular}& \begin{tabular}{c}Automatic transcription\\Handwriting recognition\end{tabular} & Dutch & \begin{tabular}{c} Early Modern \\ Manuscripts \end{tabular} & \begin{tabular}{c} Color \\ B/W\\Grayscale \end{tabular} & \begin{tabular}{c}150 - 300\\ dpi\end{tabular} &JPG & \begin{tabular}{c}Transcriptions\\ Writter-id\\Year of origin\end{tabular}& \begin{tabular}{c}UTF8\\TXT\end{tabular}  & CNN/BLSTM/CTC \cite{7814068, Nina2018NephiA} & \begin{tabular}{c}\ac{CER} (\%)\end{tabular} & $11.9 \pm 0.68$\\
  \hline

  \textbf{\begin{tabular}{l}ICDAR19\\ RASM2019\footref{rasm2019} \\ \ref{sssec:rasm2019}\end{tabular}} &
  \begin{tabular}{c}20 train images\\100 test images\end{tabular}& \begin{tabular}{c}Text\\Graphic\\Text line\end{tabular} & 
  \begin{tabular}{c}Page segmentation\\Text line detection\\OCR\end{tabular} 
  & Arabic & \begin{tabular}{c}Arabic scientific \\manuscripts from \\\nth{9}-\nth{19} CE\end{tabular}  & Color & N/A & TIFF &
  \begin{tabular}{c}Page-level: polygon text \\regions, paragraphs, graphics/\\line drawing, and text lines\\Transcriptions\end{tabular}  & \begin{tabular}{c}PAGE XML\end{tabular}& \begin{tabular}{c}\textbf{Page:} Google\\\textbf{Text-line:} RDI OCR\\\textbf{OCR:} RDI OCR\end{tabular} & 
  \begin{tabular}{c}Success rate (\%)\\Success rate (\%)\\Flex \ac{ca} (\%)\end{tabular} & \begin{tabular}{c}69.30\\77.60\\77.58\end{tabular}\\
  \hline

  
  \textbf{\begin{tabular}{l}ICDAR 2019\\HDRC-IR\\ \cite{Christlein2019ICDAR2C} - \ref{sssec:ICDAR2019-HDRC-IR}\end{tabular}} &
  \begin{tabular}{c} Train and validation\\ from \ref{sssec:Historical-WI}\\20K test images  \\\end{tabular} & 
  \begin{tabular}{c}10K writers\\Writers from manuscript \\books, letter, or charters
  \end{tabular}& \begin{tabular}{c}Image retrieval \\based on writer\\identification
  \end{tabular} & N/A & \begin{tabular}{c}Handwritten document \\pages of manuscript books,\\letters, charters and legal docs\\from  various institutions
  \end{tabular} & Color & \begin{tabular}{c}High and low\\quality factor\\resolution\\ 2K pixel\\larger dimension
  \end{tabular}
   & JPG & \begin{tabular}{c}Writer ID next \\to image name
  \end{tabular} & CSV
   & \begin{tabular}{c}SIFT \cite{LoweDavid2004DistinctiveIF} + Pathlet \cite{8978107} \\(SCUT submission) \end{tabular}& \begin{tabular}{c}\ac{acc} (\%)\\mAP (\%) \end{tabular}&\begin{tabular}{c}97.4\\92.5 \end{tabular} \\
  
  \hline

  
  \textbf{\begin{tabular}{l}HTR Benchmarks\\ \cite{SANCHEZ2019122} - \ref{sssec:htr_benchmarks}\end{tabular}} &
  \begin{tabular}{c} \textbf{ICFHR-2014:} 433 pages, 11,473 lines\\106K running words, 550K characters\\
  \textbf{ICDAR-2015:} 796 pages, 21,752 lines\\
  186K running words, 955K characters\\
  \textbf{ICFHR-2016:} 450 pages, 10K lines\\
  43K running words, 260K characters\\
  \textbf{ICDAR-2017:} 10K pages, 206K lines\\
  1.7M running words, 8M characters\end{tabular} & 
  \begin{tabular}{c}\textbf{ICFHR-2014:} 9K lexicon, 86 character set\\
  \textbf{ICDAR-2015:} 17K lexicon, 87 character set\\
  \textbf{ICFHR-2016:} 8K lexicon, 92 character set\\
  \textbf{ICDAR-2017:} 4K lexicon, 104 character set
  \end{tabular}& \begin{tabular}{c}Handwritten text\\recognition 
  \end{tabular} & \begin{tabular}{c}English\\Early modern German 
  \end{tabular} & \begin{tabular}{c}Handwritten document pages\\ from the Bentham and\\ the Ratsprotokolle\\ (1470-1805) collections
  \end{tabular} & Color & \begin{tabular}{c}300 dpi or\\ 75-300 dpi\end{tabular}
   & JPG & \begin{tabular}{c}
  Transcriptions and \\coordinates at line-level \end{tabular} & PAGE XML
   & \begin{tabular}{c} CRNN + N-gram LM\end{tabular}& \begin{tabular}{c}\ac{CER} (\%)\\\ac{WER} (\%) \end{tabular}&\begin{tabular}{ccc}&\ac{CER}&\ac{WER}\\
   ICFHR-2014 & 5.0 & 9.7\\
   ICDAR-2015 & 12.8 & 30.0\\
   ICFHR-2016 & 4.5 & 17.5\\
   ICDAR-2017 &5.8 & 17.6
   \end{tabular} \\
  
  \hline

  
  \textbf{\begin{tabular}{l}HJDataset\\\cite{Shen_2020_CVPR_Workshops} - \ref{sssec:hjdataset}\end{tabular}} & \begin{tabular}{c} 2,271 Images\\
        25K Elements
       \end{tabular} & \begin{tabular}{c} Page frame\\ Row\\ 
       Title \& Text region \\ Title, Subtitle\\ Other
       \end{tabular} & \begin{tabular}{c} Layout analysis\\
        Element extraction
       \end{tabular} & Japanese &  Biography scans  & Color & N/A & JPG 
       & \begin{tabular}{c} Page frame\\ Row region\\Text block boxes\\ Reading order\\ Hierarchical dependencies\end{tabular}& \begin{tabular}{c}JSON \\(COCO)\end{tabular} & \begin{tabular}{l}Faster-RCNN \cite{NIPS2015_14bfa6bb}\\Mask-RCNN \cite{he2017mask}\\Retinanet\cite{lin2017focal}\end{tabular} & \begin{tabular}{c}\ac{mAP} (\%) \end{tabular}& \begin{tabular}{c}81.991\\81.343\\75.223\end{tabular}\\
  \hline

  
  \textbf{\begin{tabular}{l}ICFHR 2020\\ HisFragIR20\\ \cite{Seuret2020ICFHR2C} - \ref{sssec:HisFragIR20}\end{tabular}} &
  \begin{tabular}{c}  220K fragment images\\(101,706 train - 20,019 test)\end{tabular} & 
  \begin{tabular}{c} 9,800K writers\\(8,717 train - 1,152 test)\end{tabular}& \begin{tabular}{c}Image retrieval \\per writer and \\per image 
  \end{tabular} & \begin{tabular}{c}N/A
  \end{tabular} & \begin{tabular}{c}Manuscript pages\\from books of the \\European Middle Ages\\\nth{9} - \nth{15} century CE
  \end{tabular} & \begin{tabular}{c}Color
  \end{tabular} & \begin{tabular}{c}High and low\\quality factor\\resolution\\ 2K pixel\\larger dimension
  \end{tabular}
   & \begin{tabular}{c}JPG
  \end{tabular} &  \begin{tabular}{c}Writer ID, Page ID,\\ and Fragment ID\\on image name
  \end{tabular} & Image name
   & \begin{tabular}{c}
  ResNet50 \cite{He2016DeepRL} + $\chi^2$ distance\end{tabular} & \begin{tabular}{c}
  mAP (\%)\\\ac{acc} (\%)\\Pr@10 (\%)\\Pr@100 (\%)\end{tabular} &\begin{tabular}{cc}
  Writer & Image \\ \hline33.5 & 22.6\\77.1&36.4\\53.1&31.2\\50.4&58.9\end{tabular} \\
   
  \hline

  \end{tabular}}

  
\end{minipage}
\end{adjustbox}
\end{center}
\end{sidewaystable}

\restoregeometry


\newgeometry{width=3cm, height=29.7cm, left=8.2cm}
\thispagestyle{empty}
\begin{sidewaystable}
\small
\sidewaystablefn%
\begin{center}
\begin{adjustbox}{scale=0.8,center}
\begin{minipage}{\textheight}
\renewcommand\arraystretch{1.62}
\ContinuedFloat
  \centering
  \addtolength{\tabcolsep}{-2pt}
  \caption{Historical document image datasets with information about statistics, classes, tasks, language, document type, input visual aspects, ground truth, and benchmarks present in their original papers or competitions. The datasets are presented in the same order as in Table \ref{tab:structure_content_class} (earliest to latest).}
  \scalebox{0.52}{
  \begin{tabular}{lccccccccccccc}
  
  \hline
  
  \toprule
  
  \multicolumn{6}{c}{\textbf{General Information}} & 
  
  \multicolumn{3}{c}{\textbf{Input}}&
  \multicolumn{2}{c}{\textbf{GT}}&
  \multicolumn{3}{c}{\textbf{Benchmark}} \\ 
  
  \cmidrule(lr){1-6}
  \cmidrule(lr){7-9}
  \cmidrule(lr){10-11}
  \cmidrule(lr){12-14}
  \begin{tabular}{l}\textbf{Dataset name}\\ \textbf{[reference] - section}\end{tabular}&\textbf{Statistics}&\textbf{Classes}&\textbf{Task}&\textbf{Language}& \textbf{Document type} & 
  \textbf{Mode} &
  \textbf{Resolution} &
  \textbf{Format}& \textbf{Annotation} & \textbf{Format} & \textbf{Model} & \textbf{Metric} & \textbf{Performance}\\
  
  \bottomrule
  \hline
  
  
  \textbf{\begin{tabular}{l}ICDAR 2021\\ HDC \\ \cite{10.1145/3476887.3476913} - \ref{sssec:his_doc_classification_icdar_2021}\end{tabular}} &
  \begin{tabular}{l}\textbf{Font:} 35k train, 5,506 test images\\
  \textbf{Script:} CLaMM 2017 train, 1,256 test images\\
  \textbf{Date:} 11,294 train, 2,516 test images\\
  \textbf{Location:} 5,517 train, 60
  val, 300 test images\end{tabular} & 
  \begin{tabular}{c}\textbf{Font:} same as Section \ref{sssec:fonts}\\ without "other font" and\\ "not a font" classes (10 classes)\\ 
  \textbf{Script:} same as Section \ref{sssec:ICFHR16CLAMM}\\
  \textbf{Date:} date ranges \\\textbf{Location:} Cluny, Corbie, Citeaux, Florence,\\ Fonteney, Himanis, Milan, Paris, Signy\\  MontSaintMichel, SaintBertin\\ SaintGermainDesPres, SainMatrialDeLinoges
  \end{tabular}& \begin{tabular}{c} Font/script,\\ date, and location\\classification
  \end{tabular} & Latin
   & \begin{tabular}{c} Handwritten and\\
   printed page \\images in Latin 
  \end{tabular} & \begin{tabular}{c} Color\\Grayscale
  \end{tabular} &  \begin{tabular}{c} Various
  \end{tabular}
   &  \begin{tabular}{c}TIFF\\JPG
  \end{tabular} &  \begin{tabular}{c} Label next\\to image\\name
  \end{tabular} & CSV & \begin{tabular}{l}
  \textbf{Font:} ResNeXt-50 \cite{Xie2017AggregatedRT} on multi-scales\\
  \textbf{Script:} ResNeXt-50 \cite{Xie2017AggregatedRT} + linear interpolation\\
  \textbf{Date:} CNN on text lines\\
  \textbf{Location:} ResNeXt-50 \cite{Xie2017AggregatedRT} on multi-scales
  \end{tabular}  &  \begin{tabular}{c}
  Overall \ac{acc} (\%)\\Overall \ac{acc} (\%)\\MAE (\%)
  \\Overall \ac{acc} (\%)
  \end{tabular} & \begin{tabular}{c} 99.04\\88.77\\
  21.91\\79.69
  \end{tabular} \\

  \hline
  
  
  \textbf{\begin{tabular}{l}BIR Database\\ \cite{10.1145/3476887.3476913} - \ref{sssec:bir_database}\end{tabular}} &
  \begin{tabular}{c} 35 documents\\285 pages\\2,106 bold words \\5,745 italic words\\80,168 regular words\\880,169 total words\end{tabular} & 
  \begin{tabular}{c} Bold\\Italic\\Regular\end{tabular}& \begin{tabular}{c} Style classification\\Word detection
  \end{tabular} & \begin{tabular}{c} French\\Latin\\Other
  \end{tabular} & \begin{tabular}{c} Printed pages\\from sale catalogues\\and exhibitions\\from the \nth{19}\\ and \nth{20} centuries
  \end{tabular} & Color & N/A 
   & JPG  &  \begin{tabular}{c} Word bounding\\boxes and\\word font class\\annotations
  \end{tabular} & \begin{tabular}{c}XML\\HTML
  \end{tabular}
   & \begin{tabular}{l}\textbf{Word detection}: YOLOv5m \cite{glenn_jocher_2021_4418161}\\\textbf{Style classification}: MobileNetV2 \cite{sandler2019mobilenetv2}
  \end{tabular}  &  \begin{tabular}{c}F1-score\\ (TVT)\end{tabular}  & \begin{tabular}{c} 0.91\\ 0.90 
  \end{tabular} \\

  \hline
  
  \textbf{\begin{tabular}{l}GloSAT\\ \cite{10.1145/3476887.3476890} - \ref{sssec:GloSAT}\end{tabular}} &
  \begin{tabular}{c} 500 images \\\end{tabular} & 
  \begin{tabular}{c} Heading\\Header\\Table body\\\end{tabular}& \begin{tabular}{c} Table structure\\recognition
  \end{tabular} & N/A & \begin{tabular}{c} Scanned printed\\and mixed pages \\from measurement\\logbooks from \\1700 - modern days
  \end{tabular} & Color & N/A 
   & JPG  &  \begin{tabular}{c} Full table, individual\\ cells and coarse \\segmentation cell\\  bounding boxes
  \end{tabular} & \begin{tabular}{c}\\XML\\VOC2007\end{tabular}
   & \begin{tabular}{l}\\CascadeTabNet \cite{Prasad2020CascadeTabNetAA}\\CascadeTabNet \cite{Prasad2020CascadeTabNetAA} + Postprocessing \cite{10.1145/3068335}
  \end{tabular}  & \begin{tabular}{c}\\\ac{WA} F1-score\end{tabular} & \begin{tabular}{c c c}+cTDaR19 & Ind. cells & Coarse cells \\ \hline 0.071 & 0.047& 0.385\\0.170&  0.263&  0.578
  \end{tabular} \\

  \hline
  
  
  \textbf{\begin{tabular}{l}Digital Peter\\ \cite{10.1145/3476887.3476892} - \ref{sssec:digital_peter}\end{tabular}} &
  \begin{tabular}{c}  9,694 text-line images\\
  (6,237 train, 1,930 \\validation, 1,527 test)\\
  265,788 characters\\
  50,998 words\\\end{tabular} & 
  \begin{tabular}{c}N/A \end{tabular}& \begin{tabular}{c} Line segmentation\\Line recognition
  \end{tabular} & Russian & \begin{tabular}{c} 
  Manuscripts from \\Peter the Great \\written from\\ 1709-1713
  \end{tabular} & Color & N/A
   &  JPG& \begin{tabular}{c}Line polygons\\Line transcriptions\end{tabular}& \begin{tabular}{c}COCO\\TXT\end{tabular}
   & \begin{tabular}{c}CNN-GRU-CTC \cite{10.1145/1143844.1143891}\\ (optimized)\end{tabular}
   & \begin{tabular}{c}\ac{CER} (\%)\\\ac{WER} (\%)\\String \ac{acc} (\%)
  \end{tabular} & 
  \begin{tabular}{c}3.5\\19.4\\52.3
  \end{tabular}\\

  \hline

  
  \textbf{\begin{tabular}{l}BiblIA\\ \cite{10.1145/3476887.3476892} - \ref{sssec:BiblIA}\end{tabular}} &
  \begin{tabular}{c} 202 images \\11,285 lines\\74,675 words
  \end{tabular} & 
  \begin{tabular}{c}\textbf{Script classes:} Ashkenazi,\\ Italian, Sephardi \end{tabular}& \begin{tabular}{c} Line segmentation\\Text transcription
  \end{tabular} & \begin{tabular}{c}  Hebrew\\Aramaic
  \end{tabular} & \begin{tabular}{c} 
  Hebrew\\ Manuscripts\\written from\\ \nth{11}-\nth{15}
  \end{tabular} & \begin{tabular}{c} Color\\Grayscale
  \end{tabular} & N/A
   &  JPG& \begin{tabular}{c}Baselines\\ Transcriptions\end{tabular}& XML
   & \begin{tabular}{c}Segmentation\\ and recognition\\models based on\\ kraken OCR\end{tabular}
   & \begin{tabular}{c}\ac{acc} (\%)\\\ac{WER} (\%)\\
  \end{tabular} & 
  \begin{tabular}{c}97.24 (All texts)\\8.5
  \end{tabular}\\

    \hline

  
  \textbf{\begin{tabular}{l}HisClima\\ \cite{9412210} - \ref{sssec:hisclima}\end{tabular}} &
  \begin{tabular}{c} 208 pages, 33,739 lines\\66,814 running words,\\ 15,471 relevant information
  \end{tabular} & 
  \begin{tabular}{c} 1,483 lexicon, 76 characters\end{tabular}& \begin{tabular}{c} Text recognition\\Layout analysis\\Information extraction
  \end{tabular} & N/A & \begin{tabular}{c} 
  Pages from a ship\\ weather log book\\that sailed from\\ 1880-1881\\
  \end{tabular} & \begin{tabular}{c} Color
  \end{tabular} & N/A
   & N/A & \begin{tabular}{c}Region blocks, columns,\\ rows, lines\\ Transcriptions\\Relevant information\end{tabular}& N/A
   & \begin{tabular}{c}Layout: NN-based \cite{Quirs2018MultiTaskHD}\\ OCR: CRNN + CTC + LM\\
   Information extraction based on \cite{8563224}\end{tabular}
   & \begin{tabular}{c}P - R - F1\\
   CER (\%) - WER (\%) \\P - R - F1\\
   
  \end{tabular} & 
  \begin{tabular}{c}0.91 - 0.72 - 0.80\\
   2.7 - 4.4 \\cell pos: 0.95 - 0.95 - 0.95\\
   line geom: 0.79 - 0.79 - 0.785
  \end{tabular}\\

    \hline

  
  \textbf{\begin{tabular}{l}Hugin-Munin\\ \cite{10.1007/978-3-031-06555-2_27} - \ref{sssec:hugin_munin}\end{tabular}} &
  \begin{tabular}{c}  828 pages, 23,732 lines\\164,922 words\\752,080 characters
  \end{tabular} & 
  \begin{tabular}{c}12 writers \end{tabular}& \begin{tabular}{c} Handwritten\\text recognition
  \end{tabular} & Norwegian & \begin{tabular}{c} 
  Pages from private \\correspondences and \\diaries from 12 writers\\ written between 1820-1950
  \end{tabular} & \begin{tabular}{c} Color
  \end{tabular} & N/A
   & N/A & \begin{tabular}{c} Transcriptions\end{tabular}& PAGE XML
   & \begin{tabular}{c}PyLaia expert \cite{puigcerver2018pylaia}\\Kaldi expert \cite{Arora2019UsingAM}\end{tabular}
   & \begin{tabular}{c}\ac{CER} (\%)\\\ac{WER} (\%)
  \end{tabular} & 
  \begin{tabular}{c}8.86\\22.19
  \end{tabular}\\

  \hline
  
  
  \textbf{\begin{tabular}{l}POPP\\ \cite{10.1007/978-3-031-06555-2_10} - \ref{sssec:popp}\end{tabular}} &
  \begin{tabular}{c} \textbf{Generic:} 128 train, 16 val, 16 test pages\\3,840 train, 480 val, 480 test lines\\
  \textbf{Belleville:} 38 train, 5 val, 6 test pages\\1,140 train, 150 val, 180 test lines\\
  \textbf{Chaussée d'Antin:} 625 train, 78 val, 77 test lines
  \end{tabular} & 
  \begin{tabular}{c} Generic - Belleville - Chaussée d'Antin\\80 - 1 - 10 writers\end{tabular}& \begin{tabular}{c} Handwritten\\table recognition\\and information extraction
  \end{tabular} & French & \begin{tabular}{c} 
  Pages from Paris\\ census tables\\ from 1926
  \end{tabular} & \begin{tabular}{c} Grayscale
  \end{tabular} & 200 dpi
   & TIFF & \begin{tabular}{c} Line bbox coordinates\\ Transcriptions\end{tabular}& \begin{tabular}{c} XML\\JSON\end{tabular}
   & \begin{tabular}{c}Optical model \cite{Coquenet2022} \\+ self-training\\ with extracted corpus\end{tabular}
   & \begin{tabular}{c}\ac{CER} (\%)\\\ac{WER} (\%)
  \end{tabular} & 
  \begin{tabular}{c}4.52\\13.57
  \end{tabular}\\
  
  \hline
  
  \end{tabular}}

  
\end{minipage}
\end{adjustbox}
\end{center}
\end{sidewaystable}

\restoregeometry
\bibliographystyle{sn-basic}
\bibliography{bib}

\begin{thebibliography}{195}
\providecommand{\natexlab}[1]{#1}
\providecommand{\url}[1]{{#1}}
\providecommand{\urlprefix}{URL }
\providecommand{\doi}[1]{\url{https://doi.org/#1}}
\providecommand{\eprint}[2][]{\url{#2}}
 \bibcommenthead

\bibitem[{Adam et~al(2018)Adam, Baig, Al-Maadeed, Bouridane, and
  El-Menshawy}]{adam2018kertas}
Adam K, Baig A, Al-Maadeed S, et~al (2018) {KERTAS}: dataset for automatic
  dating of ancient {A}rabic manuscripts. International Journal on Document
  Analysis and Recognition (IJDAR) 21(4):283--290

\bibitem[{Alaei et~al(2011{\natexlab{a}})Alaei, Nagabhushan, and Pal}]{6121553}
Alaei A, Nagabhushan P, Pal U (2011{\natexlab{a}}) A {N}ew {D}ataset of
  {P}ersian {H}andwritten {D}ocuments and {I}ts {S}egmentation. In: 2011 7th
  Iranian Conference on Machine Vision and Image Processing, pp 1--5,
  \doi{10.1109/IranianMVIP.2011.6121553}

\bibitem[{Alaei et~al(2011{\natexlab{b}})Alaei, Nagabhushan, and
  Pal}]{10.1007/s10044-011-0226-x}
Alaei A, Nagabhushan P, Pal U (2011{\natexlab{b}}) Piece-{W}ise {P}ainting
  {T}echnique for {L}ine {S}egmentation of {U}nconstrained {H}andwritten
  {T}ext: A {S}pecific {S}tudy with {P}ersian {T}ext {D}ocuments. Pattern Anal
  Appl 14(4):381–394. \doi{10.1007/s10044-011-0226-x}

\bibitem[{Alaei et~al(2011{\natexlab{c}})Alaei, Pal, and
  Nagabhushan}]{Alaei2011ANS}
Alaei A, Pal U, Nagabhushan PN (2011{\natexlab{c}}) A new scheme for
  unconstrained handwritten text-line segmentation. Pattern Recognit
  44:917--928

\bibitem[{Alaei et~al(2012)Alaei, Pal, and Nagabhushan}]{Alaei2012DatasetAG}
Alaei A, Pal U, Nagabhushan PN (2012) {Dataset and Ground Truth for Handwritten
  Text in Four Different Scripts}. Int J Pattern Recognit Artif Intell 26

\bibitem[{Almaz{\'a}n et~al(2012)Almaz{\'a}n, Gordo, Forn{\'e}s, and
  Valveny}]{Almazn2012EfficientEW}
Almaz{\'a}n J, Gordo A, Forn{\'e}s A, et~al (2012) {Efficient Exemplar Word
  Spotting}. In: BMVC

\bibitem[{Almazán et~al(2014)Almazán, Gordo, Fornés, and Valveny}]{6857995}
Almazán J, Gordo A, Fornés A, et~al (2014) Word {S}potting and {R}ecognition
  with {E}mbedded {A}ttributes. IEEE Transactions on Pattern Analysis and
  Machine Intelligence 36(12):2552--2566. \doi{10.1109/TPAMI.2014.2339814}

\bibitem[{Anna et~al(2021)Anna, Simon, Juliette, Ljudmila, Caroline, and
  Thibault}]{10.1145/3476887.3476913}
Anna SB, Simon G, Juliette J, et~al (2021) The {BIR} {D}atabase –
  {I}dentifying {T}ypographic {E}mphasis in {L}ist-like {H}istorical
  {D}ocuments. In: The 6th International Workshop on Historical Document
  Imaging and Processing. Association for Computing Machinery, New York, NY,
  USA, HIP '21, p 37–42, \doi{10.1145/3476887.3476913}

\bibitem[{Arora et~al(2019)Arora, Chang, Rekabdar, Povey, Etter, Raj, Hadian,
  Trmal, Garc{\'i}a, Watanabe, Manohar, Shao, and Khudanpur}]{Arora2019UsingAM}
Arora A, Chang CC, Rekabdar B, et~al (2019) {Using ASR Methods for OCR}. 2019
  International Conference on Document Analysis and Recognition (ICDAR) pp
  663--668

\bibitem[{Arvanitopoulos and Süsstrunk(2014)}]{6981106}
Arvanitopoulos N, Süsstrunk S (2014) Seam {C}arving for {T}ext {L}ine
  {E}xtraction on {C}olor and {G}rayscale {H}istorical {M}anuscripts. In: 2014
  14th International Conference on Frontiers in Handwriting Recognition, pp
  726--731, \doi{10.1109/ICFHR.2014.127}

\bibitem[{Arvanitopoulos et~al(2017)Arvanitopoulos, Chevassus, Maggetti, and
  S{\"u}sstrunk}]{Arvanitopoulos2017AHF}
Arvanitopoulos N, Chevassus G, Maggetti D, et~al (2017) A {H}andwritten
  {F}rench {D}ataset for {W}ord {S}potting: {CFRAMUZ}. Proceedings of the 4th
  International Workshop on Historical Document Imaging and Processing

\bibitem[{Belay et~al(2019)Belay, Habtegebirial, Liwicki, Belay, and
  Stricker}]{8977980}
Belay BH, Habtegebirial T, Liwicki M, et~al (2019) Amharic {T}ext {I}mage
  {R}ecognition: Database, {A}lgorithm, and {A}nalysis. In: 2019 International
  Conference on Document Analysis and Recognition (ICDAR), pp 1268--1273,
  \doi{10.1109/ICDAR.2019.00205}

\bibitem[{Binmakhashen and Mahmoud(2019)}]{10.1145/3355610}
Binmakhashen GM, Mahmoud SA (2019) Document layout analysis: A comprehensive
  survey. ACM Comput Surv 52(6). \doi{10.1145/3355610}

\bibitem[{Bishop(1994)}]{Bishop94mixturedensity}
Bishop CM (1994) Mixture density networks. Tech. rep.

\bibitem[{Boillet et~al(2019)Boillet, Bonhomme, Stutzmann, and
  Kermorvant}]{10.1145/3352631.3352633}
Boillet M, Bonhomme ML, Stutzmann D, et~al (2019) {HORAE}: {A}n {A}nnotated
  {D}ataset of {B}ooks of {H}ours. In: Proceedings of the 5th International
  Workshop on Historical Document Imaging and Processing. Association for
  Computing Machinery, New York, NY, USA, HIP '19, p 7–12,
  \doi{10.1145/3352631.3352633}

\bibitem[{Breuel(2008)}]{Breuel2008TheOO}
Breuel TM (2008) The {OCR}opus open source {OCR} system. In: Electronic Imaging

\bibitem[{Brink et~al(2012)Brink, Smit, Bulacu, and Schomaker}]{BRINK2012162}
Brink A, Smit J, Bulacu M, et~al (2012) Writer identification using directional
  ink-trace width measurements. Pattern Recognition 45(1):162--171.
  \doi{https://doi.org/10.1016/j.patcog.2011.07.005}

\bibitem[{Bromley et~al(1993)Bromley, Bentz, Bottou, Guyon, LeCun, Moore,
  S{\"a}ckinger, and Shah}]{Bromley1993SignatureVU}
Bromley J, Bentz JW, Bottou L, et~al (1993) Signature {V}erification {U}sing
  {A} "{S}iamese" {T}ime {D}elay {N}eural {N}etwork. In: Int. J. Pattern
  Recognit. Artif. Intell.

\bibitem[{Bulacu and Schomaker(2007)}]{4107573}
Bulacu M, Schomaker L (2007) Text-{I}ndependent {W}riter {I}dentification and
  {V}erification {U}sing {T}extural and {A}llographic {F}eatures. IEEE
  Transactions on Pattern Analysis and Machine Intelligence 29(4):701--717.
  \doi{10.1109/TPAMI.2007.1009}

\bibitem[{Burie et~al(2016)Burie, Coustaty, Hadi, Kesiman, Ogier, Paulus, Sok,
  Sunarya, and Valy}]{7814130}
Burie JC, Coustaty M, Hadi S, et~al (2016) {ICFHR2016} {C}ompetition on the
  {A}nalysis of {H}andwritten {T}ext in {I}mages of {B}alinese {P}alm {L}eaf
  {M}anuscripts. In: 2016 15th International Conference on Frontiers in
  Handwriting Recognition (ICFHR), pp 596--601, \doi{10.1109/ICFHR.2016.0114}

\bibitem[{Cai and Vasconcelos(2018)}]{Cai2018CascadeRD}
Cai Z, Vasconcelos N (2018) Cascade {R-CNN}: Delving {I}nto {H}igh {Q}uality
  {O}bject {D}etection. 2018 IEEE/CVF Conference on Computer Vision and Pattern
  Recognition pp 6154--6162

\bibitem[{Causer and Wallace(2012)}]{Causer2012BuildingAV}
Causer T, Wallace V (2012) {Building A Volunteer Community: Results and
  Findings from Transcribe Bentham}. Digit Humanit Q 6

\bibitem[{Chen and Wang(2000)}]{Chen2000SegmentationOS}
Chen YK, Wang JF (2000) Segmentation of {S}ingle- or {M}ultiple-{T}ouching
  {H}andwritten {N}umeral {S}tring {U}sing {B}ackground and {F}oreground
  {A}nalysis. IEEE Trans Pattern Anal Mach Intell 22:1304--1317

\bibitem[{Chollet(2017)}]{chollet2017xception}
Chollet F (2017) {Xception: Deep Learning with Depthwise Separable
  Convolutions}. 2017 IEEE Conference on Computer Vision and Pattern
  Recognition (CVPR) pp 1800--1807

\bibitem[{Christlein et~al(2015)Christlein, Bernecker, and
  Angelopoulou}]{7333893}
Christlein V, Bernecker D, Angelopoulou E (2015) {Writer identification using
  VLAD encoded contour-Zernike moments}. In: 2015 13th International Conference
  on Document Analysis and Recognition (ICDAR), pp 906--910,
  \doi{10.1109/ICDAR.2015.7333893}

\bibitem[{Christlein et~al(2017{\natexlab{a}})Christlein, Bernecker, H{\"o}nig,
  Maier, and Angelopoulou}]{Christlein2017WriterIU}
Christlein V, Bernecker D, H{\"o}nig F, et~al (2017{\natexlab{a}}) {Writer
  Identification Using GMM Supervectors and Exemplar-SVMs}. Pattern Recognit
  63:258--267

\bibitem[{Christlein et~al(2017{\natexlab{b}})Christlein, Gropp, Fiel, and
  Maier}]{christlein2017unsupervised}
Christlein V, Gropp M, Fiel S, et~al (2017{\natexlab{b}}) {Unsupervised Feature
  Learning for Writer Identification and Writer Retrieval}. 2017 14th IAPR
  International Conference on Document Analysis and Recognition (ICDAR)
  01:991--997

\bibitem[{Christlein et~al(2019)Christlein, Nicolaou, Seuret, Stutzmann, and
  Maier}]{Christlein2019ICDAR2C}
Christlein V, Nicolaou A, Seuret M, et~al (2019) {ICDAR 2019 Competition on
  Image Retrieval for Historical Handwritten Documents}. 2019 International
  Conference on Document Analysis and Recognition (ICDAR) pp 1505--1509

\bibitem[{Cilia et~al(2021)Cilia, De~Stefano, Fontanella, Marthot-Santaniello,
  and Scotto~di Freca}]{10.1007/978-3-030-68787-8_16}
Cilia ND, De~Stefano C, Fontanella F, et~al (2021) {PapyRow: A Dataset of Row
  Images from Ancient Greek Papyri for Writers Identification}. In: Del~Bimbo
  A, Cucchiara R, Sclaroff S, et~al (eds) Pattern Recognition. ICPR
  International Workshops and Challenges. Springer International Publishing,
  Cham, pp 223--234

\bibitem[{Clanuwat et~al(2018)Clanuwat, Bober-Irizar, Kitamoto, Lamb, Yamamoto,
  and Ha}]{Clanuwat2018DeepLF}
Clanuwat T, Bober-Irizar M, Kitamoto A, et~al (2018) Deep {L}earning for
  {C}lassical {J}apanese {L}iterature. ArXiv abs/1812.01718

\bibitem[{Clausner et~al(2011)Clausner, Pletschacher, and
  Antonacopoulos}]{6065274}
Clausner C, Pletschacher S, Antonacopoulos A (2011) {Aletheia - An Advanced
  Document Layout and Text Ground-Truthing System for Production Environments}.
  In: 2011 International Conference on Document Analysis and Recognition, pp
  48--52, \doi{10.1109/ICDAR.2011.19}

\bibitem[{{Clausner} et~al(2015){Clausner}, {Papadopoulos}, {Pletschacher}, and
  {Antonacopoulos}}]{7333898}
{Clausner} C, {Papadopoulos} C, {Pletschacher} S, et~al (2015) {The ENP image
  and ground truth dataset of historical newspapers}. In: 2015 13th
  International Conference on Document Analysis and Recognition (ICDAR), pp
  931--935, \doi{10.1109/ICDAR.2015.7333898}

\bibitem[{Clausner et~al(2017)Clausner, Antonacopoulos, Derrick, and
  Pletschacher}]{8270161}
Clausner C, Antonacopoulos A, Derrick T, et~al (2017) {ICDAR2017 C}ompetition
  on {R}ecognition of {E}arly {I}ndian {P}rinted {D}ocuments - {REID2017}. In:
  2017 14th IAPR International Conference on Document Analysis and Recognition
  (ICDAR), pp 1411--1416, \doi{10.1109/ICDAR.2017.230}

\bibitem[{Clausner et~al(2018)Clausner, Antonacopoulos, Mcgregor, and
  Wilson-Nunn}]{8583806}
Clausner C, Antonacopoulos A, Mcgregor N, et~al (2018) {ICFHR 2018 C}ompetition
  on {R}ecognition of {H}istorical {A}rabic {S}cientific {M}anuscripts –
  {RASM2018}. In: 2018 16th International Conference on Frontiers in
  Handwriting Recognition (ICFHR), pp 471--476,
  \doi{10.1109/ICFHR-2018.2018.00088}

\bibitem[{Clausner et~al(2019)Clausner, Antonacopoulos, Derrick, and
  Pletschacher}]{8978191}
Clausner C, Antonacopoulos A, Derrick T, et~al (2019) {ICDAR2019 C}ompetition
  on {R}ecognition of {E}arly {I}ndian {P}rinted {D}ocuments – {REID2019}.
  In: 2019 International Conference on Document Analysis and Recognition
  (ICDAR), pp 1527--1532, \doi{10.1109/ICDAR.2019.00246}

\bibitem[{Clinchant et~al(2018)Clinchant, Déjean, Meunier, Lang, and
  Kleber}]{8395184}
Clinchant S, Déjean H, Meunier JL, et~al (2018) {Comparing Machine Learning
  Approaches for Table Recognition in Historical Register Books}. In: 2018 13th
  IAPR International Workshop on Document Analysis Systems (DAS), pp 133--138,
  \doi{10.1109/DAS.2018.44}

\bibitem[{Cloppet et~al(2016)Cloppet, Églin, Kieu, Stutzmann, and
  Vincent}]{7814129}
Cloppet F, Églin V, Kieu VC, et~al (2016) {ICFHR2016 C}ompetition on the
  {C}lassification of {M}edieval {H}andwritings in {L}atin {S}cript. In: 2016
  15th International Conference on Frontiers in Handwriting Recognition
  (ICFHR), pp 590--595, \doi{10.1109/ICFHR.2016.0113}

\bibitem[{Cloppet et~al(2017)Cloppet, Eglin, Helias-Baron, Kieu, Vincent, and
  Stutzmann}]{8270155}
Cloppet F, Eglin V, Helias-Baron M, et~al (2017) {ICDAR2017 C}ompetition on the
  {C}lassification of {M}edieval {H}andwritings in {L}atin {S}cript. In: 2017
  14th IAPR International Conference on Document Analysis and Recognition
  (ICDAR), pp 1371--1376, \doi{10.1109/ICDAR.2017.224}

\bibitem[{Constum et~al(2022)Constum, Kempf, Paquet, Tranouez, Chatelain,
  Br{\'e}e, and Merveille}]{10.1007/978-3-031-06555-2_10}
Constum T, Kempf N, Paquet T, et~al (2022) {Recognition and Information
  Extraction in Historical Handwritten Tables: Toward Understanding Early
  \nth{20} Century Paris Census}. In: Uchida S, Barney E, Eglin V (eds)
  Document Analysis Systems. Springer International Publishing, Cham, pp
  143--157

\bibitem[{Coquenet et~al(2022)Coquenet, Chatelain, and Paquet}]{Coquenet2022}
Coquenet D, Chatelain C, Paquet T (2022) {End-to-end Handwritten Paragraph Text
  Recognition Using a Vertical Attention Network}. IEEE Transactions on Pattern
  Analysis and Machine Intelligence \doi{10.1109/TPAMI.2022.3144899}

\bibitem[{Dalal and Triggs(2005)}]{1467360}
Dalal N, Triggs B (2005) Histograms of oriented gradients for human detection.
  In: 2005 IEEE Computer Society Conference on Computer Vision and Pattern
  Recognition (CVPR'05), pp 886--893 vol. 1, \doi{10.1109/CVPR.2005.177}

\bibitem[{Daniel et~al(2021)Daniel, Bronson, Pawel, Hayim, Benjamin, and
  Elena}]{10.1145/3476887.3476896}
Daniel SBE, Bronson BD, Pawel J, et~al (2021) {BiblIA - a General Model for
  Medieval Hebrew Manuscripts and an Open Annotated Dataset}. In: The 6th
  International Workshop on Historical Document Imaging and Processing.
  Association for Computing Machinery, New York, NY, USA, HIP '21, p 61–66,
  \doi{10.1145/3476887.3476896}

\bibitem[{Dehak et~al(2011)Dehak, Torres-Carrasquillo, Reynolds, and
  Dehak}]{Dehak2011LanguageRV}
Dehak N, Torres-Carrasquillo PA, Reynolds DA, et~al (2011) {Language
  Recognition via i-vectors and Dimensionality Reduction}. In: INTERSPEECH

\bibitem[{Diem et~al(2017)Diem, Kleber, Fiel, Grüning, and Gatos}]{8270153}
Diem M, Kleber F, Fiel S, et~al (2017) c{BAD: ICDAR2017 C}ompetition on
  {B}aseline {D}etection. In: 2017 14th IAPR International Conference on
  Document Analysis and Recognition (ICDAR), pp 1355--1360,
  \doi{10.1109/ICDAR.2017.222}

\bibitem[{Djeddi et~al(2013)Djeddi, Siddiqi, Souici-Meslati, and
  Ennaji}]{10.1016/j.patrec.2013.03.020}
Djeddi C, Siddiqi I, Souici-Meslati L, et~al (2013) Text-{I}ndependent {W}riter
  {R}ecognition {U}sing {M}ulti-{S}cript {H}andwritten {T}exts. Pattern Recogn
  Lett 34(10):1196–1202. \doi{10.1016/j.patrec.2013.03.020}

\bibitem[{{Dolfing} et~al(2020){Dolfing}, {Bellegarda}, {Chorowski}, {Marxer},
  and {Laurent}}]{9257750}
{Dolfing} HJGA, {Bellegarda} J, {Chorowski} J, et~al (2020) {The
  “ScribbleLens” Dutch Historical Handwriting Corpus}. In: 2020 17th
  International Conference on Frontiers in Handwriting Recognition (ICFHR), pp
  67--72, \doi{10.1109/ICFHR2020.2020.00023}

\bibitem[{Dulla(2018)}]{iet:/content/conferences/10.1049/cp.2018.1286}
Dulla A (2018) {A dataset of Warped Historical Arabic Documents}. IET
  Conference Proceedings pp 10 (6 pp.)--10 (6 pp.)(1).
  \urlprefix\url{https://digital-library.theiet.org/content/conferences/10.1049/cp.2018.1286}

\bibitem[{En et~al(2016{\natexlab{a}})En, Nicolas, Petitjean, Jurie, and
  Heutte}]{10.1117/1.JEI.26.1.011010}
En S, Nicolas S, Petitjean C, et~al (2016{\natexlab{a}}) {New public dataset
  for spotting patterns in medieval document images}. Journal of Electronic
  Imaging 26(1):1 -- 15. \doi{10.1117/1.JEI.26.1.011010}

\bibitem[{En et~al(2016{\natexlab{b}})En, Petitjean, Nicolas, and
  Heutte}]{EN2016149}
En S, Petitjean C, Nicolas S, et~al (2016{\natexlab{b}}) A scalable pattern
  spotting system for historical documents. Pattern Recognition 54:149--161.
  \doi{https://doi.org/10.1016/j.patcog.2016.01.014}

\bibitem[{Everingham et~al(2009)Everingham, Gool, Williams, Winn, and
  Zisserman}]{Everingham2009ThePV}
Everingham M, Gool LV, Williams CKI, et~al (2009) The {P}ascal {V}isual
  {O}bject {C}lasses ({VOC) C}hallenge. International Journal of Computer
  Vision 88:303--338

\bibitem[{Fernández-Mota et~al(2014)Fernández-Mota, Almazán, Cirera,
  Fornés, and Lladós}]{6976764}
Fernández-Mota D, Almazán J, Cirera N, et~al (2014) {BH2M: T}he {B}arcelona
  {H}istorical, {H}andwritten {M}arriages {D}atabase. In: 2014 22nd
  International Conference on Pattern Recognition, pp 256--261,
  \doi{10.1109/ICPR.2014.53}

\bibitem[{Fiel et~al(2017)Fiel, Kleber, Diem, Christlein, Louloudis, Nikos, and
  Gatos}]{8270156}
Fiel S, Kleber F, Diem M, et~al (2017) {ICDAR2017 C}ompetition on {H}istorical
  {D}ocument {W}riter {I}dentification ({H}istorical-{WI}). In: 2017 14th IAPR
  International Conference on Document Analysis and Recognition (ICDAR), pp
  1377--1382, \doi{10.1109/ICDAR.2017.225}

\bibitem[{Fischer et~al(2009)Fischer, Wuthrich, Liwicki, Frinken, Bunke,
  Viehhauser, and Stolz}]{5306020}
Fischer A, Wuthrich M, Liwicki M, et~al (2009) {Automatic Transcription of
  Handwritten Medieval Documents}. In: 2009 15th International Conference on
  Virtual Systems and Multimedia, pp 137--142, \doi{10.1109/VSMM.2009.26}

\bibitem[{Fischer et~al(2010)Fischer, Inderm\"{u}hle, Bunke, Viehhauser, and
  Stolz}]{10.1145/1815330.1815331}
Fischer A, Inderm\"{u}hle E, Bunke H, et~al (2010) Ground {T}ruth {C}reation
  for {H}andwriting {R}ecognition in {H}istorical {D}ocuments. In: Proceedings
  of the 9th IAPR International Workshop on Document Analysis Systems.
  Association for Computing Machinery, New York, NY, USA, DAS '10, p 3–10,
  \doi{10.1145/1815330.1815331}

\bibitem[{Fischer et~al(2011)Fischer, Frinken, Forn\'{e}s, and
  Bunke}]{10.1145/2037342.2037348}
Fischer A, Frinken V, Forn\'{e}s A, et~al (2011) {Transcription Alignment of
  Latin Manuscripts Using Hidden Markov Models}. In: Proceedings of the 2011
  Workshop on Historical Document Imaging and Processing. Association for
  Computing Machinery, New York, NY, USA, HIP '11, p 29–36,
  \doi{10.1145/2037342.2037348}

\bibitem[{Fischer et~al(2012)Fischer, Keller, Frinken, and
  Bunke}]{FISCHER2012934}
Fischer A, Keller A, Frinken V, et~al (2012) {Lexicon-free handwritten word
  spotting using character HMMs}. Pattern Recognition Letters 33(7):934--942.
  \doi{https://doi.org/10.1016/j.patrec.2011.09.009}, special Issue on Awards
  from ICPR 2010

\bibitem[{Fornés et~al(2017)Fornés, Romero, Baró, Toledo, Sánchez, Vidal,
  and Lladós}]{8270158}
Fornés A, Romero V, Baró A, et~al (2017) {ICDAR2017 Competition on
  Information Extraction in Historical Handwritten Records}. In: 2017 14th IAPR
  International Conference on Document Analysis and Recognition (ICDAR), pp
  1389--1394, \doi{10.1109/ICDAR.2017.227}

\bibitem[{Frinken et~al(2012)Frinken, Fischer, Manmatha, and Bunke}]{5871643}
Frinken V, Fischer A, Manmatha R, et~al (2012) {A Novel Word Spotting Method
  Based on Recurrent Neural Networks}. IEEE Transactions on Pattern Analysis
  and Machine Intelligence 34(2):211--224. \doi{10.1109/TPAMI.2011.113}

\bibitem[{Gao et~al(2019)Gao, Huang, Déjean, Meunier, Yan, Fang, Kleber, and
  Lang}]{8978120}
Gao L, Huang Y, Déjean H, et~al (2019) {ICDAR 2019 C}ompetition on {T}able
  {D}etection and {R}ecognition (c{TDaR}). In: 2019 International Conference on
  Document Analysis and Recognition (ICDAR), pp 1510--1515,
  \doi{10.1109/ICDAR.2019.00243}

\bibitem[{Gatos et~al(2009)Gatos, Ntirogiannis, and Pratikakis}]{5277767}
Gatos B, Ntirogiannis K, Pratikakis I (2009) {ICDAR 2009 D}ocument {I}mage
  {B}inarization {C}ontest ({DIBCO 2009}). In: 2009 10th International
  Conference on Document Analysis and Recognition, pp 1375--1382,
  \doi{10.1109/ICDAR.2009.246}

\bibitem[{Gatos et~al(2011)Gatos, Kesidis, and Papandreou}]{6065492}
Gatos B, Kesidis AL, Papandreou A (2011) {Adaptive Zoning Features for
  Character and Word Recognition}. In: 2011 International Conference on
  Document Analysis and Recognition, pp 1160--1164,
  \doi{10.1109/ICDAR.2011.234}

\bibitem[{Gatos et~al(2015)Gatos, Stamatopoulos, Louloudis, Sfikas, Retsinas,
  Papavassiliou, Sunistira, and Katsouros}]{7333841}
Gatos B, Stamatopoulos N, Louloudis G, et~al (2015) {GRPOLY-DB: An old Greek
  polytonic document image database}. In: 2015 13th International Conference on
  Document Analysis and Recognition (ICDAR), pp 646--650,
  \doi{10.1109/ICDAR.2015.7333841}

\bibitem[{Gattal et~al(2016)Gattal, Djeddi, Chibani, and Siddiqi}]{7490135}
Gattal A, Djeddi C, Chibani Y, et~al (2016) Isolated {H}andwritten {D}igit
  {R}ecognition {U}sing o{BIF}s and {B}ackground {F}eatures. In: 2016 12th IAPR
  Workshop on Document Analysis Systems (DAS), pp 305--310,
  \doi{10.1109/DAS.2016.10}

\bibitem[{Gattal et~al(2017)Gattal, Chibani, and
  Hadjadji}]{Gattal2017SegmentationAR}
Gattal A, Chibani Y, Hadjadji B (2017) Segmentation and recognition system for
  unknown-length handwritten digit strings. Pattern Analysis and Applications
  20:307--323

\bibitem[{Giotis et~al(2017)Giotis, Sfikas, Gatos, and
  Nikou}]{10.1016/j.patcog.2017.02.023}
Giotis AP, Sfikas G, Gatos B, et~al (2017) {A Survey of Document Image Word
  Spotting Techniques}. Pattern Recogn 68(C):310–332.
  \doi{10.1016/j.patcog.2017.02.023},
  \urlprefix\url{https://doi.org/10.1016/j.patcog.2017.02.023}

\bibitem[{Graves et~al(2006)Graves, Fern\'{a}ndez, Gomez, and
  Schmidhuber}]{10.1145/1143844.1143891}
Graves A, Fern\'{a}ndez S, Gomez F, et~al (2006) {Connectionist Temporal
  Classification: Labelling Unsegmented Sequence Data with Recurrent Neural
  Networks}. In: Proceedings of the 23rd International Conference on Machine
  Learning. Association for Computing Machinery, New York, NY, USA, ICML '06, p
  369–376, \doi{10.1145/1143844.1143891}

\bibitem[{Graves et~al(2009)Graves, Liwicki, Fernández, Bertolami, Bunke, and
  Schmidhuber}]{Graves2009855}
Graves A, Liwicki M, Fernández S, et~al (2009) A novel connectionist system
  for unconstrained handwriting recognition. IEEE Transactions on Pattern
  Analysis and Machine Intelligence 31(5):855--868.
  \doi{10.1109/TPAMI.2008.137}

\bibitem[{{Grüning} et~al(2018){Grüning}, {Labahn}, {Diem}, {Kleber}, and
  {Fiel}}]{8395221}
{Grüning} T, {Labahn} R, {Diem} M, et~al (2018) {READ-BAD: A New Dataset and
  Evaluation Scheme for Baseline Detection in Archival Documents}. In: 2018
  13th IAPR International Workshop on Document Analysis Systems (DAS), pp
  351--356, \doi{10.1109/DAS.2018.38}

\bibitem[{Ha and Eck(2018)}]{Ha2018ANR}
Ha DR, Eck D (2018) {A Neural Representation of Sketch Drawings}. ArXiv
  abs/1704.03477

\bibitem[{Hajič and Pecina(2017)}]{8269947}
Hajič J, Pecina P (2017) {The MUSCIMA++ Dataset for Handwritten Optical Music
  Recognition}. In: 2017 14th IAPR International Conference on Document
  Analysis and Recognition (ICDAR), pp 39--46, \doi{10.1109/ICDAR.2017.16}

\bibitem[{Harley et~al(2015)Harley, Ufkes, and Derpanis}]{harley2015evaluation}
Harley AW, Ufkes A, Derpanis KG (2015) {Evaluation of deep convolutional nets
  for document image classification and retrieval}. 2015 13th International
  Conference on Document Analysis and Recognition (ICDAR) pp 991--995

\bibitem[{He et~al(2016{\natexlab{a}})He, Zhang, Ren, and Sun}]{He2016DeepRL}
He K, Zhang X, Ren S, et~al (2016{\natexlab{a}}) {Deep Residual Learning for
  Image Recognition}. 2016 IEEE Conference on Computer Vision and Pattern
  Recognition (CVPR) pp 770--778

\bibitem[{He et~al(2016{\natexlab{b}})He, Zhang, Ren, and
  Sun}]{10.1007/978-3-319-46493-0_38}
He K, Zhang X, Ren S, et~al (2016{\natexlab{b}}) Identity mappings in deep
  residual networks. In: Leibe B, Matas J, Sebe N, et~al (eds) Computer Vision
  -- ECCV 2016. Springer International Publishing, Cham, pp 630--645

\bibitem[{He et~al(2020)He, Gkioxari, Doll{\'a}r, and Girshick}]{he2017mask}
He K, Gkioxari G, Doll{\'a}r P, et~al (2020) {Mask R-CNN}. IEEE Transactions on
  Pattern Analysis and Machine Intelligence 42:386--397

\bibitem[{Huang et~al(2017)Huang, Liu, and Weinberger}]{huang2018densely}
Huang G, Liu Z, Weinberger KQ (2017) {Densely Connected Convolutional
  Networks}. 2017 IEEE Conference on Computer Vision and Pattern Recognition
  (CVPR) pp 2261--2269

\bibitem[{Huang et~al(2019)Huang, Wang, Liu, Shi, and Jin}]{8978032}
Huang S, Wang H, Liu Y, et~al (2019) {OBC306: A Large-Scale Oracle Bone
  Character Recognition Dataset}. In: 2019 International Conference on Document
  Analysis and Recognition (ICDAR), pp 681--688, \doi{10.1109/ICDAR.2019.00114}

\bibitem[{Hull(1994)}]{291440}
Hull J (1994) A database for handwritten text recognition research. IEEE
  Transactions on Pattern Analysis and Machine Intelligence 16(5):550--554.
  \doi{10.1109/34.291440}

\bibitem[{Hussain et~al(2015)Hussain, Raza, Siddiqi, Khurshid, and
  Djeddi}]{Hussain2015ACS}
Hussain R, Raza A, Siddiqi I, et~al (2015) A comprehensive survey of
  handwritten document benchmarks: structure, usage and evaluation. EURASIP
  Journal on Image and Video Processing 2015:1--24

\bibitem[{Ioffe and Szegedy(2015)}]{10.5555/3045118.3045167}
Ioffe S, Szegedy C (2015) {Batch Normalization: Accelerating Deep Network
  Training by Reducing Internal Covariate Shift}. In: Proceedings of the 32nd
  International Conference on International Conference on Machine Learning -
  Volume 37. JMLR.org, ICML'15, p 448–456

\bibitem[{Jocher et~al(2021)Jocher, Stoken, Borovec, NanoCode012,
  ChristopherSTAN, Changyu, Laughing, tkianai, yxNONG, Hogan, lorenzomammana,
  AlexWang1900, Chaurasia, Diaconu, Marc, wanghaoyang0106, ml5ah, Doug,
  Durgesh, Ingham, Frederik, Guilhen, Colmagro, Ye, Jacobsolawetz, Poznanski,
  Fang, Kim, Doan, and 于力军}]{glenn_jocher_2021_4418161}
Jocher G, Stoken A, Borovec J, et~al (2021) {ultralytics/yolov5: v4.0 -
  nn.SiLU() activations, Weights \& Biases logging, PyTorch Hub integration}.
  \doi{10.5281/zenodo.4418161}

\bibitem[{Karatzas et~al(2013)Karatzas, Shafait, Uchida, Iwamura, Bigorda,
  Mestre, Mas, Mota, Almazàn, and de~las Heras}]{6628859}
Karatzas D, Shafait F, Uchida S, et~al (2013) {ICDAR 2013 Robust Reading
  Competition}. In: 2013 12th International Conference on Document Analysis and
  Recognition, pp 1484--1493, \doi{10.1109/ICDAR.2013.221}

\bibitem[{Karatzas et~al(2015)Karatzas, Gomez-Bigorda, Nicolaou, Ghosh,
  Bagdanov, Iwamura, Matas, Neumann, Chandrasekhar, Lu, Shafait, Uchida, and
  Valveny}]{7333942}
Karatzas D, Gomez-Bigorda L, Nicolaou A, et~al (2015) Icdar 2015 competition on
  robust reading. In: {2015 13th International Conference on Document Analysis
  and Recognition (ICDAR)}, pp 1156--1160, \doi{10.1109/ICDAR.2015.7333942}

\bibitem[{Kassis and El-Sana(2014)}]{6981050}
Kassis M, El-Sana J (2014) {Word Spotting Using Radial Descriptor}. In: 2014
  14th International Conference on Frontiers in Handwriting Recognition, pp
  387--392, \doi{10.1109/ICFHR.2014.71}

\bibitem[{Kassis and El-Sana(2016)}]{7814035}
Kassis M, El-Sana J (2016) {Word Spotting Using Radial Descriptor Graph}. In:
  2016 15th International Conference on Frontiers in Handwriting Recognition
  (ICFHR), pp 31--35, \doi{10.1109/ICFHR.2016.0019}

\bibitem[{Kassis et~al(2017)Kassis, Abdalhaleem, Droby, Alaasam, and
  El-Sana}]{8067751}
Kassis M, Abdalhaleem A, Droby A, et~al (2017) {VML-HD: The historical Arabic
  documents dataset for recognition systems}. In: 2017 1st International
  Workshop on Arabic Script Analysis and Recognition (ASAR), pp 11--14,
  \doi{10.1109/ASAR.2017.8067751}

\bibitem[{Kesiman et~al(2016)Kesiman, Burie, Wibawantara, Sunarya, and
  Ogier}]{7814058}
Kesiman MWA, Burie JC, Wibawantara GNMA, et~al (2016) {AMADI\_LontarSet: The
  First Handwritten Balinese Palm Leaf Manuscripts Dataset}. In: 2016 15th
  International Conference on Frontiers in Handwriting Recognition (ICFHR), pp
  168--173, \doi{10.1109/ICFHR.2016.0042}

\bibitem[{Kesiman et~al(2018)Kesiman, Valy, Burie, Paulus, Suryani, Hadi,
  Verleysen, Chhun, and Ogier}]{8583808}
Kesiman MWA, Valy D, Burie JC, et~al (2018) {ICFHR 2018 C}ompetition {O}n
  {D}ocument {I}mage {A}nalysis {T}asks for {S}outheast {A}sian {P}alm {L}eaf
  {M}anuscripts. In: 2018 16th International Conference on Frontiers in
  Handwriting Recognition (ICFHR), pp 483--488,
  \doi{10.1109/ICFHR-2018.2018.00090}

\bibitem[{Kiessling et~al(2019)Kiessling, Ezra, and
  Miller}]{Kiessling2019BADAMAP}
Kiessling B, Ezra DSB, Miller MT (2019) {BADAM: A Public Dataset for Baseline
  Detection in Arabic-script Manuscripts}. Proceedings of the 5th International
  Workshop on Historical Document Imaging and Processing

\bibitem[{Kim et~al(2001)Kim, Jeong, Lee, and Suen}]{953781}
Kim S, Jeong S, Lee GS, et~al (2001) {Word segmentation in handwritten Korean
  text lines based on gap clustering techniques}. In: Proceedings of Sixth
  International Conference on Document Analysis and Recognition, pp 189--193,
  \doi{10.1109/ICDAR.2001.953781}

\bibitem[{Kingma and Welling(2014)}]{Kingma2014AutoEncodingVB}
Kingma DP, Welling M (2014) {Auto-Encoding Variational Bayes}. CoRR
  abs/1312.6114

\bibitem[{Ki{\v{s}}{\v{s}} et~al(2022)Ki{\v{s}}{\v{s}}, Koh{\'u}t, Bene{\v{s}},
  and Hradi{\v{s}}}]{10.1007/978-3-031-06555-2_11}
Ki{\v{s}}{\v{s}} M, Koh{\'u}t J, Bene{\v{s}} K, et~al (2022) {Importance of
  Textlines in Historical Document Classification}. In: Uchida S, Barney E,
  Eglin V (eds) Document Analysis Systems. Springer International Publishing,
  Cham, pp 158--170

\bibitem[{Krizhevsky et~al(2012)Krizhevsky, Sutskever, and
  Hinton}]{NIPS2012_c399862d}
Krizhevsky A, Sutskever I, Hinton GE (2012) Image{N}et {C}lassification with
  {D}eep {C}onvolutional {N}eural {N}etworks. In: Pereira F, Burges CJC, Bottou
  L, et~al (eds) Advances in Neural Information Processing Systems, vol~25.
  Curran Associates, Inc.

\bibitem[{{Kurar Barakat} et~al(2019){Kurar Barakat}, {El-Sana}, and
  {Rabaev}}]{8978129}
{Kurar Barakat} B, {El-Sana} J, {Rabaev} I (2019) The {P}inkas {D}ataset. In:
  2019 International Conference on Document Analysis and Recognition (ICDAR),
  pp 732--737, \doi{10.1109/ICDAR.2019.00122}

\bibitem[{Kusetogullari et~al(2020)Kusetogullari, Yavariabdi, Cheddad, Grahn,
  and Johan}]{kusetogullari2020ardis}
Kusetogullari H, Yavariabdi A, Cheddad A, et~al (2020) {ARDIS: a Swedish
  historical handwritten digit dataset}. Neural computing \& applications
  (Print) 32(21):16,505--16,518

\bibitem[{Kusetogullari et~al(2021)Kusetogullari, Yavariabdi, Hall, and
  Lavesson}]{KUSETOGULLARI2021100182}
Kusetogullari H, Yavariabdi A, Hall J, et~al (2021) {DIGITNET: A D}eep
  {H}andwritten {D}igit {D}etection and {R}ecognition {M}ethod {U}sing a {N}ew
  {H}istorical {H}andwritten {D}igit {D}ataset. Big Data Research 23:100,182.
  \doi{https://doi.org/10.1016/j.bdr.2020.100182}

\bibitem[{Lai and Jin(2019)}]{8978107}
Lai S, Jin L (2019) {Offline Writer Identification Based on the Path Signature
  Feature}. In: 2019 International Conference on Document Analysis and
  Recognition (ICDAR), pp 1137--1142, \doi{10.1109/ICDAR.2019.00184}

\bibitem[{Lang et~al(2018)Lang, Puigcerver, Toselli, and Vidal}]{8563224}
Lang E, Puigcerver J, Toselli AH, et~al (2018) {Probabilistic Indexing and
  Search for Information Extraction on Handwritten German Parish Records}. In:
  2018 16th International Conference on Frontiers in Handwriting Recognition
  (ICFHR), pp 44--49, \doi{10.1109/ICFHR-2018.2018.00017}

\bibitem[{LeCun(1998)}]{lecun1998mnist}
LeCun Y (1998) The {MNIST} database of handwritten digits. http://yann lecun
  com/exdb/mnist/

\bibitem[{Lee et~al(2020)Lee, Mears, Jakeway, Ferriter, Adams, Yarasavage,
  Thomas, Zwaard, and Weld}]{10.1145/3340531.3412767}
Lee BCG, Mears J, Jakeway E, et~al (2020) The Newspaper Navigator Dataset:
  Extracting Headlines and Visual Content from 16 Million Historic Newspaper
  Pages in Chronicling America, Association for Computing Machinery, New York,
  NY, USA, p 3055–3062.
  \urlprefix\url{https://doi.org/10.1145/3340531.3412767}

\bibitem[{Leydier et~al(2007)Leydier, Lebourgeois, and
  Emptoz}]{Leydier2007TextSF}
Leydier Y, Lebourgeois F, Emptoz H (2007) Text search for medieval manuscript
  images. Pattern Recognit 40:3552--3567

\bibitem[{Leydier et~al(2009)Leydier, Ouji, Lebourgeois, and
  Emptoz}]{Leydier2009TowardsAO}
Leydier Y, Ouji A, Lebourgeois F, et~al (2009) Towards an omnilingual word
  retrieval system for ancient manuscripts. Pattern Recognit 42:2089--2105

\bibitem[{Likforman-Sulem et~al(2006)Likforman-Sulem, Zahour, and
  Taconet}]{LikformanSulem2006TextLS}
Likforman-Sulem L, Zahour A, Taconet B (2006) Text line segmentation of
  historical documents: a survey. International Journal of Document Analysis
  and Recognition (IJDAR) 9:123--138

\bibitem[{Lin et~al(2014)Lin, Maire, Belongie, Hays, Perona, Ramanan,
  Doll{\'a}r, and Zitnick}]{Lin2014MicrosoftCC}
Lin TY, Maire M, Belongie SJ, et~al (2014) Microsoft {COCO: C}ommon {O}bjects
  in {C}ontext. In: ECCV

\bibitem[{Lin et~al(2017)Lin, Goyal, Girshick, He, and
  Doll{\'a}r}]{lin2017focal}
Lin TY, Goyal P, Girshick R, et~al (2017) Focal loss for dense object
  detection. In: Proceedings of the IEEE international conference on computer
  vision, pp 2980--2988

\bibitem[{Lombardi and Marinai(2020)}]{jimaging6100110}
Lombardi F, Marinai S (2020) Deep {L}earning for {H}istorical {D}ocument
  {A}nalysis and {R}ecognition—{A S}urvey. Journal of Imaging 6(10).
  \doi{10.3390/jimaging6100110}

\bibitem[{Louloudis et~al(2009)Louloudis, Gatos, Pratikakis, and
  Halatsis}]{10.1016/j.patcog.2008.12.016}
Louloudis G, Gatos B, Pratikakis I, et~al (2009) {Text Line and Word
  Segmentation of Handwritten Documents}. Pattern Recogn 42(12):3169–3183.
  \doi{10.1016/j.patcog.2008.12.016}

\bibitem[{LoweDavid(2004)}]{LoweDavid2004DistinctiveIF}
LoweDavid G (2004) {Distinctive Image Features from Scale-Invariant Keypoints}.
  International Journal of Computer Vision

\bibitem[{Maarand et~al(2022)Maarand, Beyer, K{\aa}sen, Fosseide, and
  Kermorvant}]{10.1007/978-3-031-06555-2_27}
Maarand M, Beyer Y, K{\aa}sen A, et~al (2022) {A Comprehensive Comparison
  of Open-Source Libraries for Handwritten Text Recognition in Norwegian}.
  In: Uchida S, Barney E, Eglin V (eds) Document Analysis Systems. Springer
  International Publishing, Cham, pp 399--413

\bibitem[{Marinai et~al(2005)Marinai, Gori, and Soda}]{1359749}
Marinai S, Gori M, Soda G (2005) Artificial neural networks for document
  analysis and recognition. IEEE Transactions on Pattern Analysis and Machine
  Intelligence 27(1):23--35. \doi{10.1109/TPAMI.2005.4}

\bibitem[{Mark et~al(2021)Mark, Denis, Alex, Vladimir, Denis, Maxim, and
  Andrey}]{10.1145/3476887.3476892}
Mark P, Denis D, Alex S, et~al (2021) Digital Peter: New Dataset, Competition
  and Handwriting Recognition Methods, Association for Computing Machinery, New
  York, NY, USA, p 43–48. \doi{https://doi.org/10.1145/3476887.3476892}

\bibitem[{Marti and Bunke(2001)}]{10.5555/505741.505745}
Marti UV, Bunke H (2001) Using a Statistical Language Model to Improve the
  Performance of an HMM-Based Cursive Handwriting Recognition Systems, World
  Scientific Publishing Co., Inc., USA, p 65–90

\bibitem[{Mehri et~al(2015)Mehri, Gomez-Kr{\"a}mer, H{\'e}roux, Boucher, and
  Mullot}]{Mehri2015ATP}
Mehri M, Gomez-Kr{\"a}mer P, H{\'e}roux P, et~al (2015) A texture-based pixel
  labeling approach for historical books. Pattern Analysis and Applications
  20:325--364

\bibitem[{Mehri et~al(2017)Mehri, H{\'e}roux, Mullot, Moreux, Co{\"u}asnon, and
  Barrett}]{mehri:hal-01637826}
Mehri M, H{\'e}roux P, Mullot R, et~al (2017) {HBA 1.0: A Pixel-based Annotated
  Dataset for Historical Book Analysis}. In: {International Workshop on
  Historical Document Imaging and Processing (HIP)}, Kyoto, Japan,
  \urlprefix\url{https://hal.archives-ouvertes.fr/hal-01637826}

\bibitem[{Mehri et~al(2019)Mehri, Héroux, Mullot, Moreux, Coüasnon, and
  Barrett}]{8978192}
Mehri M, Héroux P, Mullot R, et~al (2019) {ICDAR2019 C}ompetition on
  {H}istorical {B}ook {A}nalysis - {HBA2019}. In: 2019 International Conference
  on Document Analysis and Recognition (ICDAR), pp 1488--1493,
  \doi{10.1109/ICDAR.2019.00239}

\bibitem[{Merabti et~al(2018)Merabti, Farou, and
  Seridi}]{merabti2018segmentation}
Merabti H, Farou B, Seridi H (2018) A segmentation-recognition approach with a
  fuzzy-artificial immune system for unconstrained handwritten connected
  digits. Informatica 42(1):95--106

\bibitem[{Mohammed et~al(2017)Mohammed, M{\"a}ergner, Konidaris, and
  Stiehl}]{mohammed2017normalised}
Mohammed H, M{\"a}ergner V, Konidaris T, et~al (2017) Normalised local
  {N}a{\"\i}ve {B}ayes nearest-neighbour classifier for offline writer
  identification. In: 2017 14th IAPR International Conference on Document
  Analysis and Recognition (ICDAR), IEEE, pp 1013--1018

\bibitem[{{Mohammed} et~al(2019){Mohammed}, {Marthot-Santaniello}, and
  {Märgner}}]{8978142}
{Mohammed} H, {Marthot-Santaniello} I, {Märgner} V (2019) {GRK-Papyri}: A
  dataset of {G}reek handwriting on papyri for the task of writer
  identification. In: 2019 International Conference on Document Analysis and
  Recognition (ICDAR), pp 726--731, \doi{10.1109/ICDAR.2019.00121}

\bibitem[{Monnier and Aubry(2020)}]{monnier2020docExtractor}
Monnier T, Aubry M (2020) {docExtractor: An Off-the-Shelf Historical Document
  Element Extraction}. In: ICFHR

\bibitem[{Mota et~al(2014)Mota, Llad{\'o}s, and Forn{\'e}s}]{Mota2014AGA}
Mota DF, Llad{\'o}s J, Forn{\'e}s A (2014) A graph-based approach for
  segmenting touching lines in historical handwritten documents. International
  Journal on Document Analysis and Recognition (IJDAR) 17:293--312

\bibitem[{Namboodiri and Jain(2007)}]{Namboodiri2007}
Namboodiri AM, Jain AK (2007) Document Structure and Layout Analysis, Springer
  London, London, pp 29--48. \doi{10.1007/978-1-84628-726-8_2}

\bibitem[{Neudecker et~al(2021)Neudecker, Baierer, Gerber, Christian,
  Apostolos, and Stefan}]{10.1145/3476887.3476888}
Neudecker C, Baierer K, Gerber M, et~al (2021) {A Survey of OCR Evaluation
  Tools and Metrics}, Association for Computing Machinery, New York, NY, USA, p
  13–18. \urlprefix\url{https://doi.org/10.1145/3476887.3476888}

\bibitem[{Newell and Griffin(2014)}]{NEWELL20142255}
Newell AJ, Griffin LD (2014) Writer identification using oriented {B}asic
  {I}mage {F}eatures and the {D}elta encoding. Pattern Recognition
  47(6):2255--2265. \doi{https://doi.org/10.1016/j.patcog.2013.11.029}

\bibitem[{Nicolaou and Gatos(2009)}]{5277573}
Nicolaou A, Gatos B (2009) {Handwritten Text Line Segmentation by Shredding
  Text into its Lines}. In: 2009 10th International Conference on Document
  Analysis and Recognition, pp 626--630, \doi{10.1109/ICDAR.2009.243}

\bibitem[{Nikolaou et~al(2010)Nikolaou, Makridis, Gatos, Stamatopoulos, and
  Papamarkos}]{Nikolaou2010SegmentationOH}
Nikolaou NA, Makridis M, Gatos B, et~al (2010) Segmentation of historical
  machine-printed documents using {A}daptive {R}un {L}ength {S}moothing and
  skeleton segmentation paths. Image Vis Comput 28:590--604

\bibitem[{Nina(2018)}]{Nina2018NephiA}
Nina OA (2018) Nephi : An {O}pen {S}ource {P}ytorch {L}ibrary for {H}andwriting
  {R}ecognition

\bibitem[{Oliveira et~al(2018)Oliveira, Seguin, and
  Kaplan}]{Oliveira2018dhSegmentAG}
Oliveira SA, Seguin B, Kaplan F (2018) dh{S}egment: {A G}eneric
  {D}eep-{L}earning {A}pproach for {D}ocument {S}egmentation. 2018 16th
  International Conference on Frontiers in Handwriting Recognition (ICFHR) pp
  7--12

\bibitem[{Pantke et~al(2013)Pantke, Märgner, and Fingscheidt}]{6628824}
Pantke W, Märgner V, Fingscheidt T (2013) {On Evaluation of Segmentation-Free
  Word Spotting Approaches without Hard Decisions}. In: 2013 12th International
  Conference on Document Analysis and Recognition, pp 1300--1304,
  \doi{10.1109/ICDAR.2013.263}

\bibitem[{Pantke et~al(2014)Pantke, Dennhardt, Fecker, Märgner, and
  Fingscheidt}]{6980990}
Pantke W, Dennhardt M, Fecker D, et~al (2014) An {H}istorical {H}andwritten
  {A}rabic {D}ataset for {S}egmentation-{F}ree {W}ord {S}potting - {HADARA80P}.
  In: 2014 14th International Conference on Frontiers in Handwriting
  Recognition, pp 15--20, \doi{10.1109/ICFHR.2014.11}

\bibitem[{Papadopoulos et~al(2013)Papadopoulos, Pletschacher, Clausner, and
  Antonacopoulos}]{10.1145/2501115.2501130}
Papadopoulos C, Pletschacher S, Clausner C, et~al (2013) The {IMPACT D}ataset
  of {H}istorical {D}ocument {I}mages. In: Proceedings of the 2nd International
  Workshop on Historical Document Imaging and Processing. Association for
  Computing Machinery, New York, NY, USA, HIP '13, p 123–130,
  \doi{10.1145/2501115.2501130}

\bibitem[{Perronnin and Rodriguez-Serrano(2009)}]{5277774}
Perronnin F, Rodriguez-Serrano JA (2009) Fisher {K}ernels for {H}andwritten
  {W}ord-spotting. In: 2009 10th International Conference on Document Analysis
  and Recognition, pp 106--110, \doi{10.1109/ICDAR.2009.16}

\bibitem[{Plamondon and Srihari(2000)}]{824821}
Plamondon R, Srihari S (2000) {Online and off-line handwriting recognition: a
  comprehensive survey}. IEEE Transactions on Pattern Analysis and Machine
  Intelligence 22(1):63--84. \doi{10.1109/34.824821}

\bibitem[{Pletschacher and Antonacopoulos(2010)}]{5597587}
Pletschacher S, Antonacopoulos A (2010) {The PAGE (Page Analysis and
  Ground-Truth Elements) Format Framework}. In: 2010 20th International
  Conference on Pattern Recognition, pp 257--260, \doi{10.1109/ICPR.2010.72}

\bibitem[{Prasad et~al(2019)Prasad, Déjean, and Meunier}]{8978117}
Prasad A, Déjean H, Meunier JL (2019) {Versatile Layout Understanding via
  Conjugate Graph}. In: 2019 International Conference on Document Analysis and
  Recognition (ICDAR), pp 287--294, \doi{10.1109/ICDAR.2019.00054}

\bibitem[{Prasad et~al(2020)Prasad, Gadpal, Kapadni, Visave, and
  Sultanpure}]{Prasad2020CascadeTabNetAA}
Prasad D, Gadpal A, Kapadni K, et~al (2020) Cascade{T}ab{N}et: An approach for
  end to end table detection and structure recognition from image-based
  documents. 2020 IEEE/CVF Conference on Computer Vision and Pattern
  Recognition Workshops (CVPRW) pp 2439--2447

\bibitem[{Pratikakis et~al(2011)Pratikakis, Gatos, and Ntirogiannis}]{6065249}
Pratikakis I, Gatos B, Ntirogiannis K (2011) {ICDAR 2011 Document Image
  Binarization Contest (DIBCO 2011)}. In: 2011 International Conference on
  Document Analysis and Recognition, pp 1506--1510,
  \doi{10.1109/ICDAR.2011.299}

\bibitem[{Pratikakis et~al(2013)Pratikakis, Gatos, and Ntirogiannis}]{6628857}
Pratikakis I, Gatos B, Ntirogiannis K (2013) {ICDAR 2013 Document Image
  Binarization Contest (DIBCO 2013)}. In: 2013 12th International Conference on
  Document Analysis and Recognition, pp 1471--1476,
  \doi{10.1109/ICDAR.2013.219}

\bibitem[{Pratikakis et~al(2017)Pratikakis, Zagoris, Barlas, and
  Gatos}]{8270159}
Pratikakis I, Zagoris K, Barlas G, et~al (2017) {ICDAR2017 Competition on
  Document Image Binarization (DIBCO 2017)}. In: 2017 14th IAPR International
  Conference on Document Analysis and Recognition (ICDAR), pp 1395--1403,
  \doi{10.1109/ICDAR.2017.228}

\bibitem[{Pratikakis et~al(2018)Pratikakis, Zagori, Kaddas, and
  Gatos}]{8583809}
Pratikakis I, Zagori K, Kaddas P, et~al (2018) {ICFHR 2018 C}ompetition on
  {H}andwritten {D}ocument {I}mage {B}inarization {(H-DIBCO 2018)}. In: 2018
  16th International Conference on Frontiers in Handwriting Recognition
  (ICFHR), pp 489--493, \doi{10.1109/ICFHR-2018.2018.00091}

\bibitem[{Pratikakis et~al(2019)Pratikakis, Zagoris, Karagiannis, Tsochatzidis,
  Mondal, and Marthot-Santaniello}]{8978205}
Pratikakis I, Zagoris K, Karagiannis X, et~al (2019) {ICDAR 2019 Competition on
  Document Image Binarization (DIBCO 2019)}. In: 2019 International Conference
  on Document Analysis and Recognition (ICDAR), pp 1547--1556,
  \doi{10.1109/ICDAR.2019.00249}

\bibitem[{Puigcerver and Mocholí(2018)}]{puigcerver2018pylaia}
Puigcerver J, Mocholí C (2018) Pylaia.
  \url{https://github.com/jpuigcerver/PyLaia}

\bibitem[{Pérez et~al(2009)Pérez, Tarazón, Serrano, Castro, Terrades, and
  Juan}]{5277691}
Pérez D, Tarazón L, Serrano N, et~al (2009) The {GERMANA D}atabase. In: 2009
  10th International Conference on Document Analysis and Recognition, pp
  301--305, \doi{10.1109/ICDAR.2009.10}

\bibitem[{Quir{\'o}s(2018)}]{Quirs2018MultiTaskHD}
Quir{\'o}s L (2018) {Multi-Task Handwritten Document Layout Analysis}. ArXiv
  abs/1806.08852

\bibitem[{Quir{\'o}s et~al(2020)Quir{\'o}s, Kallio, and
  Vidal}]{Quirs2020FinnishCR}
Quir{\'o}s L, Kallio M, Vidal E (2020) {Finnish Court Records-sub500. A dataset
  of Finnish notarial records (19th Century)}

\bibitem[{Quirós et~al(2021)Quirós, Vidal, Sánchez, and Villarreal}]{vorau}
Quirós L, Vidal E, Sánchez JA, et~al (2021) {Vorau Abbey library Cod. 253
  dataset for Document Layout Analysis}.
  \urlprefix\url{https://zenodo.org/record/5443258#.YpoMti8RqJ8}

\bibitem[{Rath and Manmatha(2006)}]{Rath2006WordSF}
Rath TM, Manmatha R (2006) Word spotting for historical documents.
  International Journal of Document Analysis and Recognition (IJDAR) 9:139--152

\bibitem[{Redmon and Farhadi(2018)}]{Redmon2018YOLOv3AI}
Redmon J, Farhadi A (2018) {YOLO}v3: An {I}ncremental {I}mprovement. ArXiv
  abs/1804.02767

\bibitem[{Ren et~al(2015)Ren, He, Girshick, and Sun}]{NIPS2015_14bfa6bb}
Ren S, He K, Girshick R, et~al (2015) Faster {R-CNN: T}owards {R}eal-{T}ime
  {O}bject {D}etection with {R}egion {P}roposal {N}etworks. In: Cortes C,
  Lawrence N, Lee D, et~al (eds) Advances in Neural Information Processing
  Systems, vol~28. Curran Associates, Inc.

\bibitem[{Rezende et~al(2014)Rezende, Mohamed, and
  Wierstra}]{JimenezRezende2014StochasticBA}
Rezende DJ, Mohamed S, Wierstra D (2014) {Stochastic Backpropagation and
  Approximate Inference in Deep Generative Models}. In: ICML

\bibitem[{Rodr\'{\i}guez-Serrano and
  Perronnin(2009)}]{10.1016/j.patcog.2009.02.005}
Rodr\'{\i}guez-Serrano JA, Perronnin F (2009) {Handwritten Word-Spotting Using
  Hidden Markov Models and Universal Vocabularies}. Pattern Recogn
  42(9):2106–2116. \doi{10.1016/j.patcog.2009.02.005}

\bibitem[{Romero and Sánchez(2021)}]{9412210}
Romero V, Sánchez JA (2021) {The HisClima database: historical weather logs
  for automatic transcription and information extraction}. In: 2020 25th
  International Conference on Pattern Recognition (ICPR), pp 10,141--10,148,
  \doi{10.1109/ICPR48806.2021.9412210}

\bibitem[{Romero et~al(2013)Romero, Fornés, Serrano, Sánchez, Toselli,
  Frinken, Vidal, and Lladós}]{ROMERO20131658}
Romero V, Fornés A, Serrano N, et~al (2013) The {ESPOSALLES} database: An
  ancient marriage license corpus for off-line handwriting recognition. Pattern
  Recognition 46(6):1658--1669.
  \doi{https://doi.org/10.1016/j.patcog.2012.11.024}

\bibitem[{Ronneberger et~al(2015)Ronneberger, Fischer, and
  Brox}]{10.1007/978-3-319-24574-4_28}
Ronneberger O, Fischer P, Brox T (2015) U-{N}et: {C}onvolutional {N}etworks for
  {B}iomedical {I}mage {S}egmentation. In: Navab N, Hornegger J, Wells WM,
  et~al (eds) Medical Image Computing and Computer-Assisted Intervention --
  MICCAI 2015. Springer International Publishing, Cham, pp 234--241

\bibitem[{Rusi{\~n}ol et~al(2011)Rusi{\~n}ol, Aldavert, Toledo, and
  Llad{\'o}s}]{Rusiol2011BrowsingHD}
Rusi{\~n}ol M, Aldavert D, Toledo R, et~al (2011) Browsing {H}eterogeneous
  {D}ocument {C}ollections by a {S}egmentation-{F}ree {W}ord {S}potting
  {M}ethod. 2011 International Conference on Document Analysis and Recognition
  pp 63--67

\bibitem[{Russakovsky et~al(2015)Russakovsky, Deng, Su, Krause, Satheesh, Ma,
  Huang, Karpathy, Khosla, Bernstein, Berg, and
  Fei-Fei}]{russakovsky2015imagenet}
Russakovsky O, Deng J, Su H, et~al (2015) {ImageNet Large Scale Visual
  Recognition Challenge}. International Journal of Computer Vision 115:211--252

\bibitem[{Saini et~al(2019)Saini, Dobson, Morrey, Liwicki, and
  Simistira~Liwicki}]{8977999}
Saini R, Dobson D, Morrey J, et~al (2019) {ICDAR 2019 H}istorical {D}ocument
  {R}eading {C}hallenge on {L}arge {S}tructured {C}hinese {F}amily {R}ecords.
  In: 2019 International Conference on Document Analysis and Recognition
  (ICDAR), pp 1499--1504, \doi{10.1109/ICDAR.2019.00241}

\bibitem[{S{\'a}nchez et~al(2016)S{\'a}nchez, Romero, Toselli, and
  Vidal}]{Snchez2016ICFHR2016CO}
S{\'a}nchez JA, Romero V, Toselli AH, et~al (2016) {ICFHR2016 Competition on
  Handwritten Text Recognition on the READ Dataset}. 2016 15th International
  Conference on Frontiers in Handwriting Recognition (ICFHR) pp 630--635

\bibitem[{Sandler et~al(2018)Sandler, Howard, Zhu, Zhmoginov, and
  Chen}]{sandler2019mobilenetv2}
Sandler M, Howard AG, Zhu M, et~al (2018) {MobileNetV2: Inverted Residuals and
  Linear Bottlenecks}. 2018 IEEE/CVF Conference on Computer Vision and Pattern
  Recognition pp 4510--4520

\bibitem[{Schubert et~al(2017)Schubert, Sander, Ester, Kriegel, and
  Xu}]{10.1145/3068335}
Schubert E, Sander J, Ester M, et~al (2017) {DBSCAN R}evisited, {R}evisited:
  {W}hy and {H}ow {Y}ou {S}hould ({S}till) {U}se {DBSCAN}. ACM Trans Database
  Syst 42(3). \doi{10.1145/3068335}

\bibitem[{Serrano et~al(2010)Serrano, Castro, and
  Juan}]{serrano-etal-2010-rodrigo}
Serrano N, Castro F, Juan A (2010) The {RODRIGO} database. In: Proceedings of
  the Seventh International Conference on Language Resources and Evaluation
  ({LREC}'10). European Language Resources Association (ELRA), Valletta, Malta,
  \urlprefix\url{http://www.lrec-conf.org/proceedings/lrec2010/pdf/477_Paper.pdf}

\bibitem[{Seuret et~al(2016)Seuret, Ingold, and Liwicki}]{7814107}
Seuret M, Ingold R, Liwicki M (2016) N-light-{N}: A {H}ighly-{A}daptable {J}ava
  {L}ibrary for {D}ocument {A}nalysis with {C}onvolutional {A}uto-{E}ncoders
  and {R}elated {A}rchitectures. In: 2016 15th International Conference on
  Frontiers in Handwriting Recognition (ICFHR), pp 459--464,
  \doi{10.1109/ICFHR.2016.0091}

\bibitem[{Seuret et~al(2019)Seuret, Limbach, Weichselbaumer, Maier, and
  Christlein}]{10.1145/3352631.3352640}
Seuret M, Limbach S, Weichselbaumer N, et~al (2019) Dataset of {P}ages from
  {E}arly {P}rinted {B}ooks with {M}ultiple {F}ont {G}roups. In: Proceedings of
  the 5th International Workshop on Historical Document Imaging and Processing.
  Association for Computing Machinery, New York, NY, USA, HIP '19, p 1–6,
  \doi{10.1145/3352631.3352640}

\bibitem[{Seuret et~al(2020)Seuret, Nicolaou, Stutzmann, Maier, and
  Christlein}]{Seuret2020ICFHR2C}
Seuret M, Nicolaou A, Stutzmann D, et~al (2020) {ICFHR 2020 C}ompetition on
  {I}mage {R}etrieval for {H}istorical {H}andwritten {F}ragments. 2020 17th
  International Conference on Frontiers in Handwriting Recognition (ICFHR) pp
  216--221

\bibitem[{Seuret et~al(2021)Seuret, Nicolaou, Rodr{\'i}guez-Salas,
  Weichselbaumer, Stutzmann, Mayr, Maier, and
  Christlein}]{10.1007/978-3-030-86337-1_41}
Seuret M, Nicolaou A, Rodr{\'i}guez-Salas D, et~al (2021) {ICDAR} 2021
  {Competition} on {H}istorical {Document} {Classification}. In: Llad{\'o}s J,
  Lopresti D, Uchida S (eds) Document Analysis and Recognition -- ICDAR 2021.
  Springer International Publishing, Cham, pp 618--634

\bibitem[{Shahkolaei et~al(2018)Shahkolaei, Beghdadi, Al-maadeed, and
  Cheriet}]{8480372}
Shahkolaei A, Beghdadi A, Al-maadeed S, et~al (2018) {MHDID: A
  M}ulti-distortion {H}istorical {D}ocument {I}mage {D}atabase. In: 2018 IEEE
  2nd International Workshop on Arabic and Derived Script Analysis and
  Recognition (ASAR), pp 156--160, \doi{10.1109/ASAR.2018.8480372}

\bibitem[{Shao et~al(2012)Shao, Wang, and Xiao}]{Shao2012FastSV}
Shao Y, Wang C, Xiao B (2012) Fast self-generation voting for handwritten
  chinese character recognition. International Journal on Document Analysis and
  Recognition (IJDAR) 16:413--424

\bibitem[{Shen et~al(2020)Shen, Zhang, and Dell}]{Shen_2020_CVPR_Workshops}
Shen Z, Zhang K, Dell M (2020) {A Large Dataset of Historical Japanese
  Documents With Complex Layouts}. In: Proceedings of the IEEE/CVF Conference
  on Computer Vision and Pattern Recognition (CVPR) Workshops

\bibitem[{Shi et~al(2017)Shi, Bai, and Yao}]{Shi2017AnET}
Shi B, Bai X, Yao C (2017) An {End-to-End Trainable Neural Network for
  Image-Based Sequence Recognition and Its Application to Scene Text
  Recognition}. IEEE Transactions on Pattern Analysis and Machine Intelligence
  39:2298--2304

\bibitem[{{Simistira} et~al(2016){Simistira}, {Seuret}, {Eichenberger}, {Garz},
  {Liwicki}, and {Ingold}}]{7814109}
{Simistira} F, {Seuret} M, {Eichenberger} N, et~al (2016) {DIVA-HisDB: A
  Precisely Annotated Large Dataset of Challenging Medieval Manuscripts}. In:
  2016 15th International Conference on Frontiers in Handwriting Recognition
  (ICFHR), pp 471--476, \doi{10.1109/ICFHR.2016.0093}

\bibitem[{Simistira et~al(2017)Simistira, Bouillon, Seuret, Würsch, Alberti,
  Ingold, and Liwicki}]{8270154}
Simistira F, Bouillon M, Seuret M, et~al (2017) {ICDAR2017 C}ompetition on
  {L}ayout {A}nalysis for {C}hallenging {M}edieval {M}anuscripts. In: 2017 14th
  IAPR International Conference on Document Analysis and Recognition (ICDAR),
  pp 1361--1370, \doi{10.1109/ICDAR.2017.223}

\bibitem[{Simonyan and Zisserman(2015)}]{Simonyan2015VeryDC}
Simonyan K, Zisserman A (2015) Very {Deep Convolutional Networks for
  Large-Scale Image Recognition}. CoRR abs/1409.1556

\bibitem[{Sudholt and Fink(2016)}]{Sudholt2016PHOCNetAD}
Sudholt S, Fink GA (2016) {PHOCNet: A Deep Convolutional Neural Network for
  Word Spotting in Handwritten Documents}. 2016 15th International Conference
  on Frontiers in Handwriting Recognition (ICFHR) pp 277--282

\bibitem[{Suryani et~al(2017)Suryani, Paulus, Hadi, Darsa, and Burie}]{8270066}
Suryani M, Paulus E, Hadi S, et~al (2017) {The Handwritten Sundanese Palm Leaf
  Manuscript Dataset from 15th Century}. In: 2017 14th IAPR International
  Conference on Document Analysis and Recognition (ICDAR), pp 796--800,
  \doi{10.1109/ICDAR.2017.135}

\bibitem[{Szegedy et~al(2015)Szegedy, Liu, Jia, Sermanet, Reed, Anguelov,
  Erhan, Vanhoucke, and Rabinovich}]{7298594}
Szegedy C, Liu W, Jia Y, et~al (2015) Going deeper with convolutions. In: 2015
  IEEE Conference on Computer Vision and Pattern Recognition (CVPR), pp 1--9,
  \doi{10.1109/CVPR.2015.7298594}

\bibitem[{Sánchez et~al(2014)Sánchez, Romero, Toselli, and Vidal}]{6981116}
Sánchez JA, Romero V, Toselli AH, et~al (2014) {ICFHR2014 Competition on
  Handwritten Text Recognition on Transcriptorium Datasets (HTRtS)}. In: 2014
  14th International Conference on Frontiers in Handwriting Recognition, pp
  785--790, \doi{10.1109/ICFHR.2014.137}

\bibitem[{Sánchez et~al(2015)Sánchez, Toselli, Romero, and Vidal}]{7333944}
Sánchez JA, Toselli AH, Romero V, et~al (2015) {ICDAR 2015 competition HTRtS:
  Handwritten Text Recognition on the tranScriptorium dataset}. In: 2015 13th
  International Conference on Document Analysis and Recognition (ICDAR), pp
  1166--1170, \doi{10.1109/ICDAR.2015.7333944}

\bibitem[{Sánchez et~al(2017)Sánchez, Romero, Toselli, Villegas, and
  Vidal}]{8270157}
Sánchez JA, Romero V, Toselli AH, et~al (2017) {ICDAR2017 Competition on
  Handwritten Text Recognition on the READ Dataset}. In: 2017 14th IAPR
  International Conference on Document Analysis and Recognition (ICDAR), pp
  1383--1388, \doi{10.1109/ICDAR.2017.226}

\bibitem[{Sánchez et~al(2019)Sánchez, Romero, Toselli, Villegas, and
  Vidal}]{SANCHEZ2019122}
Sánchez JA, Romero V, Toselli AH, et~al (2019) {A set of benchmarks for
  Handwritten Text Recognition on historical documents}. Pattern Recognition
  94:122--134. \doi{https://doi.org/10.1016/j.patcog.2019.05.025}

\bibitem[{Tang et~al(1996)Tang, Lee, and Suen}]{Tang1996AutomaticDP}
Tang YY, Lee SW, Suen CY (1996) Automatic document processing: A survey.
  Pattern Recognit 29:1931--1952

\bibitem[{Toselli et~al(2007)Toselli, Romero, Rodriguez, and Vidal}]{4377054}
Toselli A, Romero V, Rodriguez L, et~al (2007) {Computer Assisted Transcription
  of Handwritten Text Images}. In: Ninth International Conference on Document
  Analysis and Recognition (ICDAR 2007), pp 944--948,
  \doi{10.1109/ICDAR.2007.4377054}

\bibitem[{Toselli et~al(2004)Toselli, Juan-C{\'i}scar, Gonz{\'a}lez, Salvador,
  Vidal, Casacuberta, Keysers, and Ney}]{Toselli2004IntegratedHR}
Toselli AH, Juan-C{\'i}scar A, Gonz{\'a}lez J, et~al (2004) {Integrated
  Handwriting Recognition And Interpretation Using Finite-State Models}. Int J
  Pattern Recognit Artif Intell 18:519--539

\bibitem[{Valy et~al(2017)Valy, Verleysen, Chhun, and
  Burie}]{10.1145/3151509.3151510}
Valy D, Verleysen M, Chhun S, et~al (2017) A {N}ew {K}hmer {P}alm {L}eaf
  {M}anuscript {D}ataset for {D}ocument {A}nalysis and {R}ecognition: Sleukrith
  {S}et. In: Proceedings of the 4th International Workshop on Historical
  Document Imaging and Processing. Association for Computing Machinery, New
  York, NY, USA, HIP2017, p 1–6, \doi{10.1145/3151509.3151510}

\bibitem[{Verma et~al(2019)Verma, Lamb, Beckham, Najafi, Mitliagkas, Lopez-Paz,
  and Bengio}]{verma2018manifold}
Verma V, Lamb A, Beckham C, et~al (2019) {Manifold Mixup: Better
  Representations by Interpolating Hidden States}. In: ICML

\bibitem[{Vinciarelli and Bengio(2002)}]{Vinciarelli2002OfflineCW}
Vinciarelli A, Bengio S (2002) Offline cursive word recognition using
  continuous density hidden markov models trained with {PCA or ICA} features.
  Object recognition supported by user interaction for service robots 3:81--84
  vol.3

\bibitem[{Voigtlaender et~al(2016)Voigtlaender, Doetsch, and Ney}]{7814068}
Voigtlaender P, Doetsch P, Ney H (2016) {Handwriting Recognition with Large
  Multidimensional Long Short-Term Memory Recurrent Neural Networks}. In: 2016
  15th International Conference on Frontiers in Handwriting Recognition
  (ICFHR), pp 228--233, \doi{10.1109/ICFHR.2016.0052}

\bibitem[{Wolf et~al(2002)Wolf, Jolion, and Chassaing}]{1048482}
Wolf C, Jolion JM, Chassaing F (2002) Text localization, enhancement and
  binarization in multimedia documents. In: 2002 International Conference on
  Pattern Recognition, pp 1037--1040 vol.2, \doi{10.1109/ICPR.2002.1048482}

\bibitem[{Wu et~al(2019)Wu, Kirillov, Massa, Lo, and
  Girshick}]{wu2019detectron2}
Wu Y, Kirillov A, Massa F, et~al (2019) Detectron2.
  \url{https://github.com/facebookresearch/detectron2}

\bibitem[{Xie et~al(2017)Xie, Girshick, Doll{\'a}r, Tu, and
  He}]{Xie2017AggregatedRT}
Xie S, Girshick RB, Doll{\'a}r P, et~al (2017) Aggregated residual
  transformations for deep neural networks. 2017 IEEE Conference on Computer
  Vision and Pattern Recognition (CVPR) pp 5987--5995

\bibitem[{Xu et~al(2019)Xu, Yin, Wang, Zhang, Zhang, and Liu}]{8978010}
Xu Y, Yin F, Wang DH, et~al (2019) {CASIA-AHCDB: A Large-Scale Chinese Ancient
  Handwritten Characters Database}. In: 2019 International Conference on
  Document Analysis and Recognition (ICDAR), pp 793--798,
  \doi{10.1109/ICDAR.2019.00132}

\bibitem[{Yang et~al(2018{\natexlab{a}})Yang, Jin, Huang, Yang, Lai, and
  Sun}]{Yang2018DenseAT}
Yang H, Jin L, Huang W, et~al (2018{\natexlab{a}}) {Dense and Tight Detection
  of Chinese Characters in Historical Documents: Datasets and a Recognition
  Guided Detector}. IEEE Access 6:30,174--30,183

\bibitem[{Yang et~al(2018{\natexlab{b}})Yang, Zhang, Yin, and
  Liu}]{Yang2018RobustCW}
Yang HM, Zhang XY, Yin F, et~al (2018{\natexlab{b}}) {Robust Classification
  with Convolutional Prototype Learning}. 2018 IEEE/CVF Conference on Computer
  Vision and Pattern Recognition pp 3474--3482

\bibitem[{Zhang et~al(2018)Zhang, Ciss{\'e}, Dauphin, and
  Lopez-Paz}]{Zhang2018mixupBE}
Zhang H, Ciss{\'e} M, Dauphin Y, et~al (2018) {mixup: Beyond Empirical Risk
  Minimization}. ArXiv abs/1710.09412

\bibitem[{Zhang et~al(2017)Zhang, Bengio, and Liu}]{Zhang2017OnlineAO}
Zhang XY, Bengio Y, Liu CL (2017) Online and offline handwritten {Chinese}
  character recognition: A comprehensive study and new benchmark. Pattern
  Recognit 61:348--360

\bibitem[{Zhong et~al(2019)Zhong, Tang, and Yepes}]{zhong2019publaynet}
Zhong X, Tang J, Yepes AJ (2019) {PubLayNet: Largest Dataset Ever for Document
  Layout Analysis}. In: 2019 International Conference on Document Analysis and
  Recognition (ICDAR), IEEE, pp 1015--1022, \doi{10.1109/ICDAR.2019.00166}

\bibitem[{Zhuang(2018)}]{Zhuang2018LadderNetMN}
Zhuang J (2018) {LadderNet: Multi-path networks based on U-Net for medical
  image segmentation}. ArXiv abs/1810.07810

\bibitem[{Ziomek and Middleton(2021)}]{10.1145/3476887.3476890}
Ziomek J, Middleton SE (2021) GloSAT Historical Measurement Table Dataset:
  Enhanced Table Structure Recognition Annotation for Downstream Historical
  Data Rescue, Association for Computing Machinery, New York, NY, USA, p
  49–54. \urlprefix\url{https://doi.org/10.1145/3476887.3476890}

\end{thebibliography}


\end{document}